\documentclass{article}
\usepackage[main,preprint]{neurips_2026}
\usepackage[utf8]{inputenc} % allow utf-8 input
\usepackage[T1]{fontenc}    % use 8-bit T1 fonts
\usepackage[hidelinks]{hyperref}

\usepackage{url}            % simple URL typesetting
\usepackage{booktabs}       % professional-quality tables
\usepackage{amsfonts}       % blackboard math symbols
\usepackage{nicefrac}       % compact symbols for 1/2, etc.
\usepackage{graphicx}

\usepackage{microtype}      % microtypography
\usepackage{xcolor}         % colors
\usepackage[most]{tcolorbox}
\usepackage{pifont} 

\usepackage{amsmath}
\usepackage{amssymb}

\usepackage{caption}
\usepackage[ruled,linesnumbered]{algorithm2e}
\usepackage{dsfont}
\usepackage{titletoc}
\usepackage{enumitem}
\usepackage{bm}

\usepackage{tabularx}
\usepackage{wrapfig}
\usepackage{multirow}

\usepackage{colortbl}

\usepackage{tcolorbox}
\definecolor{my_blue}{RGB}{0, 0, 0}
\definecolor{my_green}{RGB}{0,128,0}
\usepackage{dashrule} 
\usepackage{wasysym}
\usepackage{marvosym}
\definecolor{deltaBg}{RGB}{220,230,255} % subtle blue-grey

\title{Bridging Interleaved Multi-Modal Reasoning as \\ a Unified Decision Process
}

\author{Zican Hu\textsuperscript{12$\ast$} \quad 
Xuyang Hu\textsuperscript{3$\ast$} \quad
Yiming Liu\textsuperscript{4} \quad
Zuwei Long\textsuperscript{2} \quad
Wei Liu\textsuperscript{2} \quad \\
\textbf{Yunzhuo Hao}\textsuperscript{5} \quad 
\textbf{Jiawei Gu}\textsuperscript{\textbf{6}} \quad 
\textbf{Linjie Li}\textsuperscript{\textbf{6}} \quad
\textbf{Yu Cheng}\textsuperscript{\textbf{7}} \quad 
\textbf{Zhenhong Sun}\textsuperscript{\textbf{8}} \quad \\
\textbf{Weibo Gu}\textsuperscript{\textbf{2}} \quad
\textbf{Xing Sun}\textsuperscript{\textbf{2}} \quad
\textbf{Zhi Wang}\textsuperscript{\textbf{1\textrm{\Letter}}}
\\
\textsuperscript{1} Nanjing University
\textsuperscript{2} Tencent Youtu Lab
\textsuperscript{3} Shanghai AI Laboratory\\
\textsuperscript{4} Tsinghua University
\textsuperscript{5} Zhejiang University
\textsuperscript{6} University of Washington \\
\textsuperscript{7} The Chinese University of Hong Kong 
\textsuperscript{8} Australian National University
\\
\textbf{Contact:}
\texttt{zicanhu@smail.nju.edu.cn} \quad \texttt{zhiwang@nju.edu.cn}
}

\begin{document}

\maketitle

\renewcommand{\thefootnote}{}
\footnotetext{$^\ast$Equal contributions. This work was conducted during internship at Tencent Youtu Lab. $^\textrm{\Letter}$Corresponding authors.}
\renewcommand{\thefootnote}{\arabic{footnote}}

\begin{abstract}
% Unified multimodal models (UMMs) have demonstrated promising capabilities in generating interleaved text and image reasoning sequences, yet optimizing such trajectories via reinforcement learning (RL) remains an open challenge. 
Unified multi-modal models (UMMs) have shown promising interleaved text-image reasoning capabilities, yet effectively optimizing such multi-turn generation via reinforcement learning (RL) remains an open challenge.
% While image steps in interleaved reasoning are currently guided only by supervised constraints such as world model regularization, no unified optimization framework exists to jointly train them with text steps, leaving the full trajectory subject to long-horizon credit assignment difficulties and sparse reward signals across heterogeneous modalities.
Existing approaches apply RL exclusively to text steps, relegating image generation to supervised surrogates, preventing policy gradients from propagating through the full interleaved trajectory across heterogeneous modalities.
This leaves the potential of RL for UMMs largely untapped.
% We introduce \textbf{BRAID}, a framework for \textbf{b}ridging inte\textbf{r}leaved re\textbf{a}soning as a un\textbf{i}fied \textbf{d}ecision process, which formulates interleaved text-image-text (TIT) reasoning as a unified Markov Decision Process (MDP) and provides a principled objective for jointly optimizing text and image steps under a single RL framework. 
In the paper, we introduce \textbf{BRAID} (\textbf{B}ridging inte\textbf{R}le\textbf{A}ved mult\textbf{I}-modal reasoning as a unified \textbf{D}ecision process), a simple framework that casts multi-turn text-image-text reasoning as a unified Markov decision process (MDP), enabling joint optimization of textual and visual generation via a single, principled RL objective.
% A scalar advantage derived from rule-based terminal rewards is broadcast uniformly across all reasoning steps, and a vision-language judge provides additional reward signals to guide visual thinking toward more effective reasoning. 
BRAID computes a shared trajectory-level advantage and propagates it coherently into both text tokens and image denoising paths, each optimized through its modality-native policy gradient mechanism.
To further address long-horizon credit assignment, BRAID employs a vision-language model (VLM) judge that scores each intermediate image on its reasoning utility, supplying dense turn-level feedback to sharpen learning at critical visual branches.
% Experiments on perception and special reasoning benchmarks demonstrate that BRAID outperforms competitive baselines, validating that a unified MDP formulation with vision-thinking guidance is key to stronger multi-modal reasoning.
Experiments on spatial reasoning and visual perception benchmarks show that BRAID consistently outperforms various baselines, confirming that a unified MDP formulation with vision-thinking guidance is essential for effective multi-modal reasoning.
\end{abstract}

\section{Introduction} \label{sec:intro}
% 1. UMMs(Bagel, Janus, Emu, etc.) -> interleave cot -> visual superiority hypothesis
% 2. ThinkMorph/TWGI proved that interleaved CoT was valid (SFT) -> data-scaling bottleneck
% 3. RL success across language and multimodal -> unify RL
% 4. Previous work:
%    a. COOPER(text RL), DeepEyes(External tool)-> image RL formulation
%    b. DDPO,DPOK(denoise process as MDP)-> flow_grpo(flow matching RL), DiffusionNFT(optimize forward process dexterously)-> merge is technically supported
%    c. UniGRPO(concurrent work do similar thing but focus on generation task)
% 5. BRAID merge two modal-> 
%    a. challenge: long-horizon sparse reward, two modal modeling gap -> credit assignment
%    b. introuduce a vision contribution to reasoning reward
%    c. performance -> summary
% Unified multimodal models that jointly master visual understanding and generation~\cite{team2024chameleon,wang2024emu3,deng2025bagel,wu2024janus,chen2025blip3} have enabled \emph{interleaved text-image chain-of-thought} (TIT-CoT), a paradigm that interleaves textual deliberation with intermediate image synthesis to externalize perceptual steps beyond the reach of language~\cite{gu2025thinkmorph,chern2025twgi,gao2024icot}. 

\begin{wrapfigure}{r}{0.49\textwidth}
  \vspace{-23pt}
  \centering 
  \includegraphics[width=\linewidth]{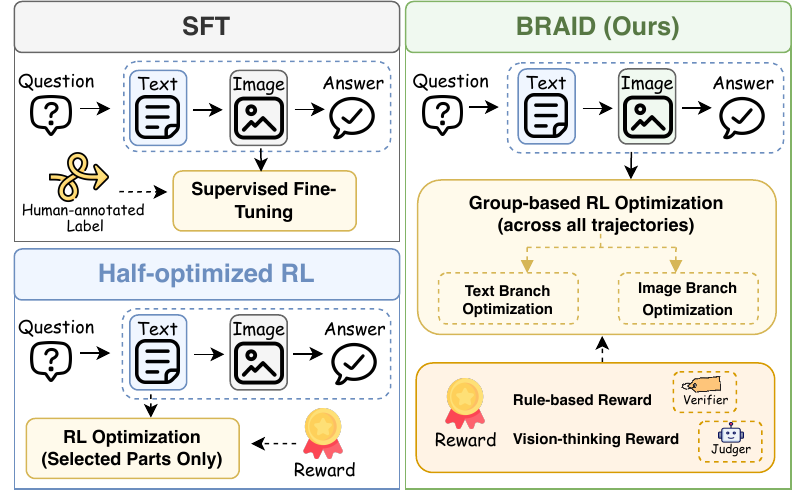}
  % \caption{Comparison of SFT, Partial RL Optimization, and BRAID.}
  \caption{SFT, Half-optimized RL, and BRAID.}
  \label{fig:intro}
  \vspace{-9pt}
\end{wrapfigure}

Recent emergence of unified multi-modal models (UMMs)~\citep{team2024chameleon,deng2025bagel} that seamlessly bridge understanding and generation across modalities within a single framework has opened a new frontier in multi-modal reasoning: interleaved Text-Image-Text Chain-of-Though (TIT-CoT), where verbal thinking alternates with intermediate image generation to externalize perceptual steps beyond the reach of language~\citep{gao2024icot,wang2024emu3,cui2025emu3.5}.
% \cite{wu2026visworld} formalize this premise as the \emph{visual superiority hypothesis}: for physically grounded tasks, visual generation affords a more veridical world model than verbal reasoning, whose bandwidth is bounded by the symbolic nature of language.
The underlying premise, often termed the \textit{visual superiority hypothesis}~\citep{wu2026visworld}, is that for physically grounded tasks, visual generation affords a more faithful world model than verbal reasoning, whose expressiveness is inherently bounded by the symbolic nature of language.
% 
% Operationalizing this premise has so far meant supervised fine-tuning on curated TIT-CoT traces~\cite{gu2025thinkmorph,chern2025twgi}, an approach whose gains are capped by a steep data-scaling bottleneck in acquiring high-quality interleaved traces.
So far, this premise has been pursued exclusively through supervised fine-tuning (SFT) on curated TIT-CoT traces~\citep{gu2025thinkmorph,chern2025twgi}, an approach bottlenecked by the cost and scarcity of high-quality interleaved data. 
% Reinforcement learning offers a principled way past this ceiling, with demonstrated success across language~\cite{guo2025deepseekr1,shao2024deepseekmath} and multimodal reasoning~\cite{meng2025mmeureka,yang2025r1onevision}.
Reinforcement learning (RL) offers a principled alternative: by learning from task rewards rather than curated data, it enables models to autonomously discover novel reasoning strategies that go beyond what static data can teach, as already shown in language~\citep{guo2025deepseekr1,shao2024deepseekmath} and multi-modal reasoning~\citep{meng2025mmeureka,yang2025r1onevision}.

% Nevertheless, existing formulations leave TIT-CoT half-optimized: gradients flow through text tokens alone, while the intermediate image generation that defines interleaved thinking stays outside the optimization loop. The resulting asymmetry confines exploration to the verbal trajectory and forfeits the very advantage on which interleaved reasoning is premised. 
However, existing formulations leave TIT-CoT half-optimized: policy gradients flow through text tokens alone, while intermediate image generation remains outside the training loop, undermining the very advantage on which interleaved reasoning is premised.
% Recent work has begun to tackle this gap from different angles. COOPER~\cite{zhang2025cooper} retains visual generation but places it outside RL, training it through a separate supervised stage; DeepEyes~\cite{zheng2025deepeyes} keeps RL end-to-end but strips the visual steps down to deterministic grounding. Neither admits image generation into the reward-driven objective. 
COOPER~\cite{zhang2025cooper} optimizes visual generation, but through a separate supervised stage.
While DeepEyes~\cite{zheng2025deepeyes} is trained end-to-end with RL, its policy gradients reach only the visual tool calling rather than the image generation process, still leaving visual reasoning outside the optimization loop.
% From the generation side, a parallel line of work has already cast image synthesis as an RL problem.  DDPO~\cite{black2023ddpo} and DPOK~\cite{fan2023dpok} formulate diffusion denoising as a Markov decision process; Flow-GRPO\cite{liu2025flowgrpo} extends this to flow matching by relaxing the deterministic ODE into an SDE to restore stochastic exploration, and DiffusionNFT~\cite{zheng2025diffusionnft} further refines this by moving optimization onto the forward process via a negative-aware flow-matching objective, eliminating likelihood estimation. 
Meantime, RL for image generation is well established: DDPO~\citep{black2023ddpo} and DPOK~\citep{fan2023dpok} cast diffusion denoising as an MDP, and subsequent work extends this to flow matching in FlowGRPO~\citep{liu2025flowgrpo} and DiffusionNFT~\citep{zheng2025diffusionnft}. 
% Together, these formulations provide a mature RL toolkit for the visual branch of a unified model, yet none of them has been integrated into an interleaved reasoning trajectory.
These formulations provide a ready-made RL toolkit for the visual branch of a unified model, yet none has been integrated into an interleaved reasoning setting. 
% The concurrent UniGRPO~\cite{liu2026unigrpo}  applies RL jointly to both modalities, but only within single-turn text-to-image generation.
The concurrent work, UniGRPO~\citep{liu2026unigrpo}, applies RL to both modalities, but limited to single-turn text-to-image generation.
These observations point to a critical missing piece: \textit{a unified RL framework that jointly optimizes textual and visual reasoning within an interleaved TIT-CoT trajectory, enabling end-to-end credit assignment across modality boundaries and exploration over the full combinatorial space of interleaved strategies}.

% We close exactly this gap with \textbf{BRAID} (\textbf{B}ridging Inte\textbf{r}leaved Re\textbf{a}soning as a Un\textbf{i}fied \textbf{D}ecision Process). 
To close this gap, we propose \textbf{BRAID} (\textbf{B}ridging inte\textbf{R}le\textbf{A}ved mult\textbf{I}-modal reasoning as a unified \textbf{D}ecision process), which casts multi-turn text-image-text reasoning as a unified MDP.
This formulation allows both textual and visual generation turns to be optimized jointly under one RL objective, without requiring separate training stages or modality-specific reward designs.
% Built on BAGEL~\cite{deng2025bagel}, BRAID generates trajectories of the form $\{\texttt{text}\!\to\!\texttt{image}\!\to\!\texttt{text}\!\to\!\cdots\!\to\!\texttt{answer}\}$, treating every step as a policy action subject to RL optimization. 
Built on BAGEL~\cite{deng2025bagel}, BRAID formulates every turn in a $\{\texttt{text}\!\to\!\texttt{image}\!\to\!\texttt{text}\!\to\!\cdots\!\to\!\texttt{answer}\}$ trajectory as a macro-action. 
% Unifying these two modalities under a single objective is non-trivial: text steps produce discrete autoregressive tokens while image steps traverse continuous denoising trajectories, and naively combining them collapses the optimization.
However, unifying the two modalities under a single objective is non-trivial: the text branch yields per-token autoregressive likelihoods while the image branch traverses continuous denoising paths via flow matching, and naively combining them collapses the optimization due to incompatible scales.
% BRAID resolves this with a \emph{unified trajectory likelihood} that joins token-level text log-probabilities with trajectory-level image log-probabilities from flow-matching policy gradients~\cite{liu2025flowgrpo,zheng2025diffusionnft}, letting a single RL objective back-propagate coherently across modality boundaries.
% BRAID resolves this with a unified trajectory likelihood that re-normalizes both modalities onto a common scale, while a group-level advantage computed over the full trajectory broadcasts a unified policy gradient coherently across modality boundaries.
BRAID addresses the modality boundary by deriving a shared trajectory-level advantage and channeling it into \textit{modality-native policy gradients}: autoregressive token generation for text and a likelihood-free flow-matching objective for image.
% A unified objective sets the stage, but the cast still has to be directed. Every image action unrolls into dozens of denoising sub-steps, stretching a TIT-CoT rollout to hundreds of actions whose text and image branches update on mismatched scales, while only a single terminal correctness reward travels back through this long chain. Credit assignment thus becomes acutely ill-posed, with no hint as to which intermediate image moved the reasoning forward. To tackle these challenges, BRAID supplies that direction through a \emph{vision-thinking process reward}: a vision-language judge scores each intermediate image on its credit contribution to the reasoning trajectory. This dense, step-level feedback sharpens credit assignment precisely where it matters most at the visual reasoning steps that define TIT-CoT.
Further, BRAID introduces a vision-thinking process reward to address credit assignment over the long-horizon TIT-CoT rollout: a VLM judge scores each intermediate image on its utility to the reasoning trajectory, providing turn-level feedback to sharpen learning at critical visual branches.
We hope BRAID serves as a step toward simple, general-purpose RL frameworks for UMMs, and inspires further investigation into principled, modality-agnostic policy optimization paradigms.

\iffalse 
Experiments show that the unified formulation and process-level visual guidance enable BRAID to outperform SFT baselines and modality-decoupled RL approaches on [\textbf{todo:}~benchmarks].
In summary, our main contributions are as follows:
\begin{itemize}
    \item We cast interleaved text-image reasoning as a single Markov decision process, coupling discrete text tokens with continuous flow-matching image trajectories under one RL objective.
    \item We introduce a vision-thinking process reward over intermediate images that resolves the credit assignment bottleneck in long, modality-mismatched rollouts.
    \item BRAID surpasses SFT baselines and modality-decoupled RL approaches on [\textbf{todo:}~benchmarks].
\end{itemize}
\fi 

Experiments show that BRAID improves BAGEL by \textbf{+5.73} avg.\ across seven benchmarks, surpassing GPT-4o with only 7B parameters, with pronounced gains on SAT (\textbf{+14.00}) and V*Bench (\textbf{+10.76}).
% Ablations confirm that image-branch RL and vision-thinking process reward are complementary, Maj@$n$ shows the solution space grows with sampling budget while baselines plateau, and case studies reveal that the optimized image branch produces faithful intermediates and transfers instruction-following to standalone generation.
Notably, Maj@$n$ shows BRAID's solution space scales efficiently with sampling budget while baselines plateau, and case studies reveal that the optimized image branch produces faithful intermediates and also transfers strong instruction-following capability to text-to-image generation.

% Experimental results demonstrate that BRAID consistently outperforms standard SFT and modality-decoupled RL approaches across [benchmarks].
% Notably, [highlight some impressive results].
% Our in-depth analysis reveals that [insightful findings].

\section{Related Work}

\textbf{Interleaved Multi-Modal Reasoning.}
% Chain-of-Thought prompting~\citep{wei2022chain} improves complex problem solving by turning a single prediction into a sequence of intermediate steps, but its text-only form is a poor fit for visual tasks where the relevant quantities, such as spatial positions, shape transformations, or perception regions, are easier to draw than to describe~\citep{hu2024visual,menon2024whiteboard}. 
Chain-of-Thought prompting~\citep{wei2022chain,kojima2022large} decomposes complex reasoning into intermediate steps, yet pure text limits expressiveness for visually grounded tasks where spatial/geometric relations are more naturally captured by images~\citep{hu2024visual,menon2024whiteboard}.
% Interleaved multimodal reasoning addresses this mismatch by letting images appear inside the reasoning trace, so that visual intermediate states can be produced and revised alongside text~\citep{li2025imagine,meng2023chain}. 
% Recent unified models push this idea further by folding image generation and understanding into a single backbone~\citep{team2024chameleon,xie2024show,deng2025bagel}, as exemplified by MetaMorph~\citep{tong2025metamorph}, Zebra-CoT~\citep{li2025zebra}, and ThinkMorph~\citep{gu2025thinkmorph}, which contribute training signals, interleaved-trajectory data, and evidence that cross-modal emission can be made intrinsic to the model.
% Interleaved multimodal reasoning addresses this gap by embedding generated images within the reasoning trace, enabling intermediate visual states to be produced, inspected, and revised jointly with text~\citep{meng2023chain}. Recent unified architectures advance this paradigm by integrating image generation and understanding within a single backbone, incrementally building the reasoning trajectory with each text/image segment conditioning subsequent turns~\citep{cheng2025visual}.
Recent unified architectures address this gap by integrating generation and understanding within a single backbone, enabling interleaved text-image reasoning traces with intermediate visual states~\citep{meng2023chain,cheng2025visual}.
These architectures follow two designs: fully autoregressive (AR) models that treat both modalities as discrete tokens under a unified next-token objective~\citep{team2024chameleon,xie2024show}, and hybrid AR-diffusion models that employ selective activation of modality-specific parameters (AR for text, diffusion for images)~\citep{deng2025bagel,li2025zebra}.
Both lines demonstrate that cross-modal reasoning can be made intrinsic to a single model through dedicated objectives and curated interleaved data~\citep{gu2025thinkmorph,li2025imagine,tong2025metamorph}.
% Across these settings the trajectory is built turn by turn, with each segment shaping the context for the next, framing learning as a sequential decision process rather than one-shot generation.
% BRAID builds on this setting but employs RL to optimize the entire interleaved trajectory, letting the model learn what to visualize through reward-driven exploration rather than supervised imitation.
BRAID builds on the hybrid backbone for its superiority in delegating each modality to its best-suited generative mechanism~\citep{wu2024janus,zhou2024transfusion,dong2024dreamllm}.

\textbf{RL Fine-tuning.}
% Verifiable-reward RL has emerged as a principled paradigm for improving reasoning capabilities beyond what supervised data alone can provide, first demonstrated in large language models~\citep{guo2025deepseekr1,shao2024deepseekmath} and subsequently extended to vision-language models whose action space remains purely textual~\citep{meng2025mmeureka,huang2025vision}. 
Verifiable-reward RL has emerged as a principled paradigm for enhancing reasoning beyond the ceiling of supervised data, first in LLMs~\citep{guo2025deepseekr1,shao2024deepseekmath} and subsequently in VLMs that are still limited to textual action space~\citep{meng2025mmeureka,huang2025vision}.
% A parallel line adapts RL to image generation, where DDPO~\citep{black2023ddpo} and DPOK~\citep{fan2023dpok} formulate diffusion denoising as a Markov decision process, FlowGRPO~\citep{liu2025flowgrpo} generalizes this formulation to flow-matching models via stochastic relaxation, DiffusionNFT~\citep{zheng2025diffusionnft} directly optimizes the forward process to accelerate reward maximization, and DanceGRPO~\citep{xue2025dancegrpo} unifies these approaches under a single policy-gradient framework. 
A parallel line adapts RL to image generation by casting diffusion denoising as an MDP~\citep{black2023ddpo,fan2023dpok,liu2026beyond}, with subsequent work generalizing to flow-matching models in FlowGRPO~\citep{liu2025flowgrpo}, accelerating learning through direct forward-process optimization in DiffusionNFT~\citep{zheng2025diffusionnft}, and unifying these variants under a single policy-gradient framework in DanceGRPO~\citep{xue2025dancegrpo}.
% While each front has matured within its own modality, neither accommodates the interleaved regime in which a unified model jointly emits textual and visual tokens along a single trajectory, a setting that motivates the multimodal-reasoning RL formulations reviewed below.
However, these two lines of work operate within single-modality generation, and neither accommodates the setting where a unified model emits interleaved text-image outputs. 
BRAID aims to bridge this divide by formulating RL over the joint multi-modal reasoning trajectory.

% Among these, DiffusionNFT's forward-process formulation is particularly appealing for unified multimodal architectures, as it propagates rewards directly to the forward pass without requiring fine-grained temporal credit assignment, providing a simple yet efficient way to bridge the inherent modality-fusion gap.

\textbf{RL for Interleaved Multi-Modal Reasoning.}
% A growing body of work applies RL to unified models that emit both text and images, and existing methods divide along architectural lines. 
A growing body of work applies RL to fine-tune unified models.
% Fully autoregressive backbones treat images as discrete tokens, admitting a single token-level objective: X-Omni~\citep{geng2025xomni} applies multi-reward GRPO to discrete AR image generation; Emu3.5~\citep{cui2025emu3.5} scales verifiable-reward RL on unified AR backbones; 
For AR backbones, Emu3.5~\citep{cui2025emu3.5} scales post-training with verifiable-reward RL, and other studies adopt the same paradigm with a hybrid reward design~\citep{geng2025xomni,nie2025towards}.
% Hybrid AR–Diffusion architectures, in contrast, often place one branch outside the RL loop: COOPER~\citep{zhang2025cooper} on BAGEL optimizes image generation through a separate supervised stage rather than RL, DeepEyes~\citep{zheng2025deepeyes} trains end-to-end with RL but routes gradients only to discrete visual tool calls, and BLIP3o-NEXT~\citep{chen2025blip3} trains its AR front-end and DiT back-end under separate per-branch signals. UniRL-Zero~\citep{wang2025unirl} moves further by coordinating RL across language and diffusion experts within a single model, though optimization is still issued per expert. 
For hybrid architectures, prior work typically excludes the visual branch from the RL loop: COOPER~\citep{zhang2025cooper} trains image generation in a separate supervised stage, and DeepEyes~\citep{zheng2025deepeyes} backpropagates policy gradients to the visual tool calling rather than the image generation process itself.
% Concurrent UniGRPO~\citep{liu2026unigrpo} does jointly optimize both modalities under a single RL objective, but only for single-turn text-to-image generation. 
Further, the concurrent work UniGRPO~\citep{liu2026unigrpo} does jointly optimize both modalities under a single RL objective, but is limited to single-turn text-to-image generation. 
% Across these settings, no existing pipeline applies RL to multi-turn interleaved traces in which earlier visual outputs condition later textual reasoning and revised text drives subsequent image emissions. 
% We address this regime by casting interleaved multimodal reasoning as a sequential decision process and optimizing a single policy across both modalities end-to-end.
In summary, no existing pipeline applies RL to the multi-turn interleaved reasoning setting where each text/image turn conditions subsequent generation.
BRAID fills this gap by casting the full interleaved trajectory as a unified decision process and deriving modality-native gradients end-to-end within a single policy.

% \paragraph{Unified Multi-Modal Models.}
% UMMs integrate visual understanding and generation in a single model.
% Early efforts adopt fully autoregressive transformers over discrete image tokens~\citep{team2024chameleon,wang2024emu3,wang2025emu3}, followed by hybrids of autoregressive text and diffusion-based image generation~\citep{zhou2024transfusion,xie2024show,wu2024janus,chen2025blip3}, and mixture-of-transformer (MoT) architectures that couple modality-specialized encoders (ViT, VAE) and experts via shared self-attention, balancing specialization with training stability~\citep{deng2025bagel}, our backbone.
% Recent works extend RL to UMMs from different angles.
% COOPER~\citep{zhang2025cooper} optimizes image generation through a separate supervised stage outside the RL loop.
% DeepEyes~\citep{zheng2025deepeyes} trains end-to-end with RL but routes gradients only to discrete visual tool calls, leaving image generation outside optimization.
% Concurrent UniGRPO~\citep{liu2026unigrpo} jointly optimizes both modalities under RL but only in single-turn text-to-image generation.
% BRAID instead propagates gradients end-to-end through both textual reasoning and image generation along a multi-turn TIT-CoT trajectory.

\section{Preliminaries}\label{sec:pre}

\paragraph{Group Relative Policy Optimization.} GRPO~\cite{shao2024deepseekmath}
samples, for each prompt $q$, a group of $G$ rollouts $\{\tau_i\}_{i=1}^{G}$ from the behavior policy $\pi_{\theta_{\mathrm{old}}}$, scores each $\tau_i$ with a scalar reward $r(\tau_i)$, and assigns a group-normalized advantage as $\hat{A}(\tau_i) = \frac{r(\tau_i)-\mu(r(\tau))}{\sigma(r(\tau)) + \kappa}$, where $\mu(\cdot)$ and $\sigma(\cdot)$ denote the mean and standard derivation, and $\kappa$ is a small constant for numerical stability. 
The policy $\pi_\theta$ is then updated by a clipped surrogate objective
with a KL penalty against a frozen reference model $\pi_{\mathrm{ref}}$ as
\begin{equation}
\label{eq:grpo_obj}
  \mathcal{J}_{\text{GRPO}}(\pi_\theta)
  = \mathbb{E}\!\Biggl[
      \frac{1}{G}\sum_{i=1}^{G}
      \frac{1}{|\tau_i|}\sum_{t=1}^{|\tau_i|}
      \min\bigl(
        \rho_{i,t}\,\hat{A}_i,\;
        \mathrm{clip}(\rho_{i,t},1\!-\!\varepsilon,1\!+\!\varepsilon)
        \,\hat{A}_i
      \bigr)
    \Biggr]
  - \eta D_{\mathrm{KL}}\!\bigl[\pi_\theta\|\pi_{\mathrm{ref}}\bigr],
\end{equation}
where $|\tau_i|$ is the length of rollout $\tau_i$,
$\rho_{i,t}=\pi_\theta(\tau_{i,t}|q,\tau_{i,<t})/
            \pi_{\theta_{\mathrm{old}}}(\tau_{i,t}|q,\tau_{i,<t})$
is the per-token importance ratio, $\varepsilon$ is the clipping
threshold, and $\eta$ controls the KL regularization strength.

\paragraph{RL for Image generation.}
Image steps are modeled by flow matching~\cite{lipman2022flow}. 
Given context $c$, the policy parameterizes a conditional velocity field $v_\theta(x_t, c, t)$ that transports a noisy sample $\epsilon\sim\mathcal{N}(0,I)$ to a clean image
$x_0$ via the probability-flow ODE along the continuous time axis $t\in [0,1]$. 
On interpolated samples $x_t=(1-t)x_0 + t\epsilon$,
flow matching regresses $v_\theta$ onto the ground-truth velocity:
\begin{equation}
\label{eq:fm}
  \mathcal{L}_{\text{FM}}(\theta)
  = \mathbb{E}_{t,x_0,x_1,c}
    \bigl\lVert v_\theta(x_t, c, t)-(\epsilon-x_0)\bigr\rVert^{2}.
\end{equation}

Applying RL to flow-matching generators requires injecting reward signals into this objective.
Rather than estimating the intractable likelihood $\log p_\theta(x_0|c)$, DiffusionNFT~\cite{zheng2025diffusionnft} turns Eq.~\eqref{eq:fm} into a reward-aware regression by splitting each sample into a positive and a negative branch weighted by a soft label $r(x_0, c)\in[0,1]$, i.e., the reward representing the optimality probability:
\begin{equation}
\label{eq:nft}
  \mathcal{L}_{\text{NFT}}(\theta)
  = \mathbb{E}\Bigl[\,
      r(x_0, c)\,\bigl\lVert v^{+}_\theta(x_t, c, t)-(\epsilon-x_0)\bigr\rVert^{2}
      +(1-r(x_0, c))\,\bigl\lVert v^{-}_\theta(x_t, c, t)-(\epsilon-x_0)\bigr\rVert^{2}
    \Bigr],
\end{equation}
with implicit predictors $v^{\pm}_\theta:=(1\mp\beta)\,v_{\theta_{\mathrm{old}}}\pm\beta\,v_\theta$ that linearly mix the current policy $v_\theta$ with a frozen behavior velocity $v_{\theta_{\mathrm{old}}}$, where $\beta$ is a hyperparameter. 
High-reward samples pull $v_\theta$ toward the data target, while low-reward ones push it away, injecting the RL signal directly into the forward training loss without any likelihood or score estimation.

\section{Method} \label{sec:method}

\begin{figure}[t]
\centering
\includegraphics[width=0.98\linewidth]{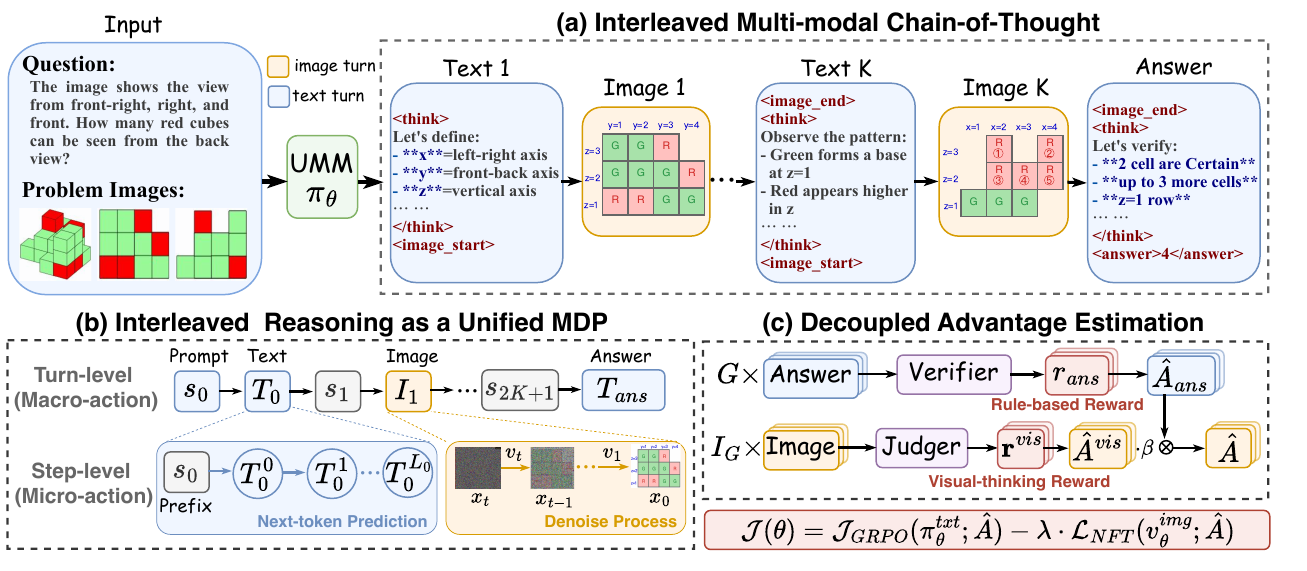}
\caption{Overview of BRAID. (a) Interleaved TIT-CoT reasoning formulated as a two-level MDP. (b) Unified policy optimization via GRPO for text and DiffusionNFT for image. (c) Decoupled advantage estimation combining terminal reward and vision-thinking process reward.}
\label{fig:framwork}
\end{figure}

% In this section, we first recap the building blocks our framework relies on (\S\ref{subsec:prelim}). We then cast TIT-CoT reasoning as a two-level Markov Decision Process that unifies text and image turns under a shared advantage (\S\ref{subsec:mdp}), and introduce a vision-thinking process reward that densifies the sparse terminal signal into per-turn credit (\S\ref{subsec:reward}). Building on these two ingredients, we derive a unified trajectory-level objective that jointly optimizes both modalities via GRPO and DiffusionNFT, yielding the overall training objective of BRAID (\S\ref{subsec:opt}).
Here, we first cast TIT-CoT reasoning as a two-level MDP that unifies text and image generation in (\S\ref{subsec:mdp}).
Then, we introduce the unified optimization mechanism that channels the shared trajectory-level advantage into modality-native policy gradients in (\S\ref{subsec:opt}).
Finally, we present the vision-thinking process reward design that enables efficient credit assignment for image turns in (\S\ref{subsec:reward}).

% \subsection{Preliminaries} \label{subsec:prelim}
\subsection{Problem Formulation}
% \paragraph{Problem Statement.} We build on a UMM $\pi_\theta$ that unifies text and image generation within a unify model~\cite{deng2025bagel}. 
We formulate a UMM policy $\pi_\theta$ that unifies text and image generation within a single model~\cite{deng2025bagel}.
% Given a prompt $c=(q,\{\mathcal{I}^{(n)}_q\}_{n=1}^{N})$ pairing a textual question $q$ with $N\!\ge\!1$ problem images, the model produces an \emph{interleaved} trajectory $\tau\sim\pi_\theta(\cdot\,|\,c)$,
Given a prompt $c=(q,\{\mathcal{I}^n_q\}_{i=1}^{M})$ pairing a textual query $q$ with $M\!\ge\!1$ images, the model produces an \emph{interleaved} reasoning trajectory $\tau\sim\pi_\theta(\cdot\,|\,c)$ as
\begin{equation}\label{eq:traj}
    \tau = \bigl(T_1,\, I_1,\, T_2,\, I_2,\, \ldots,\, T_K,\, I_K,\, T_{\mathrm{ans}}\bigr),
\end{equation}
% alternating between text chunks $t_k$ and intermediate images $I_k$ as visual reasoning turns, and ending with a textual answer $t_{\mathrm{ans}}$. 
which alternates between text chunks $T_k$ and intermediate images $I_k$ across multiple turns, and ends with a textual answer $T_{\mathrm{ans}}$.
% Both modalities are sampled from the policy $\pi_\theta$ under a shared multi-modal attention backbone~\cite{gu2025thinkmorph}. We call this structure a \emph{text-image-text chain-of-thought} (TIT-CoT).
Both modalities are generated from $\pi_\theta$ using a shared multi-modal attention backbone~\cite{gu2025thinkmorph}. 
We denote this structure as \emph{Text-Image-Text Chain-of-Thought} (TIT-CoT).

\subsection{Interleaved Reasoning as a Unified MDP}\label{subsec:mdp}
% To optimize text and image turns under a joint RL framework, we formulate a TIT-CoT trajectory as a two-level Markov Decision Process $\mathcal{M}=(\mathcal{M}_{\mathrm{turn}},\mathcal{M}_{\mathrm{tok}})$.
To optimize text and image turns under a joint RL framework, we formulate the generation of a TIT-CoT trajectory as a two-level MDP as $\mathcal{M}=(\mathcal{M}_{\mathrm{turn}},\mathcal{M}_{\mathrm{step}})$.
% The outer level abstracts each reasoning turn as a \emph{macro-action} that emits either a text chunk or an image, while the inner level unrolls it into a sequence of \emph{micro-actions}, namely tokens for text turns and denoising velocities for image turns, so that a turn-level advantage can be consistently propagated to every micro-action it contains.
The outer level $\mathcal{M}_{\mathrm{turn}}$ abstracts each reasoning turn as a \emph{macro-action} that emits either a text chunk $T_k$ or an intermediate image $I_k$.
The inner level $\mathcal{M}_{\mathrm{step}}$ unrolls each macro-action into a sequence of \emph{micro-actions}, i.e., tokens for text generation and denoising paths for image generation, where the turn-level advantage can be consistently propagated to every micro-action within that turn.

\textbf{Turn-level MDP.}
The interleaved reasoning over $2K{+}1$ turns is mapped to the following MDP:
% We map the interleaved reasoning procedure, consisting of $2K{+}1$ reasoning turns in Eq.~(\ref{eq:traj}), to the following MDP:
\begin{alignat*}{3}
&s_k \triangleq (c,\, \tau_{<k}) 
\qquad 
&&\pi(a_k \mid s_k) 
  \triangleq \pi_\theta(a_k \mid c,\, \tau_{<k}) 
\qquad 
&&P(s_{k+1} \mid s_k, a_k) 
  \triangleq \delta_{\,s_k \oplus a_k} 
\\[6pt]
&a_k \triangleq 
  \begin{cases} 
    T_k\in \mathcal{A}_{\text{txt}} \\[2pt] 
    I_k\in \mathcal{A}_{\text{img}}
  \end{cases}
\qquad 
&&\rho(s_0) 
  \triangleq \delta_{(c,\,\varnothing)} 
\qquad 
&&R(s_k, a_k) 
  \triangleq 
  \begin{cases} 
    r(\tau) & \text{if } k \!=\! 2K\!+\!1 \\[2pt] 
    0       & \text{otherwise}
  \end{cases}
\end{alignat*}
The state $s_k$ concatenates the prompt $c$ with all preceding turns $\tau_{<k}$, and $\rho(\cdot)$ is the initial state distribution.
The action $a_k$ is either a text chunk $T_k$ or an image $I_k$, sampled from the policy $\pi_\theta(a_k|s_k)$ whose text and image experts share a unified backbone. 
The transition $P$ is the deterministic append $s_{k+1}\!=\!s_k\!\oplus\!a_k$, the reward $R$ is a sparse rule-based signal assigned only at the final answer $T_{\mathrm{ans}}$.
% By collapsing each text chunk or intermediate image into a single macro-action, this abstraction renders the two modalities \emph{homogeneous} at the outer level: a single group-normalized trajectory advantage $\hat{A}(\tau)$ is computed once from the terminal reward and broadcast uniformly to every turn, regardless of modality.
By collapsing each text chunk or intermediate image into a single macro-action, this abstraction renders the two modalities \textit{homogeneous} at the outer level: text and image turns are seamlessly integrated into a single MDP that shares a unified state space, transition dynamics, and reward signal, thereby enabling a coherent RL formulation over the entire interleaved reasoning trajectory.

\textbf{Step-level MDP.}
Each turn-level action $a_k$ is realized by rolling out a lower-level MDP $\mathcal{M}_{\mathrm{step}}$ that operates at a fine-grained scale.
The specific form depends on the modality:
\begin{itemize}[itemsep=0.25em, leftmargin=1.25em]\vspace{-0.5em}
    \item For a text turn $a_k\!=\!T_k$, $\mathcal{M}_{\mathrm{step}}$ is the standard autoregressive token generation: at step $i$, the micro-state is the running prefix $s_k^i:=\left(s_k, a_k^{<i}\right)$, the micro-action is the next token $a_k^i:=T_k^i$, and $\pi_\theta$ is factorized as $\pi_\theta(T_k|s_k)=\prod_{i=1}^{|T_k|}\pi_\theta\left(T_k^i\mid s_k,T_k^{<i}\right)$, where $|T_k|$ is the text length.
    
    \item For an image turn $a_k\!=\!I_k$, $\mathcal{M}_{\mathrm{step}}$ is the iterative denoising procedure via flow matching: at denoising timestep $t\!\in\! [0,1]$, the micro-state is the noisy image $s_k^t:=x_t$, the micro-action is the predicted flow velocity $a_k^t:=v_\theta(x_t,s_k,t)$, and the state transition follows the probability-flow ODE as $x_{t_{n-1}}\!=\! x_{t_n}+v_\theta(x_{t_n},s_k,t_n)\cdot\mathrm{d}t$, with the clean image $I_k=x_0$ as the terminal state, where $n=N, N\!-\!1,...,1$ is the denoising step along the time axis. 
\end{itemize}

Let $\mathcal{K}_{\mathrm{txt}}$ and $\mathcal{K}_{\mathrm{img}}$ index text and image turns, respectively.
Because all turns reside in a single MDP, the joint policy over a complete trajectory factorizes naturally along the turn sequence:
\begin{equation}\label{eq:traj-uni}
\log \pi_\theta(\tau \mid c) \;=\; \sum\nolimits_{k=1}^{2K+1} \log \pi_\theta(a_k \mid s_k)\,.
\end{equation}
Expanding each macro-action into its step-level form then reveals two \textit{modality-native} branches:
% Factorizing the trajectory-level log-probability along the heterogeneous turns yields a text term and an image term:
\begin{equation}
\log\pi_\theta(\tau\mid c) \;=\;
\underbrace{\sum_{k\in\mathcal{K}_{\mathrm{txt}}}\sum_{i=1}^{|T_k|}
\log\pi_\theta\bigl(T_k^i\mid s_k,T_k^{<i}\bigr)}_{\text{text branch}}
\;+\;
\underbrace{\sum_{k\in\mathcal{K}_{\mathrm{img}}}\sum_{n=1}^{N}
\log\pi_\theta\bigl(x^k_{t_{n-1}}\mid x^k_{t_n},\,s_k\bigr)}_{\text{image branch (flow-matching)}}.
\label{eq:traj-decomp}
\end{equation}
% This decomposition provides two modality-specific gradient branches through which a shared turn-level advantage is consistently backpropagated, directly motivating the two-branch RL objective in \S\ref{subsec:opt}.
Although the two branches differ in computational mechanism (autoregressive token generation for text and flow-matching denoising for image generation), their structural parallelism ensures that a shared trajectory-level advantage can be consistently back-propagated through both modality-native policy gradients, a property we exploit in the unified RL objective in \S\ref{subsec:opt}.

% ================================
% ========== Subsection ==========
% ================================
% \subsection{Unified Trajectory Optimization}\label{subsec:opt}
%%% 把这个subsection放到前面，主要考虑到文章的核心亮点是 unified RL objective
%%% vision-thinking process reward是这个框架下的一个辅助，所以放在后面
%%% 这样跟abstract/introduction的写法顺序也保持一致
% \subsection{Bridging Interleaved Turns into Unified Policy Optimization}\label{subsec:opt}
\subsection{Unified Policy Optimization}\label{subsec:opt}
% With the combined advantages $\hat{A}_k$ and $A^{\mathrm{vis}}_I$ in hand, we train the unified policy $\pi_\theta=(\pi^{\mathrm{txt}}_\theta,v^{\mathrm{img}}_\theta)$ by jointly optimizing its two branches on a grouped rollouts, following the trajectory decomposition in Eq.~(\ref{eq:traj-decomp}): the text branch consumes $\hat{A}_k$ through a PPO-style clipped objective, while the image branch consumes $A^{\mathrm{vis}}_I$ through DiffusionNFT~\citep{zheng2025diffusionnft}, yielding
% \begin{equation}
% \mathcal{J}(\theta) \;=\;
% \underbrace{\mathcal{J}_{\mathrm{ppo}}\!\bigl(\pi^{\mathrm{txt}}_\theta;\hat{A}\bigr)}_{\text{text branch}}
% \;-\;\lambda_{\mathrm{img}}\cdot
% \underbrace{\mathcal{L}_{\mathrm{nft}}\!\bigl(v^{\mathrm{img}}_\theta;A^{\mathrm{vis}}\bigr)}_{\text{image branch}} .
% \label{eq:joint}
% \end{equation}
% With the advantage $\hat{A}(\tau)$ estimated from the terminal reward $r(\tau)$ via Eq.~\eqref{eq:grpo_adv}, 
We train the unified policy $\pi_\theta = (\pi_\theta^{\text{txt}},\, \pi_\theta^{\text{img}})$ by jointly optimizing both branches on grouped rollouts, leveraging the structural parallelism established in Eq.~\eqref{eq:traj-decomp}: a shared turn-level advantage $\hat{A}(s_k, a_k)$ is back-propagated through each modality-native surrogate, a GRPO-style clipped objective for the text branch and a DiffusionNFT loss~\citep{zheng2025diffusionnft} for the image branch, yielding the unified objective:
\begin{equation}
\mathcal{J}(\theta) \!=\!
\sum_{k\in\mathcal{K}_{\mathrm{txt}}}\!\!\mathcal{J}_{\mathrm{GRPO}}\bigl(\pi^{\mathrm{txt}}_\theta(T_k|s_k);\hat{A}(s_k,T_k)\bigr)
\;-\;\lambda\cdot\!\!
\sum_{k\in\mathcal{K}_{\mathrm{img}}}\!\mathcal{L}_{\mathrm{NFT}}\bigl(\pi^{\mathrm{img}}_\theta(I_k|s_k);\hat{A}(s_k,I_k)\bigr),
\label{eq:joint}
\end{equation}
where $\lambda$ balances the two modalities.
% Together, these components define BRAID: a single RL loop in which text reasoning and image generation share one trajectory, one advantage signal, and one unified backward pass.
Eqs.~\eqref{eq:traj-uni}-\eqref{eq:joint} together realize BRAID's core principle: despite distinct generative mechanisms, text and image branches are optimized within a unified RL loop, sharing one interleaved trajectory, one turn-level advantage, and one joint backward pass.

\iffalse 
To inject $\hat{A}_k$ into flow matching without likelihood
estimation, we map it to a soft label
$r_k=\sigma(\hat{A}_k/T)\in[0,1]$ with temperature $T$, and plug it
into the DiffusionNFT loss of Eq.~(\ref{eq:nft}):
%
\begin{equation}
\mathcal{L}_{\mathrm{nft}}
= \mathbb{E}_{I_k\in\mathcal{I}_{\mathcal{G}},\,t,\,\mathbf{x}_1}\!\Bigl[
r_k\,\bigl\lVert v^{+}_\theta(\mathbf{x}_t,t\mid s_k)-v\bigr\rVert^{2}
+(1-r_k)\,\bigl\lVert v^{-}_\theta(\mathbf{x}_t,t\mid s_k)-v\bigr\rVert^{2}
\Bigr],
\label{eq:nft-ours}
\end{equation}
%
with implicit predictors
$v^{\pm}_\theta=(1\!\mp\!\beta_{\mathrm{nft}})v_{\theta_{\mathrm{old}}}
\pm\beta_{\mathrm{nft}}v_\theta$ as in \S\ref{subsec:prelim}.
High-$\hat{A}_k$ image turns pull $v_\theta$ toward the generated
$I_k$, while low-$\hat{A}_k$ ones push it away, delivering turn-level
credit that jointly reflects terminal correctness and vision-thinking
quality directly through the flow-matching loss.

Up to now, these components specify BRAID: a single RL loop in which
text reasoning and image thinking share one trajectory, one
advantage stream, and one backward pass.
\fi

To inject $\hat{A}$ into flow matching without likelihood estimation, we map it to a soft reward label $r_k=\sigma(\hat{A}(s_k,I_k)/\upsilon)\in[0,1]$ with temperature $\upsilon$, and plug it into the DiffusionNFT loss of Eq.~(\ref{eq:nft}):
\begin{equation}
\mathcal{L}_{\mathrm{NFT}}(\theta)
= \mathbb{E}\Bigl[
r_k\,\bigl\lVert v^{+}_\theta\left(x_t,s_k,t\right)-(\epsilon-x_0)\bigr\rVert^{2}
+(1-r_k)\,\bigl\lVert v^{-}_\theta\left(x_t,s_k,t\right)-(\epsilon-x_0)\bigr\rVert^{2}
\Bigr],
\label{eq:nft-ours}
\end{equation}
where $x_0=I_k$ is the clean image, $\epsilon$ is Gaussian noise, $x_t=(1-t)x_0+t\epsilon$ is the interpolated sample, and $v^{\pm}_\theta=(1\!\mp\!\beta)v_{\theta_{\mathrm{old}}} \pm\beta v_\theta$ are implicit predictors as described in \S\ref{sec:pre}.
Intuitively, a high $\hat{A}_k$ steers the velocity field $v_\theta$ toward the generated image $I_k$, while a low $\hat{A}_k$ steers it away, thereby propagating the terminal reward signal to image generation directly via the flow-matching loss.

% The full procedure is summarized in Algorithm~\ref{alg:braid}.

% \begin{algorithm}[!h]
% \caption{BRAID training iteration}
% \label{alg:braid}

% \KwIn{Prompt $c$;~~Group size $G$;~~Policy $\pi_\theta$;~~Behavior $\pi_{\theta_{\mathrm{old}}}$;~~Answer scorer $R^{\mathrm{ans}}$;~~VLM judge $R^{\mathrm{vis}}$}

% \For{each iteration}{
%     Sample $G$ TIT-CoT trajectories $\{\tau_i\}_{i=1}^{G}\sim\pi_{\theta_{\mathrm{old}}}(\cdot\mid c)$

%     Score terminal answers with $R^{\mathrm{ans}}$ and grade each image turn with $R^{\mathrm{vis}}$

%     Compute $A^{\mathrm{ans}}_i$ and $A^{\mathrm{vis}}_{I_k}$ via Eqs.~(\ref{eq:adv-ans})--(\ref{eq:adv-vis}), then assemble $\hat{A}_k$ via Eq.~(\ref{eq:adv-combined})

%     Update $\theta$ by one gradient step on the joint objective in Eq.~(\ref{eq:joint})
% }
% \end{algorithm}

% ================================
% ========== Subsection ==========
% ================================
% \subsection{Credit Assignment via Vision-Thinking Process Reward} \label{subsec:reward}
\subsection{Vision-Thinking Process Reward} \label{subsec:reward}
% With the two-level MDP in place, it remains to densify the sparse terminal reward into per-turn advantages that can drive both modalities.
% \paragraph{Process Reward Design.}
% An image turn is a \emph{macro-action} whose reasoning quality is invisible to both pixel-level statistics and the terminal reward $r^{\mathrm{ans}}$.
The trajectory-level advantage $\hat{A}(\tau)$ broadcasts a single terminal-reward scalar uniformly to every turn. 
As the TIT-CoT horizon grows, this undifferentiated signal increasingly obscures which intermediate steps truly matter for the final answer.
The issue is amplified for image turns: unlike autoregressive text, image generation must produce coherent, pixel-level content in each non-autoregressive pass, demanding a far sharper learning signal than a sparse, delayed reward can provide.

% To address this credit-assignment bottleneck, we introduce a \emph{vision-thinking process reward} that provides step-level feedback to individual image turns, enabling the policy to concentrate its updates on the visual steps that are most critical to the reasoning trajectory.
To address this, we introduce a \emph{vision-thinking process reward} that scores each intermediate image on its utility to the reasoning trajectory, providing dense credit assignment for image turns.
% We therefore grade each $I_k$ via a VLM judge\footnote{We use GPT-5.4 in all experiments; see Appendix~\ref{app:exp} for the full prompt and grader ablations.} along a set of complementary criteria $\mathcal{C}$, yielding a per-turn \emph{vector} of process rewards:
Specifically, we employ a VLM judge 
\footnote{We use GPT-5.2 in all experiments; see Appendix~\ref{app:prompt} for the full prompt.}
to score each intermediate image $I_k$ according to a set of complementary criteria $\mathcal{C}$, yielding a per-turn \emph{vector} of process rewards: $\bm{r}^{\mathrm{vis}}=\bigl[r^{(c)}\bigr]_{c\in\mathcal{C}}\in \mathbb{R}^{|\mathcal{C}|}$.
%\begin{equation}
%\bm{r}^{\mathrm{vis}} \;=\;
%\bigl[r^{(c)}\bigr]_{c\in\mathcal{C}} \;\in\; \mathbb{R}^{|\mathcal{C}|},
%\label{eq:rvis}
%\end{equation}
% each coordinate of which is group-normalized independently before contributing to the image-turn advantage. 
% In this work we instantiate $\mathcal{C}=\{\texttt{ac},\texttt{rc},\texttt{vf}, \texttt{tr}\}$, covering four complementary aspects of a visual thought: (1) \texttt{Action Correctness} ($r^{(\mathrm{ac})}_k$), measuring whether the right visual operation is chosen and executed accurately; (2) \texttt{Reasoning Contribution} ($r^{(\mathrm{rc})}_k$), measuring whether the image provides useful evidence toward the final answer; (3) \texttt{Visual Fidelity} ($r^{(\mathrm{vf})}_k$), measuring whether the image is coherent and free of hallucinated content; and (4) \texttt{Trustworthiness} ($r^{(\mathrm{tr})}_k$), measuring whether the image is unlikely to mislead a reasonable reasoner.
In practice, we instantiate $\mathcal{C}=\{\texttt{vc},\texttt{vf},\texttt{ru}, \texttt{tr}\}$, covering four salient aspects of a visual thought: 
\begin{itemize}[itemsep=0em, leftmargin=1.25em]\vspace{-0.5em}
    \item \texttt{Visual Correctness} $r^{(\mathrm{vc})}$: whether the intended visual operation is executed accurately;

    \item \texttt{Visual Fidelity} $r^{(\mathrm{vf})}$: whether the image is coherent and free of hallucinated content;

    \item \texttt{Reasoning Utility} $r^{(\mathrm{ru})}$: whether the image provides useful evidence to the final answer;

    \item \texttt{Trustworthiness} $r^{(\mathrm{tr})}$: whether the image is unlikely to mislead a reasonable reasoner.
\end{itemize}

\iffalse 
\paragraph{Advantage Estimation.}
For the terminal answer reward $r^{\mathrm{ans}}$, we follow the
standard group-relative scheme: within each rollout group
$\mathcal{G}$ of size $G$ sharing the same prompt, the \emph{answer
advantage} is
\begin{equation}
A^{\mathrm{ans}}_i \;=\;
\frac{r^{\mathrm{ans}}_i-\mu_{\mathcal{G}}(r^{\mathrm{ans}})}
     {\sigma_{\mathcal{G}}(r^{\mathrm{ans}})},
\qquad i\in\mathcal{G},
\label{eq:adv-ans}
\end{equation}
and broadcast uniformly to all text tokens in the $i$-th rollout.
\fi

% The image branch, by contrast, warrants a more careful design.
% Image turns within a rollout differ in both quantity and conditioning context, making a per-rollout baseline statistically fragile and prone to conflating unrelated sub-problems. 
% Moreover, the visual reward is intrinsically multi-objective, and collapsing $\mathcal{C}$ into a scalar before normalization would let gains on one axis silently offset losses on another, washing out the per-axis structure of the learning signal. 
% We therefore pool all image turns in the group into a single set $\mathcal{I}_{\mathcal{G}}$ and standardize each criterion independently over it before aggregating into a per-image advantage:
%\begin{equation}
%A^{\mathrm{vis}}_{I} \;=\;
%\sum_{c\in\mathcal{C}} \alpha_c \cdot
%\frac{r^{(c)}_{I}-\mu_{\mathcal{I}_{\mathcal{G}}}(r^{(c)})}
%     {\sigma_{\mathcal{I}_{\mathcal{G}}}(r^{(c)})},
%\qquad I\in\mathcal{I}_{\mathcal{G}},
%\label{eq:adv-vis}
%\end{equation}
\textbf{Decoupled Advantage Estimation.}
Since the visual process reward is intrinsically multi-objective, its components typically operate on disparate and dynamically shifting scales during training.
This complicates the development of a flexible weighting scheme for aggregating the process reward components.
To circumvent this issue, we pool all image turns within the group of rollouts into a visual set $\mathcal{I}_{\mathcal{G}}$, compute the advantages for each reward component independently, and simply merge them with equal weight at the advantage level:
\begin{equation}
\hat{A}^{\mathrm{vis}}(s_k,I_k) \;=\;
\sum\nolimits_{c\in\mathcal{C}}
\frac{r^{(c)}_k-\mu_{\mathcal{I}_{\mathcal{G}}}(r^{(c)})}
     {\sigma_{\mathcal{I}_{\mathcal{G}}}(r^{(c)})},
\qquad I_k\in\mathcal{I}_{\mathcal{G}},
\label{eq:adv-vis}
\end{equation}
where $\mu_{\mathcal{I}_{\mathcal{G}}}(\cdot)$ and $\sigma_{\mathcal{I}_{\mathcal{G}}}(\cdot)$ denote the mean and standard derivation calculated from $\mathcal{I}_{\mathcal{G}}$.
Analogously, we incorporate the terminal reward $r(\tau)$ with the visual process reward in a decoupled way as
\begin{equation}
\hat{A}(s_k,a_k) \;=\;
\begin{cases}
\hat{A}^{\mathrm{ans}}(s_k, T_k), & \text{for text turns},\\[2pt]
\hat{A}^{\mathrm{ans}}(s_k, I_k) + \lambda_{\text{vis}}\cdot\hat{A}^{\mathrm{vis}}(s_k,I_k), & \text{for image turns},
\end{cases}
\label{eq:adv-combined}
\end{equation}
where $\hat{A}^{\mathrm{ans}}(\cdot)$ is the advantage calculated from the terminal reward $r(\tau)$, and $\lambda_{\text{vis}}$ balances two advantages.
By decoupling advantage estimation, each reward signal is normalized relative to its own baseline before aggregation.
This architecture mitigates sensitivity to scale mismatch and temporal fluctuations, facilitating more robust and interpretable control over the process reward integration.

\iffalse
% with the weights $\{\alpha_c\}_{c\in\mathcal{C}}$
% satisfying $\sum_c\alpha_c=1$. 

In summary, the per-token advantage used by the policy gradient is
%
\begin{equation}
\hat{A}_{k}(\tau_i) \;=\;
\begin{cases}
A^{\mathrm{ans}}_i, & a_k \text{ is a text token},\\[2pt]
A^{\mathrm{ans}}_i + \beta\,A^{\mathrm{vis}}_{I_k}, & a_k \text{ belongs to image turn } I_k,
\end{cases}
\label{eq:adv-combined}
\end{equation}
%
where $\beta\!\ge\!0$ balances the reasoning turn-level signals; 
$A^{\mathrm{vis}}_{I_k}$ additionally serves as the image branch through DiffusionNFT.
\fi

\section{Experiments}\label{sec:exp}

% \subsection{Experimental Settings}\label{subsec:exp_set}
\textbf{Dataset and Evaluation.} 
% Following ThinkMorph and its follow-ups~\citep{gu2025thinkmorph}, our training corpus spans eight task families (\texttt{Matterport3D}~\citep{chang2017matterport3d}, \texttt{RealSee3D}~\citep{Li2025realsee3d}, \texttt{Perspective}, \texttt{DirectionalQuery}, \texttt{$360{+}x$}~\citep{chen2024360+}, \texttt{PositionedDirection}, \texttt{Jigsaw}~\citep{wang2025jigsaw}, \texttt{VisualSearch}~\citep{shao2024visual}), partitioned into SFT and RL pools under three regimes ($\mathcal{D}^{2{:}1}$, $\mathcal{D}^{1{:}1}$, $\mathcal{D}^{\cap}$). 
Following the literature~\citep{gu2025thinkmorph}, our training corpus spans a wide range of eight task families: \texttt{Matterport3D}~\citep{chang2017matterport3d}, \texttt{RealSee3D}~\citep{Li2025realsee3d}, \texttt{Perspective}, \texttt{DirectionalQuery}, \texttt{$360{+}x$}~\citep{chen2024360+}, \texttt{PositionedDirection}, \texttt{Jigsaw}~\citep{wang2025jigsaw}, and \texttt{VisualSearch}~\citep{shao2024visual}.
In each task, the dataset is partitioned into SFT and RL pools under three ratios: $\mathcal{D}^{2{:}1}$, $\mathcal{D}^{1{:}1}$, and $\mathcal{D}^{\cap}$.
We default to the partial-overlap regime $\mathcal{D}^{\cap}$, which bridges supervised initialization and on-policy exploration. 
We evaluate on seven benchmarks spanning two categories: 1) spatial reasoning, including \texttt{MMSI-Bench}~\citep{mmsi}, \texttt{SAT}~\citep{sat}, \texttt{MMVP}~\citep{mmvp}, and \texttt{CV-Bench 3D}~\citep{cvbench}; 2) visual perception, covering \texttt{BLINK}~\citep{blink}, \texttt{V$^{*}$Bench}~\citep{vstar}, and \texttt{CV-Bench 2D}~\citep{cvbench}. All evaluations are conducted under the \texttt{VLMEvalKit} framework~\citep{vlmevalkit}.
% Detailed per-task sample counts and allocation specifications are provided in Appendix~\ref{app:data}.
Appendix~\ref{app:data} shows details of datasets and benchmarks.

\textbf{Baselines and Training Recipe.}
% We compare our method against two categories of baselines: 1) proprietary and open-source VLMs: \texttt{GPT-5.4}, \texttt{GPT-4o}~\citep{gpt4o}, and \texttt{Qwen2.5-VL}~\citep{qwen25vl}; 2) unified multi-modal models: \texttt{Janus-Pro}~\citep{wu2024janus}, \texttt{Chameleon}~\citep{team2024chameleon}, and \texttt{BAGEL}~\citep{deng2025bagel}. 
We compare our method to two categories of baselines: 1) proprietary and open-source VLMs: \texttt{GPT-5.4}, \texttt{GPT-4o}~\citep{gpt4o}, and \texttt{Qwen2.5-VL}~\citep{qwen25vl}; 2) competitive UMMs: \texttt{Janus-Pro}~\citep{wu2024janus}, \texttt{Chameleon}~\citep{team2024chameleon}, and \texttt{BAGEL}~\citep{deng2025bagel}. 
% For RL training, we remove the KL loss term and the rollout batch size is $64$, the update batch size is $32$, with $8$ rollouts per prompt. For NFT, we set $\beta=0.4$ and use $20$ sampling steps for training and evaluation. All experimental details are documented in Appendix~\ref{app:exp}.
For RL training, we remove the KL loss term, and set the key hyperparameters: rollout batch size $64$, update batch size $32$, $8$ rollouts per prompt. 
For DiffusionNFT, we set $\beta=0.4$ and use $20$ sampling steps for training and evaluation. All experimental details are documented in Appendix~\ref{app:exp}.

\subsection{Main Results}\label{subsec:results}
\begin{table}[ht]
\centering
\vspace{-10pt}
\caption{Performance comparison on multimodal spatial reasoning and visual perception benchmarks. Best results in \textbf{bold} and second best \underline{underlined}.}
\label{tab:main_results}
\setlength{\tabcolsep}{3pt}
\renewcommand{\arraystretch}{1.1}
\resizebox{\textwidth}{!}{%
\begin{tabular}{lc cccc ccc >{\columncolor{cyan!10}}c}
\toprule
\multirow{2}{*}{\textbf{Model}} & \multirow{2}{*}{\textbf{Size}}
 & \multicolumn{4}{c}{\textbf{Spatial Reasoning}}
 & \multicolumn{3}{c}{\textbf{Visual Perception}}
 & \multirow{2}{*}{\textbf{Avg.}} \\
\cmidrule(lr){3-6} \cmidrule(lr){7-9}
 & & \textbf{MMSI} & \textbf{SAT} & \textbf{MMVP} & \textbf{CV-Bench 3D}
 & \textbf{BLINK} & \textbf{VStar} & \textbf{CV-Bench 2D} & \\
\midrule
\multicolumn{10}{c}{\textit{Visual Understanding-only VLM}} \\
\midrule
GPT-5.2     & --  & \textbf{32.90}    & 43.33             & \underline{82.33} & \underline{86.92} & \textbf{63.70}    & 65.97             & \underline{76.81} & 64.57 \\
GPT-4o      & --  & 25.40             & 42.00             & \textbf{85.33}    & 83.08             & 58.13             & 67.54             & 75.77             & 62.46 \\
Qwen2.5-VL  & 7B  & 28.80             & 52.67             & 77.00             & 74.25             & 56.81             & \underline{76.96} & 68.25             & 62.11 \\
Qwen2.5-VL  & 72B & 31.70             & \textbf{66.00}    & 81.67             & \textbf{87.58}    & \underline{61.81} & \textbf{85.86}    & \textbf{79.09}    & \textbf{70.53} \\
\midrule
\multicolumn{10}{c}{\textit{Unified Models}} \\
\midrule
Janus-Pro   & 7B  & 27.70 & 21.33 & 63.33 & 66.50 & 38.82 & 40.83 & 63.82 & 46.05 \\
Chameleon   & 7B  & 22.20 & 10.67 & 41.67 & 46.58 & 23.19 & 22.51 & 25.59 & 27.49 \\
BAGEL       & 7B  & 26.60 & 44.67 & 75.67 & 83.41 & 54.76 & 57.59 & 73.53 & 59.46 \\
\midrule
\rowcolor{cyan!10}
\multicolumn{10}{c}{\textbf{Our Methods}} \\
\midrule
SFT
  & 7B & 29.20 & 49.67 & 78.67 & 81.67 & 57.02 & 62.82 & 74.85 & 61.99 \\
$\Delta$ (vs BAGEL) &
  & \textcolor{my_green}{\textbf{+2.60}}
  & \textcolor{my_green}{\textbf{+5.00}}
  & \textcolor{my_green}{\textbf{+3.00}}
  & \textcolor{red}{\textbf{-1.74}}
  & \textcolor{my_green}{\textbf{+2.26}}
  & \textcolor{my_green}{\textbf{+5.23}}
  & \textcolor{my_green}{\textbf{+1.32}}
  & \textcolor{my_green}{\textbf{+2.53}} \\
BRAID
  & 7B & \underline{31.80} & \underline{58.67} & 81.33 & 82.17 & 57.76 & 68.35 & 76.26 & \underline{65.19} \\
$\Delta$ (vs BAGEL) &
  & \textcolor{my_green}{\textbf{+5.20}}
  & \textcolor{my_green}{\textbf{+14.00}}
  & \textcolor{my_green}{\textbf{+5.66}}
  & \textcolor{red}{\textbf{-1.24}}
  & \textcolor{my_green}{\textbf{+3.00}}
  & \textcolor{my_green}{\textbf{+10.76}}
  & \textcolor{my_green}{\textbf{+2.73}}
  & \textcolor{my_green}{\textbf{+5.73}} \\
\bottomrule
\vspace{-10pt}
\end{tabular}
}
\end{table}

\textbf{Comparison among Unified Models.}
% Table~\ref{tab:main_results} summarizes performance on spatial reasoning and visual perception benchmarks. 
Table~\ref{tab:main_results} summarizes the performance of all tested methods on all benchmarks. 
% Among unified multimodal models of comparable scale, BAGEL already establishes a strong baseline (59.46 avg.), substantially outperforming Janus-Pro (46.05) and Chameleon (27.49), confirming that its architecture is amenable to absorbing diverse multimodal signals even from limited data.
Among UMMs of comparable scale, BAGEL already establishes a strong baseline (59.46 avg.), substantially outperforming Janus-Pro (46.05) and Chameleon (27.49), which validates its hybrid AR-diffusion architecture as a capable backbone for multi-modal reasoning (even from limited data) and motivates our choice of building upon it.
% Nevertheless, a considerable gap persists between BAGEL and understanding-specialized VLMs such as Qwen2.5-VL-72B (70.53) and GPT-4o (62.46), indicating that unified architectures do not inherently excel at visual understanding without targeted optimization.
Nevertheless, a notable gap remains between BAGEL and understanding-only VLMs: even the same-scale Qwen2.5-VL-7B (62.11) surpasses it by 2.65 points, with the margin widening further against Qwen2.5-VL-72B (70.53).
It suggests that the generative versatility of unified architectures does not automatically translate into strong visual reasoning without dedicated post-training.

% \paragraph{Spatial Reasoning and Perception Performance.}
\textbf{Spatial Reasoning and Visual Perception Performance.}
% A prerequisite for effective on-policy exploration is a backbone that can already produce reasonable rollouts; our SFT stage fulfills this role, yielding consistent gains over BAGEL across nearly all benchmarks (\textbf{+2.53} avg.) and providing a stable initialization for RL. 
Effective on-policy exploration presupposes a policy that can already generate reasonable rollouts. 
Our SFT stage fulfills precisely this prerequisite: it improves over BAGEL by \textbf{+2.53} points on average, with gains observed on nearly every benchmark, thereby furnishing a reliable initialization for the subsequent RL phase.
% Building on this foundation, BRAID further amplifies improvements through reinforcement learning, achieving \textbf{+2.73} point over BAGEL on average and surpassing GPT-4o (\textbf{65.19 vs. 62.46}) with only a 7B model.
From this initialization, RL yields an additional \textbf{+2.73}-point average gain over BAGEL, lifting BRAID to \textbf{65.19} and surpassing GPT-4o (62.46) despite employing only 7B parameters.
% Notably, the most pronounced gains appear on SAT (\textbf{+14.00}) and VStar (\textbf{+10.76}), both of which require mental viewpoint transformation or fine-grained visual perception, confirming that RL effectively expands the reasoning frontier of unified models on precisely the axes where they previously lagged behind specialist systems. 
Notably, the most pronounced gains concentrate on SAT (\textbf{+14.00}) and V*Bench (\textbf{+10.76}), which demand mental viewpoint transformation or fine-grained visual discrimination -- precisely the capabilities where unified models have historically trailed understanding-only counterparts.
This confirms that BRAID can narrow such deficits without any architectural change.
% The only minor regression occurs on CV-Bench 3D ($-$1.24), which we attribute to a distribution mismatch in the SFT stage, though RL partially alleviates this gap.
The sole regression appears on CV-Bench 3D (-1.24), which we attribute to a distributional gap in the SFT data; encouragingly, BRAID recovers part of this loss, reducing the deficit from -1.74 (SFT) to -1.24.

\begin{figure}[ht]
    \vspace{-5pt}
    \centering
    \includegraphics[width=0.99\linewidth]{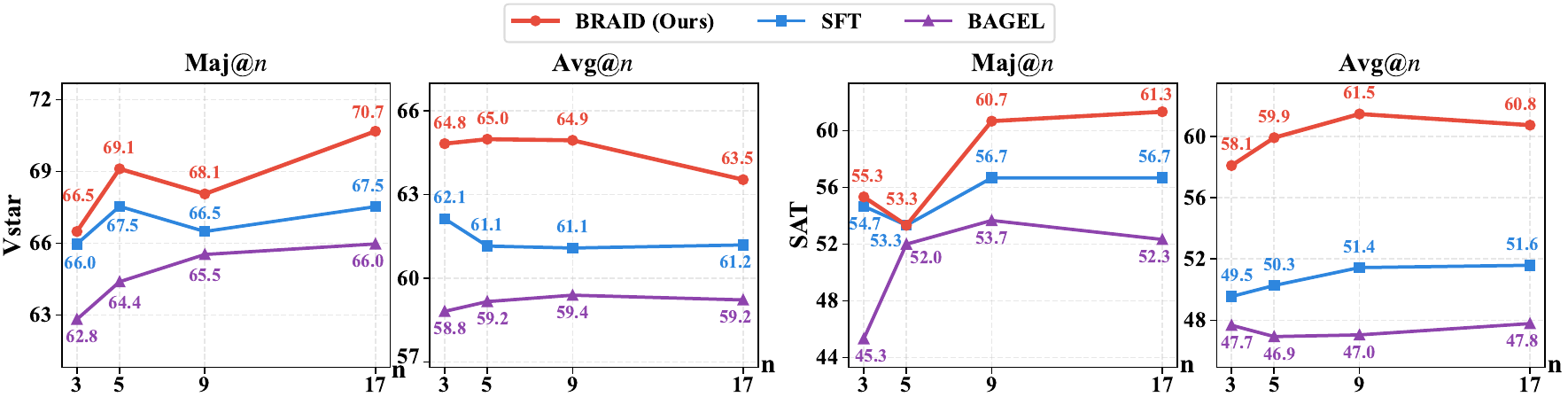}
    \caption{Maj@$n$ and Avg@$n$ curves on VStar and SAT benchmarks with varying number of samples $n=\{3,5,9,17\}$. Complete results are available in Table~\ref{tab:maj_avg}.}
    \label{fig:maj_avg}
    \vspace{-5pt}
\end{figure}

\textbf{Reasoning Capacity Boundary.}
Figure~\ref{fig:maj_avg} examines the reasoning ceiling via Maj@$n$ and Avg@$n$ metrics across varying samples. 
% BRAID's Maj@$n$ curve on VStar rises from 66.5 to 70.7 as $n$ increases from 3 to 17, whereas SFT and BAGEL plateau, suggesting that RL endows the model with a broader and more exploitable solution space. 
On VStar, BRAID's Maj@$n$ curve climbs steadily from 66.5 to 70.7 as $n$ grows from 3 to 17, while curves of SFT and BAGEL plateau early, suggesting that RL endows the model with a broader and more exploitable solution space. 
% A similar trend holds on SAT, where BRAID reaches 61.3 at Maj@17, surpassing SFT (56.7) and BAGEL (52.3) by \textbf{+4.6} and \textbf{+9.0} respectively.
SAT exhibits the same pattern: BRAID attains 61.3 at Maj@17, exceeding SFT (56.7) and BAGEL (52.3) by \textbf{+4.6} and \textbf{+9.0}, respectively.
% The Avg@$n$ curves corroborate this finding, showing that BRAID's per-sample quality is consistently higher rather than relying on diversity for majority-vote gains. 
The Avg@$n$ curves reinforce this interpretation: BRAID's mean per-sample quality is consistently higher, ruling out the possibility that its majority-vote advantage stems solely from greater output diversity.
% These results confirm that BRAID does not merely memorize reward-aligned patterns but genuinely acquires stronger reasoning capabilities that can be surfaced through sampling.
Taken together, these scaling trends confirm that RL not only aligns the model with the reward signal but also incentivizes genuinely stronger reasoning capabilities.

\subsection{Ablation Study}\label{subsec:ablation}
\begin{figure}[ht]
  \centering 
  \vspace{-8pt}
  \includegraphics[width=0.93\linewidth]{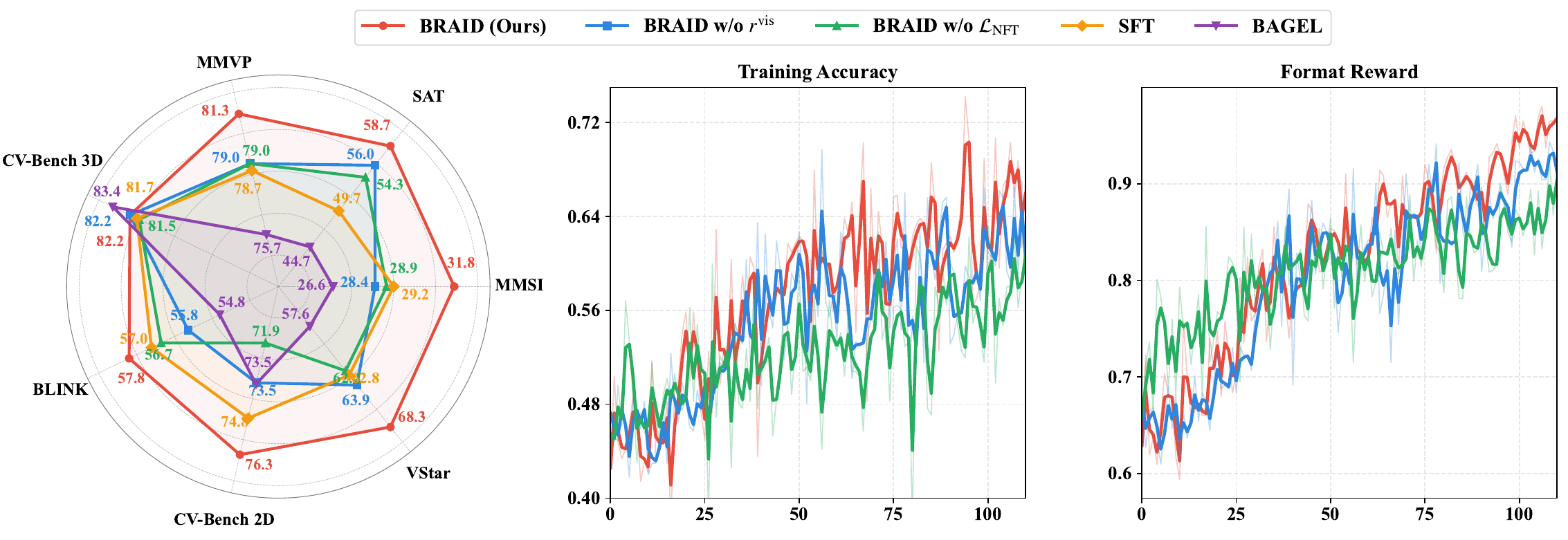}
  \caption{Ablation results. Left: Final test accuracy across seven benchmarks. Middle: Training average accuracy. Right: Training average format reward. Complete results are available in Table~\ref{tab:ablation}.}
  \vspace{-12pt}
  \label{fig:ablation}
\end{figure}

To isolate the contribution of each component, we evaluate two ablated variants under the same training recipe: (1)~\texttt{BRAID w/o $\mathcal{L}_{\text{NFT}}$}, which freezes the image generation parameters during RL and only optimizes the text branch, following the design of COOPER~\citep{zhang2025cooper}; and (2)~\texttt{BRAID w/o $r^{\text{vis}}$}, which removes the vision-thinking process reward and relies solely on the terminal reward $r(\tau)$ for advantage estimation. Results are reported in Table~\ref{tab:ablation} and Figure~\ref{fig:ablation} with key findings as
\begin{itemize}[itemsep=0em, leftmargin=1.25em]\vspace{-0.5em}
\item Dropping $r^{\text{vis}}$ (BRAID \textit{vs.} w/o $r^{\text{vis}}$) results in a moderate decline (65.19 vs.\ 62.68 avg.), mainly on VStar ($-$4.48) and SAT ($-$2.67), two benchmarks reliant on fine-grained visual reasoning.
Training accuracy, however, does not collapse: the shared trajectory-level advantage still couples the two modalities, allowing the terminal reward signal to backpropagate to image turns.

\item Excluding the image branch (w/o $\mathcal{L}_{\text{NFT}}$) incurs a steeper decline, notably on VStar ($-$6.00) and CV-Bench 2D ($-$4.32). 
However, its format reward curve converges faster, presumably because the text branch receives an undiluted reward as the sole trainable modality. 
In contrast, BRAID trades slower early format convergence for a higher accuracy ceiling, confirming that jointly optimizing the image branch produces richer visual intermediates that benefit downstream reasoning.

\item $\mathcal{L}_{\text{NFT}}$ opens the gradient pathway for image-turn optimization, while $r^{\text{vis}}$ supplies a dedicated signal that steers those gradients toward reasoning-relevant outputs. 
Neither alone suffices: BRAID outperforms the two ablations by \textbf{+3.09} and \textbf{+2.51} on average, confirming that both components are essential.
Notably, removing $\mathcal{L}_{\text{NFT}}$ is more damaging than removing $r^{\text{vis}}$, underscoring that bringing the image branch into the RL loop (BRAID's central premise) is the more critical factor.
\end{itemize}

\subsection{Analysis}\label{subsec:analysic}
\begin{wrapfigure}{r}{0.52\textwidth}
    \vspace{-25pt}
    \centering
    \includegraphics[width=1.02\linewidth]{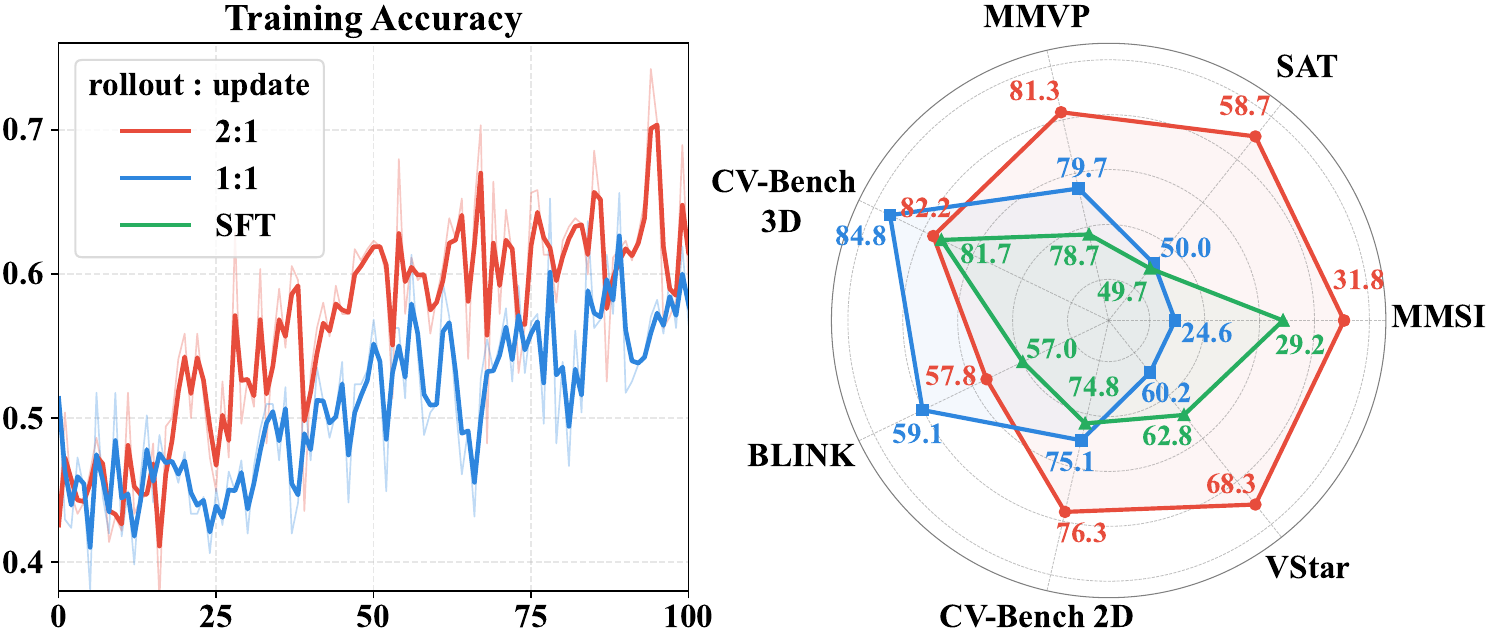}
  \caption{Complete results are available in Table~\ref{tab:on_off_policy}.}
  \vspace{-1pt}
    \label{fig:on_off_policy}
\end{wrapfigure}

\textbf{Rollout-to-Update Ratio.}
We study how the rollout-to-update batch ratio affects training stability. Our default (2:1, rollout 64 / update 32) maintains a rollout buffer that stabilizes gradient estimates.
% We investigate the effect of the rollout-to-update batch size ratio on training stability and final performance. 
% The default setting (Ratio = 2:1, rollout batch 64, update batch 32) generates twice as many rollouts as consumed per update, maintaining a buffer that stabilizes gradient estimates. 
We compare this against a more aggressive setting (Ratio = 1:1, rollout and update batch is 64) that immediately consumes all rollouts in a single update.
% As shown in Table~\ref{tab:on_off_policy} and Figure~\ref{fig:on_off_policy}, the 1:1 ratio leads to a noticeable decline on several benchmarks, resulting in a lower average (\textbf{61.93} vs.\ \textbf{65.19}). 
As shown in Table~\ref{tab:on_off_policy} and Figure~\ref{fig:on_off_policy}, the 1:1 ratio leads to a notable decline of avg. 65.19 to 61.93. 
% The training accuracy curve exhibits pronounced oscillations and recovers slowly, suggesting that aggressive updates on stale rollouts destabilize the policy before it can consolidate improvements.
The accuracy curve exhibits pronounced oscillations and recovers slowly, indicating that aggressive updates can destabilize the policy.
% Interestingly, despite the overall degradation, CV-Bench 3D improves to \textbf{84.83} under the 1:1 ratio, even surpassing BRAID and recovering the regression observed in the SFT stage. 
Interestingly, despite the overall degradation, CV-Bench 3D improves to \textbf{84.83} under the 1:1 ratio, even surpassing the default 2:1 setting (82.2). 
We conjecture that higher update frequency encourages aggressive exploration that coincidentally favors this task but destabilizes others; the 2:1 ratio therefore better balances exploration and stable improvement.
% We conjecture that the higher update frequency encourages more aggressive exploration that coincidentally favors this particular task distribution, but at the expense of training stability across other benchmarks. 
% The 2:1 ratio therefore strikes a more favorable balance between sufficient exploration and stable policy improvement.

\begin{wrapfigure}{r}{0.45\textwidth}
    \vspace{-12pt}
    \centering
    \hspace{-12pt}
    \includegraphics[width=0.98\linewidth]{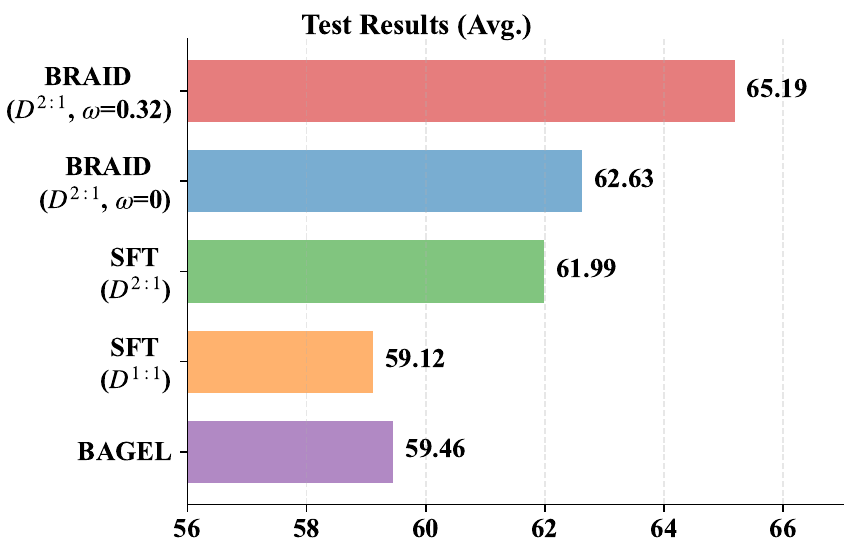}
    \vspace{-2pt}
  \caption{Complete results are available in Table~\ref{tab:data_split_result}.}
    \label{fig:data_split_result}
\end{wrapfigure}

\textbf{Data Allocation Strategy.}
We compare three data-allocation regimes (detailed in Appendix~\ref{app:data}): a disjoint split with equal budget ($\mathcal{D}^{1:1}$, $\omega{=}0$), a disjoint split favoring SFT ($\mathcal{D}^{2:1}$, $\omega{=}0$), and a replay regime ($\mathcal{D}^{\cap}$, $\omega{\approx}0.32$) where the RL pool partially overlaps with SFT data. Results are reported in Table~\ref{tab:data_split_result} and Figure~\ref{fig:data_split_result}.
% The $\mathcal{D}^{1:1}$ SFT checkpoint already exhibits weaknesses on several benchmarks (e.g., CV-Bench 2D drops to 65.24), indicating that an insufficient SFT budget produces a poor initialization that limits downstream RL. We therefore did not proceed with RL from this checkpoint.
The $\mathcal{D}^{1:1}$ SFT checkpoint already exhibits weakness on several benchmarks (e.g., CV-Bench 2D drops to 65.24), indicating that an insufficient SFT budget yields a suboptimal initialization that constrains downstream RL gains. We therefore did not proceed with RL training from this checkpoint.
With adequate SFT initialization ($\mathcal{D}^{2:1}$), RL brings consistent gains and reaches \textbf{62.63} avg. 
% The replay regime $\mathcal{D}^{\cap}$ further improves this to \textbf{65.19} avg., with particularly large gains on SAT (\textbf{+9.34} over $\mathcal{D}^{2:1}$ RL) and VStar (\textbf{+3.43}).
The replay regime $\mathcal{D}^{\cap}$ further improves this to \textbf{65.19} avg., with pronounced gains on SAT (\textbf{+9.34} over $\mathcal{D}^{2:1}$ RL) and VStar (\textbf{+3.43}).
% This confirms that replaying a portion of SFT data during RL bridges imitation learning and on-policy exploration, allowing the model to revisit familiar prompts under its own rollouts and thereby consolidate supervised knowledge while still discovering novel reasoning strategies.
This confirms that replaying a portion of SFT data during RL bridges imitation learning and on-policy exploration, enabling the model to revisit familiar prompts under its own policy and thereby consolidate supervised knowledge while continuing to discover novel reasoning strategies.

\subsection{Case Study}\label{subsec:case}
\begin{wrapfigure}{r}{0.49\textwidth}
  \vspace{-10pt}
  \centering
  \setlength{\tabcolsep}{0.5pt}
  \begin{tabular}{@{}c@{\hspace{1pt}}c@{\hspace{1pt}}c@{\hspace{1pt}}c@{\hspace{1pt}}c@{\hspace{1pt}}c@{}}
  & \textbf{\tiny BAGEL}  & \textbf{\tiny w/o $\mathcal{L}_{\text{NFT}}$} & \textbf{\tiny w/o $r^{\text{vis}}$} & \textbf{\tiny BRAID} \\[1pt]
  \rotatebox{90}{\parbox{1.3cm}{\centering\tiny A cat sitting inside a fishbowl}} &
  \includegraphics[width=0.185\linewidth]{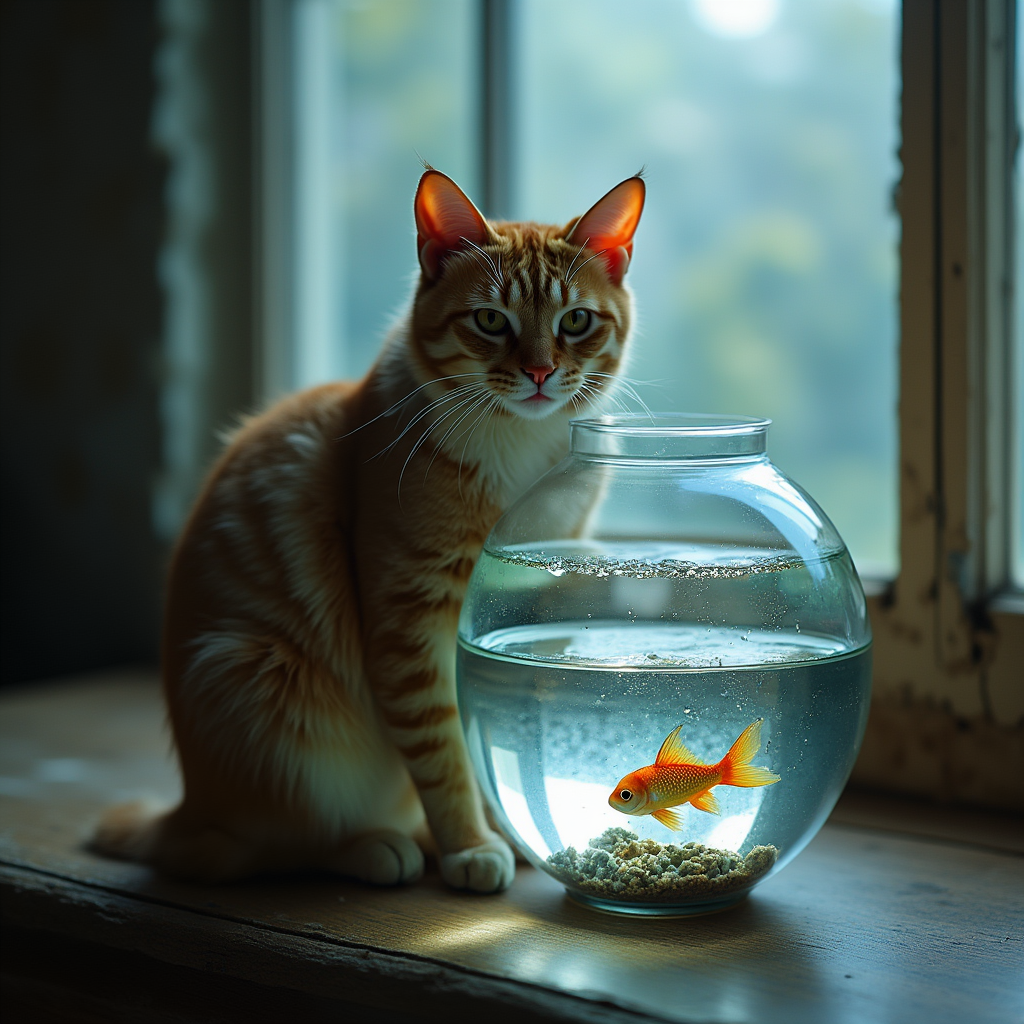} &
  \includegraphics[width=0.185\linewidth]{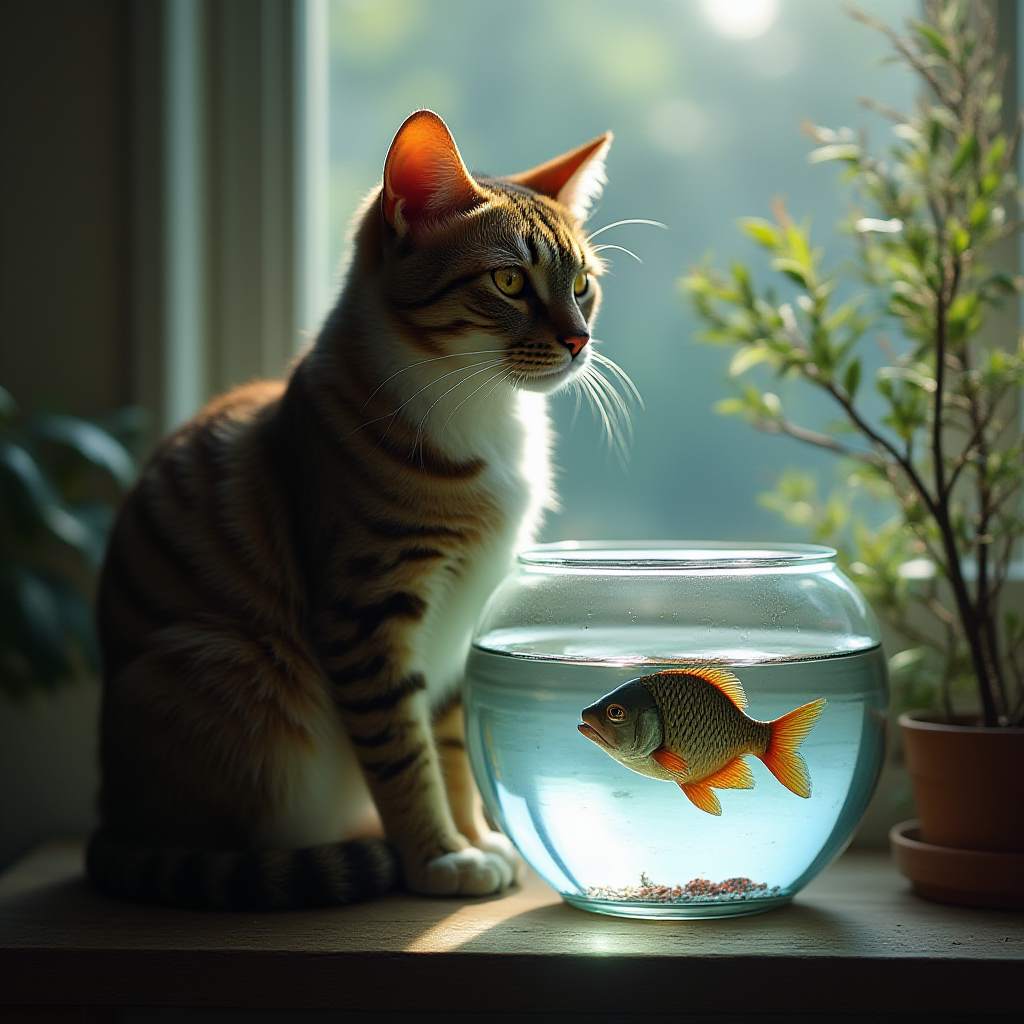} &
  \includegraphics[width=0.185\linewidth]{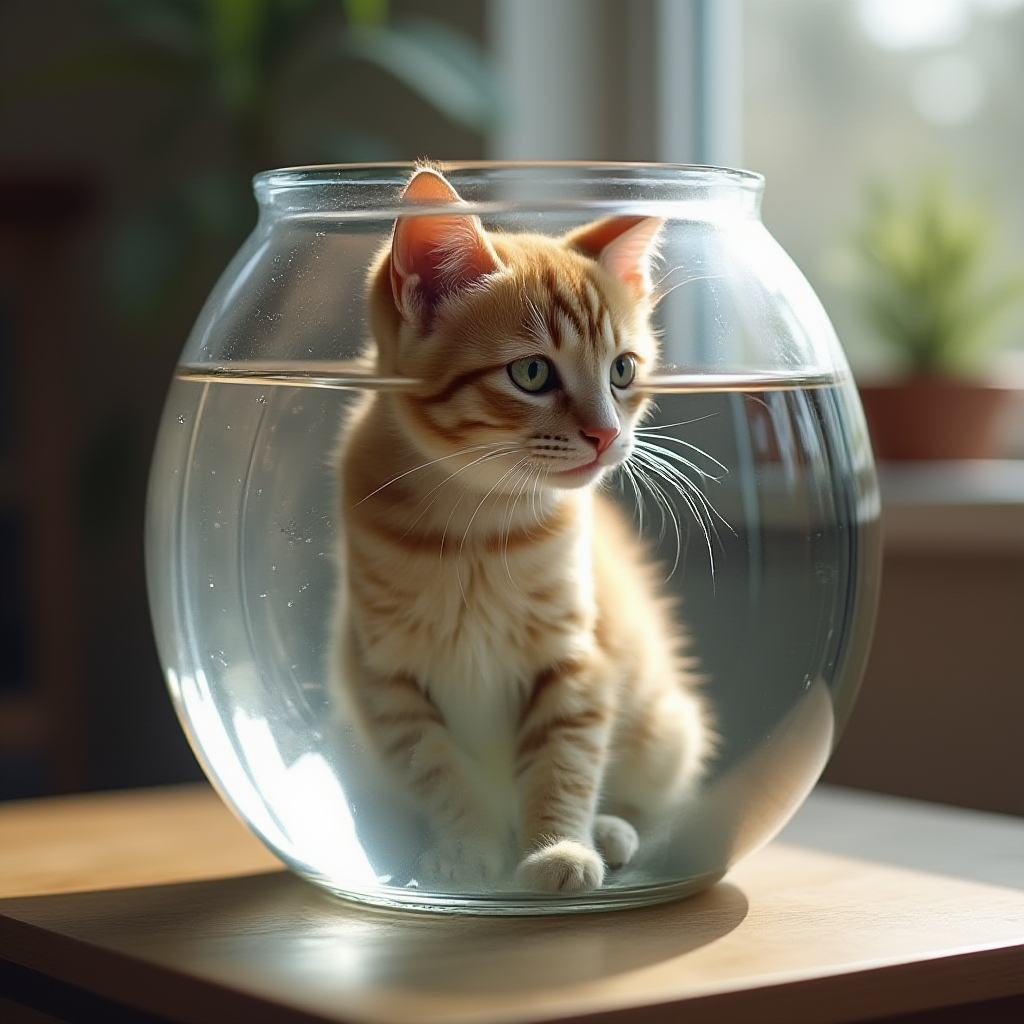} &
  \includegraphics[width=0.185\linewidth]{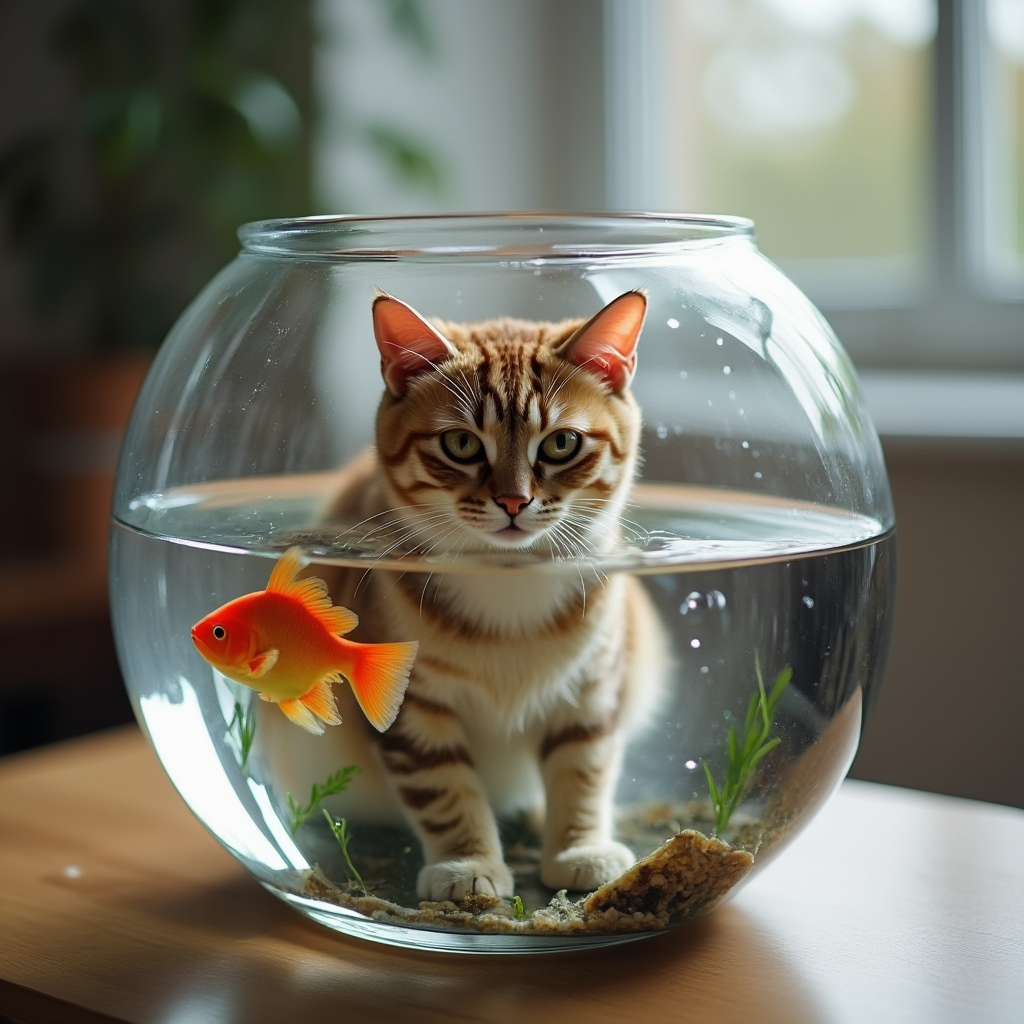} \\[1pt]
  \rotatebox{90}{\parbox{1.3cm}{\centering\tiny Exactly five balloons}} &
  \includegraphics[width=0.185\linewidth]{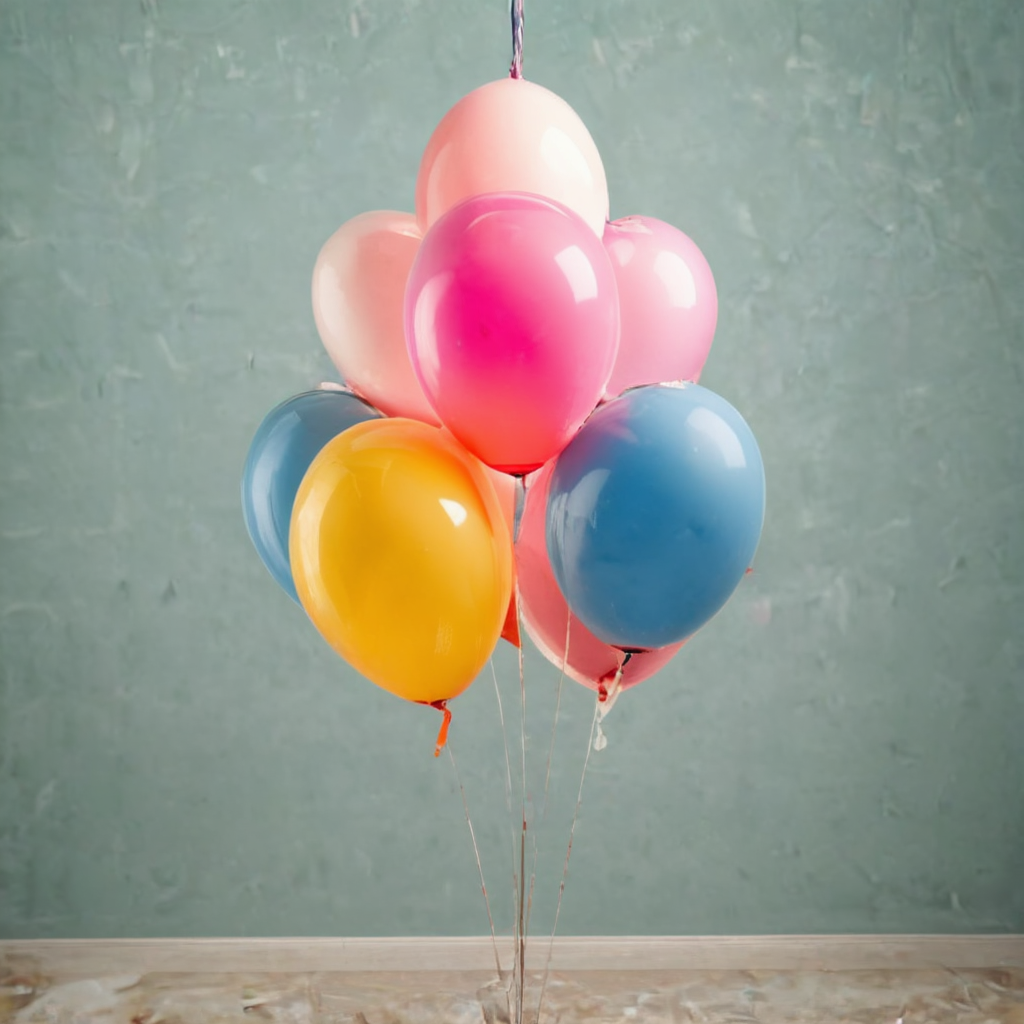} &
  \includegraphics[width=0.185\linewidth]{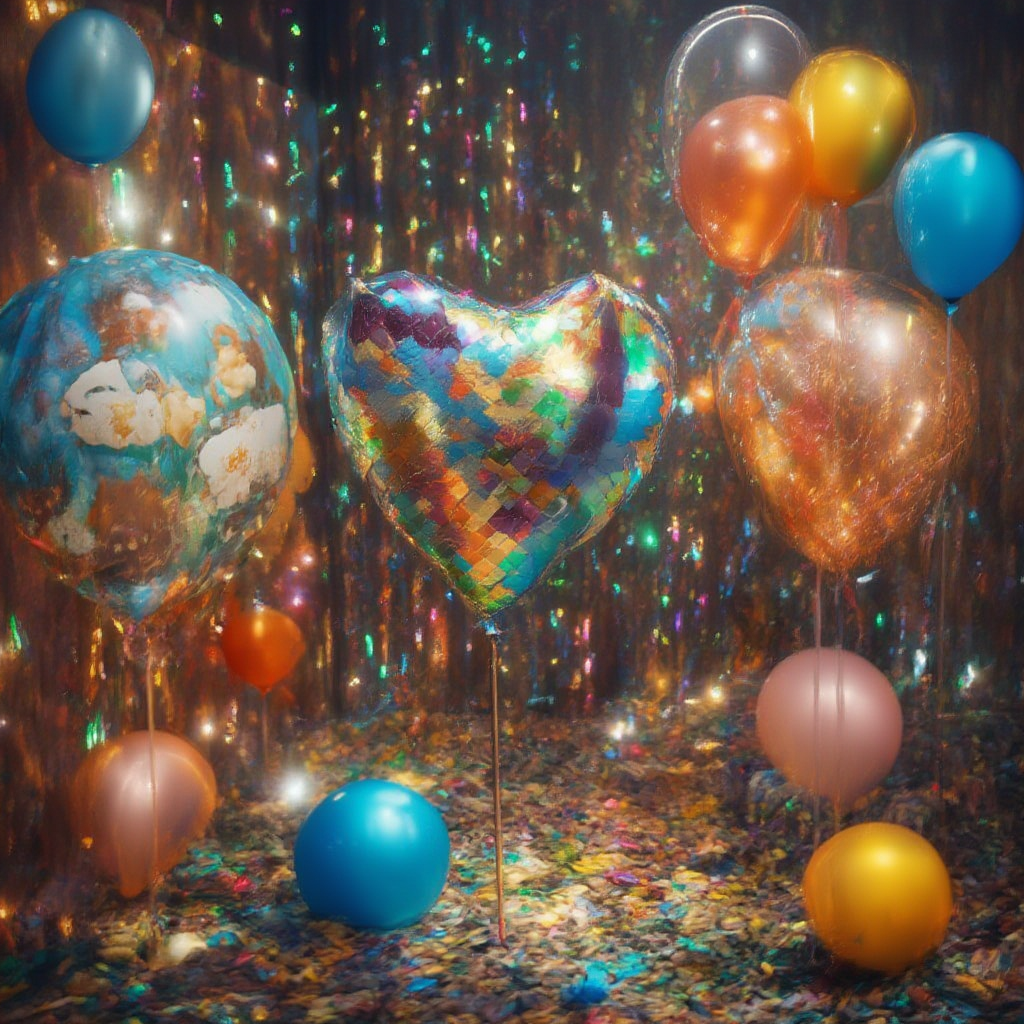} &
  \includegraphics[width=0.185\linewidth]{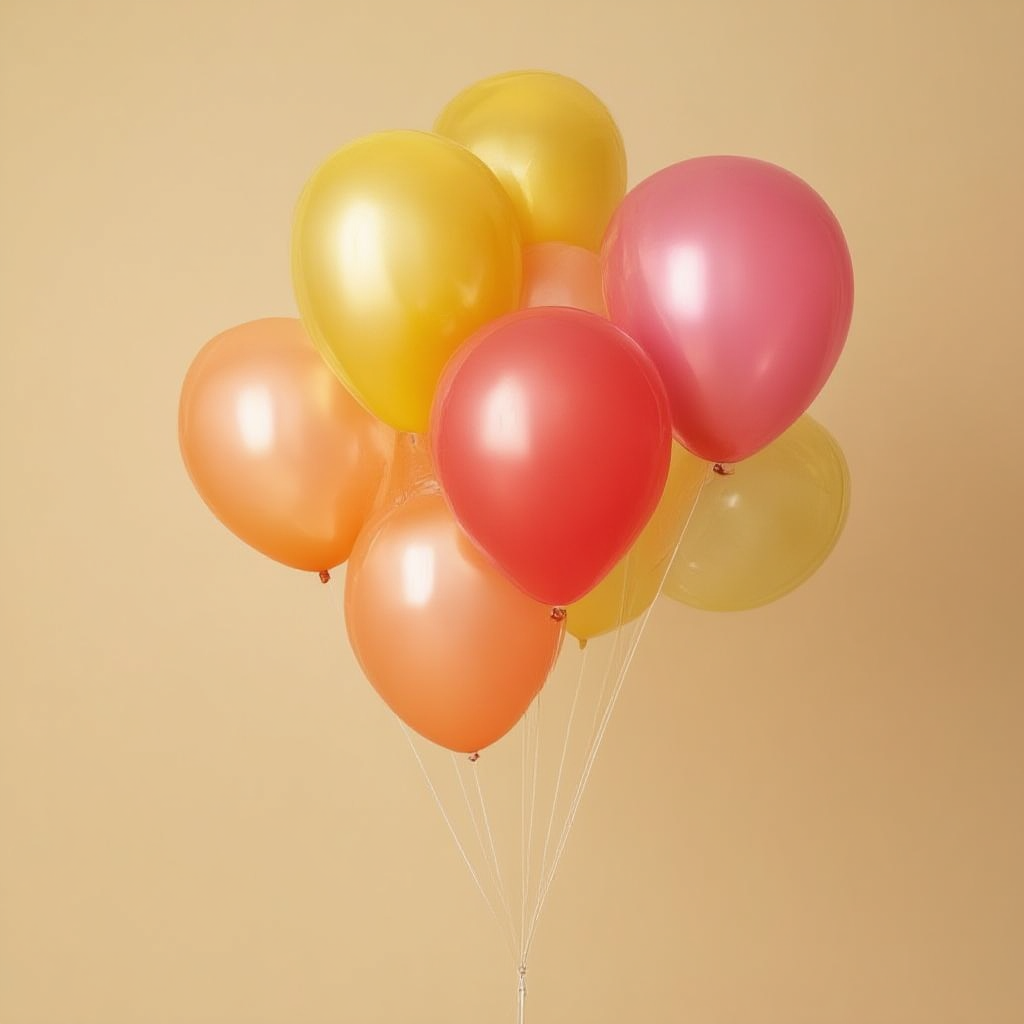} &
  \includegraphics[width=0.185\linewidth]{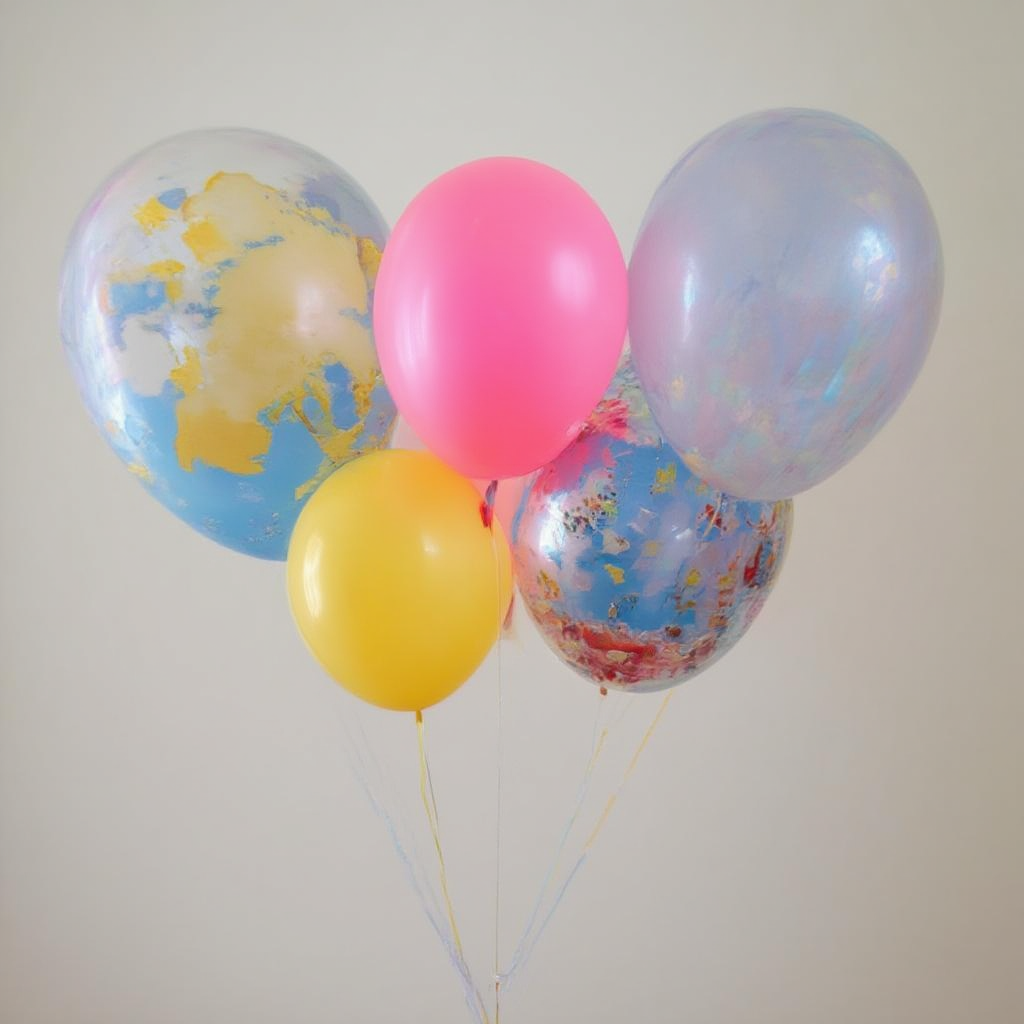} \\
  \end{tabular}
  % \caption{Text-to-image generation comparison. Full results in Appendix~\ref{app:t2i_case}.}
  \caption{Text-to-image generation comparison.}
  \label{fig:t2i_main}
  \vspace{-8pt}
\end{wrapfigure}
% \begin{wrapfigure}{r}{0.49\textwidth}
%   \vspace{-10pt}
%   \centering
%   \setlength{\tabcolsep}{1.5pt}
%   \begin{tabular}{@{}c@{\hspace{2pt}}c@{\hspace{2pt}}c@{\hspace{2pt}}c@{}}
%   & \textbf{\scriptsize w/o $\mathcal{L}_{\text{NFT}}$} & \textbf{\scriptsize w/o $r^{\text{vis}}$} & \textbf{\scriptsize BRAID} \\[2pt]
%   \rotatebox{90}{\parbox{1.5cm}{\centering\tiny A cat sitting inside a fishbowl}} &
%   \includegraphics[width=0.3\linewidth]{Figs/t2i_case/braid_wo_nft_1.png} &
%   \includegraphics[width=0.3\linewidth]{Figs/t2i_case/braid_wo_r_vis_case_1.png} &
%   \includegraphics[width=0.3\linewidth]{Figs/t2i_case/braid_case_1.png} \\[2pt]
%   \rotatebox{90}{\parbox{1.5cm}{\centering\tiny Exactly five balloons}} &
%   \includegraphics[width=0.3\linewidth]{Figs/t2i_case/braid_wo_nft_2.png} &
%   \includegraphics[width=0.3\linewidth]{Figs/t2i_case/braid_wo_r_vis_case_2.png} &
%   \includegraphics[width=0.3\linewidth]{Figs/t2i_case/braid_case_2.png} \\
%   \end{tabular}
%   \caption{Text-to-image generation of ablated variants. Full comparison is in Appendix~\ref{app:t2i_case}.}
%   \label{fig:t2i_main}
%   \vspace{-8pt}
% \end{wrapfigure}
\textbf{Interleaved Reasoning.}
We qualitatively compare BRAID and its ablations on a visual search task (Appendix~\ref{app:case}). BRAID generates diverse, high-fidelity intermediate images that directly support correct reasoning. In contrast, w/o $\mathcal{L}_{\text{NFT}}$ produces distorted visuals that induce hallucinated text reasoning, and w/o $r^{\text{vis}}$ introduces fabricated artifacts. Although all variants reach the correct answer on this instance, only BRAID's reasoning chain is grounded in faithful visual evidence, consistent with its superior robustness under Maj@$n$ sampling.

\textbf{Text-to-Image Generation.}
Figure~\ref{fig:t2i_main} shows whether BRAID's RL training transfers to standalone generation on demo prompts requiring spatial compositionality and numerical accuracy. Without $\mathcal{L}_{\text{NFT}}$, the image branch fails to follow instructions (cat placed beside the fishbowl; balloon count exceeds five). Removing $r^{\text{vis}}$ improves spatial relations but still lacks numerical precision. Only BRAID correctly satisfies both prompts, indicating that the vision-thinking reward provides optimization direction that generalizes beyond interleaved reasoning (full comparison in Appendix~\ref{app:t2i_case}).
% \section{Conclusions, Limitations, and Future Work}

\section{Conclusions, Limitations, and Future Work}  \label{sec:con}
We presented BRAID, a framework that casts interleaved text-image-text reasoning as a unified MDP, enabling joint optimization over both textual and visual generation within a unified policy.
By deriving a shared trajectory-level advantage and channeling it into modality-native policy gradients, BRAID propagates reward signals end-to-end across modality boundaries.
A complementary vision-thinking process reward, scored by a VLM judge, supplies turn-level feedback that sharpens credit assignment at critical image branches.
Experiments on seven spatial-reasoning and visual-perception benchmarks demonstrate that BRAID consistently outperforms various baselines.

Several limitations point to promising future directions.
First, BRAID is built upon BAGEL's hybrid AR-diffusion backbone; extending to fully autoregressive UMMs~\cite{wang2024emu3}, where discrete image tokens are generated via next-token prediction, would broaden the framework's applicability.
Second, the vision-thinking reward currently relies on an external VLM judge, inducing inference cost and dependency; distilling it into a compact, jointly trained PRM would improve both efficiency and reproducibility.
Finally, the current formulation assumes a fixed interleaving pattern; exploring adaptive turn structures and richer credit-assignment mechanisms, such as hierarchical advantage decomposition across turns, could further unlock RL's potential for interleaved multi-modal reasoning.

\onecolumn
\newpage
\bibliography{reference}

@article{team2024chameleon,
  title     = {Chameleon: Mixed-Modal Early-Fusion Foundation Models},
  author    = {{Chameleon Team}},
  journal   = {arXiv preprint arXiv:2405.09818},
  year      = {2024}
}

@article{wang2024emu3,
  title     = {Emu3: Next-Token Prediction is All You Need},
  author    = {Xinlong Wang and Xiaosong Zhang and Zhengxiong Luo and
               Quan Sun and Yufeng Cui and Jinsheng Wang and Fan Zhang and
               Yueze Wang and Zhen Li and Qiying Yu and others},
  journal   = {arXiv preprint arXiv:2409.18869},
  year      = {2024}
}

@article{cui2025emu3.5,
  title={Emu3.5: Native multimodal models are world learners},
  author={Cui, Yufeng and Chen, Honghao and Deng, Haoge and Huang, Xu and Li, Xinghang and Liu, Jirong and Liu, Yang and Luo, Zhuoyan and Wang, Jinsheng and Wang, Wenxuan and others},
  journal={arXiv preprint arXiv:2510.26583},
  year={2025}
}

@article{deng2025bagel,
  title     = {Emerging Properties in Unified Multimodal Pretraining},
  author    = {Chaorui Deng and Deyao Zhu and Kunchang Li and
               Chenhui Gou and Feng Li and Zeyu Wang and Shu Zhong and
               Weihao Yu and Xiaonan Nie and Ziang Song and Guang Shi and
               Haoqi Fan},
  journal   = {arXiv preprint arXiv:2505.14683},
  year      = {2025}
}

@article{wu2024janus,
  title     = {Janus: Decoupling Visual Encoding for Unified Multimodal
               Understanding and Generation},
  author    = {Chengyue Wu and Xiaokang Chen and Zhiyu Wu and
               Yiyang Ma and Xingchao Liu and Zizheng Pan and Wen Liu and
               Zhenda Xie and Xingkai Yu and Chong Ruan and Ping Luo},
  journal   = {arXiv preprint arXiv:2410.13848},
  year      = {2024}
}

@article{xie2024show,
  title={Show-o: One single transformer to unify multimodal understanding and generation},
  author={Xie, Jinheng and Mao, Weijia and Bai, Zechen and Zhang, David Junhao and Wang, Weihao and Lin, Kevin Qinghong and Gu, Yuchao and Chen, Zhijie and Yang, Zhenheng and Shou, Mike Zheng},
  journal={arXiv preprint arXiv:2408.12528},
  year={2024}
}

@article{zhou2024transfusion,
  title={Transfusion: Predict the next token and diffuse images with one multi-modal model},
  author={Zhou, Chunting and Yu, Lili and Babu, Arun and Tirumala, Kushal and Yasunaga, Michihiro and Shamis, Leonid and Kahn, Jacob and Ma, Xuezhe and Zettlemoyer, Luke and Levy, Omer},
  journal={arXiv preprint arXiv:2408.11039},
  year={2024}
}

@inproceedings{wei2022chain,
  title={Chain-of-thought prompting elicits reasoning in large language models},
  author={Wei, Jason and Wang, Xuezhi and Schuurmans, Dale and Bosma, Maarten and Xia, Fei and Chi, Ed and Le, Quoc V and Zhou, Denny and others},
  booktitle={Advances in Neural Information Processing Systems},
  volume={35},
  pages={24824--24837},
  year={2022}
}

@inproceedings{kojima2022large,
  title={Large language models are zero-shot reasoners},
  author={Kojima, Takeshi and Gu, Shixiang Shane and Reid, Machel and Matsuo, Yutaka and Iwasawa, Yusuke},
  booktitle={Advances in Neural Information Processing Systems},
  volume={35},
  pages={22199--22213},
  year={2022}
}

@article{hu2024visual,
  title={Visual sketchpad: Sketching as a visual chain of thought for multimodal language models},
  author={Hu, Yushi and Shi, Weijia and Fu, Xingyu and Roth, Dan and Ostendorf, Mari and Zettlemoyer, Luke and Smith, Noah A and Krishna, Ranjay},
  journal={Advances in Neural Information Processing Systems},
  volume={37},
  pages={139348--139379},
  year={2024}
}

@inproceedings{menon2024whiteboard,
  title={Whiteboard-of-thought: Thinking step-by-step across modalities},
  author={Menon, Sachit and Zemel, Richard and Vondrick, Carl},
  booktitle={Proceedings of the 2024 Conference on Empirical Methods in Natural Language Processing},
  pages={20016--20031},
  year={2024}
}

@inproceedings{tong2025metamorph,
  title={Metamorph: Multimodal understanding and generation via instruction tuning},
  author={Tong, Shengbang and Fan, David and Li, Jiachen and Xiong, Yunyang and Chen, Xinlei and Sinha, Koustuv and Rabbat, Michael and LeCun, Yann and Xie, Saining and Liu, Zhuang},
  booktitle={Proceedings of the IEEE/CVF International Conference on Computer Vision},
  pages={17001--17012},
  year={2025}
}

@article{li2025zebra,
  title={Zebra-cot: A dataset for interleaved vision language reasoning},
  author={Li, Ang and Wang, Charles and Fu, Deqing and Yue, Kaiyu and Cai, Zikui and Zhu, Wang Bill and Liu, Ollie and Guo, Peng and Neiswanger, Willie and Huang, Furong and others},
  journal={arXiv preprint arXiv:2507.16746},
  year={2025}
}

@article{li2025imagine,
  title={Imagine while reasoning in space: Multimodal visualization-of-thought},
  author={Li, Chengzu and Wu, Wenshan and Zhang, Huanyu and Xia, Yan and Mao, Shaoguang and Dong, Li and Vuli{\'c}, Ivan and Wei, Furu},
  journal={arXiv preprint arXiv:2501.07542},
  year={2025}
}

@article{meng2023chain,
  title={Chain of images for intuitively reasoning},
  author={Meng, Fanxu and Yang, Haotong and Wang, Yiding and Zhang, Muhan},
  journal={arXiv preprint arXiv:2311.09241},
  year={2023}
}

@article{gu2025thinkmorph,
  title     = {ThinkMorph: Emergent Properties in Multimodal Interleaved
               Chain-of-Thought Reasoning},
  author    = {Jiawei Gu and Yunzhuo Hao and Huichen Will Wang and
               Linjie Li and Michael Qizhe Shieh and Yejin Choi and
               Ranjay Krishna and Yu Cheng},
  journal   = {arXiv preprint arXiv:2510.27492},
  year      = {2025}
}

@article{chern2025twgi,
  title={Thinking with generated images},
  author={Chern, Ethan and Hu, Zhulin and Chern, Steffi and Kou, Siqi and Su, Jiadi and Ma, Yan and Deng, Zhijie and Liu, Pengfei},
  journal={arXiv preprint arXiv:2505.22525},
  year={2025}
}

@article{gao2024icot,
  title     = {Interleaved-Modal Chain-of-Thought},
  author    = {Jun Gao and Yongqi Li and Ziqiang Cao and Wenjie Li},
  journal   = {arXiv preprint arXiv:2411.19488},
  year      = {2024}
}

@article{wu2026visworld,
  title     = {Visual Generation Unlocks Human-Like Reasoning through
               Multimodal World Models},
  author    = {Jialong Wu and Xiaoying Zhang and Hongyi Yuan and
               Xiangcheng Zhang and Tianhao Huang and Changjing He and
               Chaoyi Deng and Renrui Zhang and Youbin Wu and
               Mingsheng Long},
  journal   = {arXiv preprint arXiv:2601.19834},
  year      = {2026}
}

@article{guo2025deepseekr1,
  title     = {DeepSeek-R1: Incentivizing Reasoning Capability in {LLM}s
               via Reinforcement Learning},
  author    = {Daya Guo and Dejian Yang and Haowei Zhang and
               Junxiao Song and Ruoyu Zhang and Runxin Xu and
               Qihao Zhu and Shirong Ma and Peiyi Wang and Xiao Bi
               and others},
  journal   = {arXiv preprint arXiv:2501.12948},
  year      = {2025}
}

@article{shao2024deepseekmath,
  title     = {DeepSeekMath: Pushing the Limits of Mathematical Reasoning
               in Open Language Models},
  author    = {Zhihong Shao and Peiyi Wang and Qihao Zhu and Runxin Xu and
               Junxiao Song and Xiao Bi and Haowei Zhang and
               Mingchuan Zhang and Y. K. Li and Y. Wu and others},
  journal   = {arXiv preprint arXiv:2402.03300},
  year      = {2024}
}

@article{meng2025mmeureka,
  title     = {MM-Eureka: Exploring the Frontiers of Multimodal Reasoning
               with Rule-based Reinforcement Learning},
  author    = {Fanqing Meng and Lingxiao Du and Zongkai Liu and
               Zhixiang Zhou and Quanfeng Lu and Daocheng Fu and
               Tiancheng Han and Botian Shi and Wenhai Wang and
               Junjun He and Kaipeng Zhang and Ping Luo and Yu Qiao and
               Qiaosheng Zhang and Wenqi Shao},
  journal   = {arXiv preprint arXiv:2503.07365},
  year      = {2025}
}

@article{yang2025r1onevision,
  title     = {R1-Onevision: Advancing Generalized Multimodal Reasoning
               through Cross-Modal Formalization},
  author    = {Yi Yang and Xiaohui He and Haotian Pan and Xiaojiang Jiang and
               Yao Deng and Xiaoyang Yang and Hao Lu and Dawei Yin and
               Fei Rao and Miao Zhu and others},
  journal   = {arXiv preprint arXiv:2503.10615},
  year      = {2025}
}

@article{huang2025vision,
  title={Vision-r1: Incentivizing reasoning capability in multimodal large language models},
  author={Huang, Wenxuan and Jia, Bohan and Zhai, Zijie and Cao, Shaosheng and Ye, Zheyu and Zhao, Fei and Xu, Zhe and Tang, Xu and Hu, Yao and Lin, Shaohui},
  journal={arXiv preprint arXiv:2503.06749},
  year={2025}
}

@article{zheng2025deepeyes,
  title     = {DeepEyes: Incentivizing ``Thinking with Images'' via
               Reinforcement Learning},
  author    = {Ziwei Zheng and Michael Yang and Jack Hong and
               Chenxiao Zhao and Guohai Xu and Le Yang and Chao Shen and
               Xing Yu},
  journal   = {arXiv preprint arXiv:2505.14362},
  year      = {2025}
}

@article{zhang2025cooper,
  title     = {{COOPER}: A Unified Model for Cooperative Perception and
               Reasoning in Spatial Intelligence},
  author    = {Zefeng Zhang and Xiangzhao Hao and Hengzhu Tang and
               Zhenyu Zhang and Jiawei Sheng and Xiaodong Li and
               Zhenyang Li and Li Gao and Daiting Shi and Dawei Yin and
               Tingwen Liu},
  journal   = {arXiv preprint arXiv:2512.04563},
  year      = {2025}
}

@article{geng2025xomni,
  title={X-omni: Reinforcement learning makes discrete autoregressive image generative models great again},
  author={Geng, Zigang and Wang, Yibing and Ma, Yeyao and Li, Chen and Rao, Yongming and Gu, Shuyang and Zhong, Zhao and Lu, Qinglin and Hu, Han and Zhang, Xiaosong and others},
  journal={arXiv preprint arXiv:2507.22058},
  year={2025}
}

@article{lipman2022flow,
  title={Flow matching for generative modeling},
  author={Lipman, Yaron and Chen, Ricky TQ and Ben-Hamu, Heli and Nickel, Maximilian and Le, Matt},
  journal={arXiv preprint arXiv:2210.02747},
  year={2022}
}

@article{black2023ddpo,
  title     = {Training Diffusion Models with Reinforcement Learning},
  author    = {Kevin Black and Michael Janner and Yilun Du and
               Ilya Kostrikov and Sergey Levine},
  journal   = {arXiv preprint arXiv:2305.13301},
  year      = {2023}
}

@article{fan2023dpok,
  title={Dpok: Reinforcement learning for fine-tuning text-to-image diffusion models},
  author={Fan, Ying and Watkins, Olivia and Du, Yuqing and Liu, Hao and Ryu, Moonkyung and Boutilier, Craig and Abbeel, Pieter and Ghavamzadeh, Mohammad and Lee, Kangwook and Lee, Kimin},
  journal={Advances in Neural Information Processing Systems},
  volume={36},
  pages={79858--79885},
  year={2023}
}

@article{liu2025flowgrpo,
  title     = {Flow-GRPO: Training Flow Matching Models via Online {RL}},
  author    = {Jie Liu and Gongye Liu and Jiajun Liang and Yangguang Li and
               Jiaheng Liu and Xintao Wang and Pengfei Wan and Di Zhang and
               Wanli Ouyang},
  journal   = {arXiv preprint arXiv:2505.05470},
  year      = {2025}
}

@article{zheng2025diffusionnft,
  title     = {DiffusionNFT: Online Diffusion Reinforcement with Forward
               Process},
  author    = {Kaiwen Zheng and Huayu Chen and Haotian Ye and
               Haoxiang Wang and Qinsheng Zhang and Kai Jiang and
               Hang Su and Stefano Ermon and Jun Zhu and Ming-Yu Liu},
  journal   = {arXiv preprint arXiv:2509.16117},
  year      = {2025}
}

@article{xue2025dancegrpo,
  title={Dancegrpo: Unleashing grpo on visual generation},
  author={Xue, Zeyue and Wu, Jie and Gao, Yu and Kong, Fangyuan and Zhu, Lingting and Chen, Mengzhao and Liu, Zhiheng and Liu, Wei and Guo, Qiushan and Huang, Weilin and others},
  journal={arXiv preprint arXiv:2505.07818},
  year={2025}
}

@article{liu2026unigrpo,
  title     = {{UniGRPO}: Unified Policy Optimization for Reasoning-Driven
               Visual Generation},
  author    = {Jie Liu and Zilyu Ye and Linxiao Yuan and Shenhan Zhu and
               Yu Gao and Jie Wu and Kunchang Li and Xionghui Wang and
               Xiaonan Nie and Weilin Huang and Wanli Ouyang},
  journal   = {arXiv preprint arXiv:2603.23500},
  year      = {2026}
}

@article{chang2017matterport3d,
  title={Matterport3d: Learning from rgb-d data in indoor environments},
  author={Chang, Angel and Dai, Angela and Funkhouser, Thomas and Halber, Maciej and Niessner, Matthias and Savva, Manolis and Song, Shuran and Zeng, Andy and Zhang, Yinda},
  journal={arXiv preprint arXiv:1709.06158},
  year={2017}
}

@misc{Li2025realsee3d,
  doi = {10.5281/zenodo.17826243},
  url = {https://doi.org/10.5281/zenodo.17826243},
  author = {Li, Linyuan and Wu, Yan and Li, Xi and Wang, Lingli and Rao, Tong and Zhou, Jie and Pan, Cihui and Hui, Xinchen},
  title = {Realsee3D: A Large-Scale Multi-View RGB-D Dataset of Indoor Scenes (Version 1.0)},
  publisher = {Zenodo},
  year = {2025}
}

@inproceedings{chen2024360+,
  title={360+ x: A panoptic multi-modal scene understanding dataset},
  author={Chen, Hao and Hou, Yuqi and Qu, Chenyuan and Testini, Irene and Hong, Xiaohan and Jiao, Jianbo},
  booktitle={Proceedings of the IEEE/CVF Conference on Computer Vision and Pattern Recognition},
  pages={19373--19382},
  year={2024}
}

@article{shao2024visual,
  title={Visual cot: Unleashing chain-of-thought reasoning in multi-modal language models},
  author={Shao, Hao and Qian, Shengju and Xiao, Han and Song, Guanglu and Zong, Zhuofan and Wang, Letian and Liu, Yu and Li, Hongsheng},
  journal={arXiv preprint arXiv:2403.16999},
  volume={2},
  year={2024}
}

@article{wang2025jigsaw,
  title={Jigsaw-r1: A study of rule-based visual reinforcement learning with jigsaw puzzles},
  author={Wang, Zifu and Zhu, Junyi and Tang, Bo and Li, Zhiyu and Xiong, Feiyu and Yu, Jiaqian and Blaschko, Matthew B},
  journal={arXiv preprint arXiv:2505.23590},
  year={2025}
}

@inproceedings{vstar,
  title={V*: Guided Visual Search as a Core Mechanism in Multimodal LLMs},
  author={Wu, Penghao and Xie, Saining},
  booktitle={Proceedings of the IEEE/CVF Conference on Computer Vision and Pattern Recognition},
  pages={13084--13094},
  year={2024}
}

@inproceedings{mmvp,
  title={Eyes wide shut? exploring the visual shortcomings of multimodal llms},
  author={Tong, Shengbang and Liu, Zhuang and Zhai, Yuexiang and Ma, Yi and LeCun, Yann and Xie, Saining},
  booktitle={Proceedings of the IEEE/CVF conference on computer vision and pattern recognition},
  pages={9568--9578},
  year={2024}
}

@inproceedings{blink,
  title={Blink: Multimodal large language models can see but not perceive},
  author={Fu, Xingyu and Hu, Yushi and Li, Bangzheng and Feng, Yu and Wang, Haoyu and Lin, Xudong and Roth, Dan and Smith, Noah A and Ma, Wei-Chiu and Krishna, Ranjay},
  booktitle={European Conference on Computer Vision},
  pages={148--166},
  year={2024},
  organization={Springer}
}

@article{mmsi,
  title={Mmsi-bench: A benchmark for multi-image spatial intelligence},
  author={Yang, Sihan and Xu, Runsen and Xie, Yiman and Yang, Sizhe and Li, Mo and Lin, Jingli and Zhu, Chenming and Chen, Xiaochen and Duan, Haodong and Yue, Xiangyu and others},
  journal={arXiv preprint arXiv:2505.23764},
  year={2025}
}

@article{sat,
  title={Sat: Spatial aptitude training for multimodal language models},
  author={Ray, Arijit and Duan, Jiafei and Tan, Reuben and Bashkirova, Dina and Hendrix, Rose and Ehsani, Kiana and Kembhavi, Aniruddha and Plummer, Bryan A and Krishna, Ranjay and Zeng, Kuo-Hao and others},
  journal={arXiv preprint arXiv:2412.07755},
  volume={3},
  year={2024}
}

@article{cvbench,
  title={Cambrian-1: A fully open, vision-centric exploration of multimodal llms},
  author={Tong, Shengbang and Brown, Ellis and Wu, Penghao and Woo, Sanghyun and Middepogu, Manoj and Akula, Sai C and Yang, Jihan and Yang, Shusheng and Iyer, Adithya and Pan, Xichen and others},
  journal={Advances in Neural Information Processing Systems},
  volume={37},
  pages={87310--87356},
  year={2024}
}

@inproceedings{vlmevalkit,
  title={Vlmevalkit: An open-source toolkit for evaluating large multi-modality models},
  author={Duan, Haodong and Yang, Junming and Qiao, Yuxuan and Fang, Xinyu and Chen, Lin and Liu, Yuan and Dong, Xiaoyi and Zang, Yuhang and Zhang, Pan and Wang, Jiaqi and others},
  booktitle={Proceedings of the 32nd ACM international conference on multimedia},
  pages={11198--11201},
  year={2024}
}

@article{gpt4o,
  title={Gpt-4o system card},
  author={Hurst, Aaron and Lerer, Adam and Goucher, Adam P and Perelman, Adam and Ramesh, Aditya and Clark, Aidan and Ostrow, AJ and Welihinda, Akila and Hayes, Alan and Radford, Alec and others},
  journal={arXiv preprint arXiv:2410.21276},
  year={2024}
}

@misc{qwen25vl,
      title={Qwen2.5-VL Technical Report}, 
      author={Shuai Bai and Keqin Chen and Xuejing Liu and Jialin Wang and Wenbin Ge and Sibo Song and Kai Dang and Peng Wang and Shijie Wang and Jun Tang and Humen Zhong and Yuanzhi Zhu and Mingkun Yang and Zhaohai Li and Jianqiang Wan and Pengfei Wang and Wei Ding and Zheren Fu and Yiheng Xu and Jiabo Ye and Xi Zhang and Tianbao Xie and Zesen Cheng and Hang Zhang and Zhibo Yang and Haiyang Xu and Junyang Lin},
      year={2025},
      eprint={2502.13923},
      archivePrefix={arXiv},
      primaryClass={cs.CV},
      url={https://arxiv.org/abs/2502.13923}, 
}

@inproceedings{cheng2025visual,
  title={Visual Thoughts: A Unified Perspective of Understanding Multimodal Chain-of-Thought},
  author={Cheng, Zihui and Chen, Qiguang and Xu, Xiao and WANG, Jiaqi and Wang, Weiyun and Fei, Hao and Wang, Yidong and Wang, Alex Jinpeng and Chen, Zhi and Che, Wanxiang and others},
  booktitle={Advances in Neural Information Processing Systems},
  year={2025}
}

@article{liu2026beyond,
  title={Beyond the {Dirac Delta}: Mitigating Diversity Collapse in Reinforcement Fine-Tuning for Versatile Image Generation},
  author={Liu, Jinmei and Li, Haoru and Sun, Zhenhong and Chen, Chaofeng and Bian, Yatao and Wang, Bo and Dong, Daoyi and Chen, Chunlin and Wang, Zhi},
  journal={arXiv preprint arXiv:2601.12401},
  year={2026}
}

@inproceedings{dong2024dreamllm,
  title={{DreamLLM}: Synergistic Multimodal Comprehension and Creation},
  author={Dong, Runpei and Han, Chunrui and Peng, Yuang and others},
  booktitle={International Conference on Learning Representations},
  year={2024}
}

@inproceedings{nie2025towards,
  title={Towards Unified Multimodal Interleaved Generation via Group Relative Policy Optimization},
  author={Ming Nie and Chunwei Wang and Jianhua Han and Hang Xu and Li Zhang},
  booktitle={Advances in Neural Information Processing Systems},
  year={2025},
}
\bibliographystyle{unsrt}

%%%%%%%%%%%%%%%%%%%%%%%%%%%%%%%%%%%%%%%%%%%%%%%%%%%%%%%%%%%%

\clearpage
\appendix
\part*{Appendix}
\addcontentsline{toc}{part}{Appendix}  % Add to main TOC
\startcontents[appendix]
\printcontents[appendix]{}{1}{}

% \section{Limitaions}

\section{Experimental Details}\label{app:exp}

\subsection{Data Configuration} \label{app:data}
Following ThinkMorph and its follow-ups~\citep{gu2025thinkmorph}, our training corpus spans eight task families (\texttt{Matterport3D}, \texttt{RealSee3D}, \texttt{Perspective}, \texttt{DirectionalQuery}, \texttt{$360{+}x$}, \texttt{PositionedDirection}, \texttt{Jigsaw}, \texttt{VisualSearch}), each contributing both SFT and RL split. To study how the SFT/RL data budget interacts with our two-stage pipeline, we define an \emph{overlap ratio} $\omega \triangleq |\mathcal{D}_{\text{SFT}}\cap\mathcal{D}_{\text{RL}}|/|\mathcal{D}_{\text{RL}}| \in [0,1]$, and consider three data-allocation regimes (per-task counts in Table~\ref{tab:data_config}): \textbf{(1)} a \emph{disjoint} regime ($\mathcal{D}^{2{:}1}$, $\omega{=}0$) with a $2{:}1$ SFT/RL budget, favoring supervised initialization; \textbf{(2)} a \emph{disjoint} regime ($\mathcal{D}^{1{:}1}$, $\omega{=}0$) with a $1{:}1$ budget, isolating the effect of equal-sized splits; and \textbf{(3)} a \emph{replay} regime ($\mathcal{D}^{\cap}$) where the RL pool augments the original RL split with additional samples drawn from the SFT pool ($20{,}000$/$15{,}200$, $\omega{\approx}0.32$), so that part of the RL prompts revisit SFT examples under on-policy rollouts. We adopt regime~\textbf{(3)} as our default.
\begin{table}[ht]
\centering
\small
\caption{Per-task sample counts under the three data-allocation regimes. In $\mathcal{D}^{\cap}$, the RL pool extends the original RL split with $4{,}800$ samples drawn from the SFT pool, yielding an overlap ratio $\omega\approx 0.32$.}
\setlength{\tabcolsep}{4pt}
\renewcommand{\arraystretch}{1.2}
\begin{tabular}{l|cc|cc|cc}
\toprule
 & \multicolumn{2}{c|}{$\mathcal{D}^{2{:}1}$ ($\omega{=}0$)} & \multicolumn{2}{c|}{$\mathcal{D}^{1{:}1}$ ($\omega{=}0$)} & \multicolumn{2}{c}{$\mathcal{D}^{\cap}$ ($\omega{\approx}0.32$)} \\
\textbf{Task} & SFT & RL & SFT & RL & SFT & RL \\
\midrule
\texttt{Matterport3D}        & 4{,}000 & 2{,}000 & 3{,}000 & 3{,}000 & 4{,}000 & 3{,}000 \\
\texttt{RealSee3D}           & 4{,}000 & 2{,}000 & 3{,}000 & 3{,}000 & 4{,}000 & 3{,}000 \\
\texttt{Perspective}         & 4{,}000 & 2{,}000 & 3{,}000 & 3{,}000 & 4{,}000 & 3{,}000 \\
\texttt{DirectionalQuery}    & 1{,}400 & 1{,}000 & 1{,}200 & 1{,}200 & 1{,}400 & 1{,}200 \\
\texttt{$360{+}x$}           &   400   &   200   &   300   &   300   &   400   &   300   \\
\texttt{PositionedDirection} &   200   &   200   &   200   &   200   &   200   &   200   \\
\texttt{Jigsaw}              & 2{,}000 & 1{,}000 & 2{,}000 & 2{,}000 & 2{,}000 & 2{,}000 \\
\texttt{VisualSearch}        & 4{,}000 & 2{,}000 & 2{,}500 & 2{,}500 & 4{,}000 & 2{,}500 \\
\midrule
\textbf{Total}               & \textbf{20{,}000} & \textbf{10{,}400} & \textbf{15{,}200} & \textbf{15{,}200} & \textbf{20{,}000} & \textbf{15{,}200} \\
\bottomrule
\end{tabular}
\label{tab:data_config}
\end{table}

\subsection{Evaluation Benchmarks} \label{app:benchmark}
We evaluate on seven benchmarks spanning two categories. All evaluations are conducted under the VLMEvalKit~\citep{vlmevalkit} framework with accuracy as the primary metric.

\paragraph{Spatial Reasoning.}
(1)~\textbf{MMSI-Bench}~\citep{mmsi} is a multi-image spatial intelligence benchmark comprising 1,000 multiple-choice questions across 10 fundamental spatial tasks (e.g., object motion tracking, camera ego-motion, scene reconstruction), curated from over 120,000 images by 3D-vision experts.
(2)~\textbf{SAT}~\citep{sat} evaluates dynamic and static spatial reasoning, including egocentric movement, allocentric perspective shifts, depth relations, and action consequence prediction. It contains over 218,000 QA pairs across 22,000 synthetic scenes generated with the ProcTHOR engine, plus 150 real-image QA pairs for out-of-distribution testing.
(3)~\textbf{MMVP}~\citep{mmvp} targets systematic failures of vision encoders by constructing ``CLIP-blind pairs''---image pairs that CLIP encodes similarly despite clear visual differences in orientation, counting, color, or viewpoint. It contains approximately 150 image pairs formulated as multiple-choice questions.
(4)~\textbf{CV-Bench 3D}~\citep{cvbench} is the 3D split of the Cambrian Vision-Centric Benchmark, evaluating depth ordering and relative distance understanding via natural language questions derived from OMNI3D annotations.

\paragraph{Visual Perception.}
(1)~\textbf{BLINK}~\citep{blink} contains 3,807 multiple-choice questions reformulated from 14 classic computer vision tasks (relative depth, visual correspondence, multi-view reasoning, forensic detection, spatial relations, etc.) that humans can solve ``in a blink'' but multimodal LLMs find challenging.
(2)~\textbf{V$^{*}$Bench}~\citep{vstar} evaluates fine-grained visual perception in high-resolution, visually crowded images, comprising 191 multiple-choice questions on attribute recognition (color, material) and spatial relationship reasoning for small or occluded objects that require targeted visual search.
(3)~\textbf{CV-Bench 2D}~\citep{cvbench} is the 2D split of the Cambrian Vision-Centric Benchmark, assessing spatial relationship understanding and object counting via questions derived from ADE20k and COCO annotations.

\subsection{Training Recipe}
\begin{table}[ht]
\centering
\caption{Key hyperparameters for RL training.}
\label{tab:training_recipe}
\setlength{\tabcolsep}{6pt}
\renewcommand{\arraystretch}{1.15}
\begin{tabular}{lrlr}
\toprule
\textbf{Hyperparameter} & \textbf{Value} & \textbf{Hyperparameter} & \textbf{Value} \\
\midrule
Rollout batch size & 64 & NFT $\beta$ & 0.4 \\
Update batch size & 32 & NFT training timesteps & 1 \\
Rollouts per prompt & 8 & Denoising steps & 20 \\
PPO epochs & 1 & Timestep shift & 3.0 \\
Training steps & 120 & CFG text scale & 4.0 \\
Clip ratio (high / low) & 0.28 / 0.20 & CFG image scale & 2.0 \\
Entropy coefficient & 0.0 & CFG interval & $[0.4, 1.0]$ \\
KL loss & disabled & Noise level & 0.7 \\
Sampling temperature & 1.0 & Max latent size & 64 \\
Max interleaved rounds & 3 & VLM judge & GPT-5.2 \\
Max thinking tokens & 2048 & Reward weights & 0.25 each \\
Max total tokens & 32768 & & \\
\bottomrule
\end{tabular}
\end{table}
We build on BAGEL-7B~\citep{deng2025bagel} and train with a two-stage pipeline: SFT followed by RL. Both stages use FSDP2 with hybrid sharding, gradient checkpointing, and the VAE frozen throughout. During RL, all other modules (LLM, ViT, understanding head, generation head) are unfrozen and jointly optimized.

For the SFT stage, we follow ThinkMorph~\citep{gu2025thinkmorph} and train on curated TIT-CoT trajectories under the $\mathcal{D}^{2:1}$ data regime (detailed in Section~\ref{app:data}). For the RL stage, we adopt the replay regime $\mathcal{D}^{\cap}$ ($\omega{\approx}0.32$) as default, where the RL pool partially overlaps with SFT data to bridge imitation learning and on-policy exploration. We sample rollouts with temperature 1.0 and allow up to 3 interleaved image rounds per trajectory, with a maximum of 2048 thinking tokens per round and 32768 total tokens. Image generation uses 20 denoising steps (for both training and evaluation) with a timestep shift of 3.0, classifier-free guidance scales of 4.0 (text) and 2.0 (image), and a guidance interval of $[0.4, 1.0]$. The noise level for intermediate images is set to 0.7. Table~\ref{tab:training_recipe} summarizes the key hyperparameters.

\subsection{Vision-Thinking Process Reward Prompt}
\label{app:prompt}

To instantiate the per-turn visual process reward $\bm{r}^{\text{vis}}$, we employ \texttt{GPT-5.2} as the VLM judge. Given the question, the problem image(s), and the generated intermediate image $I_k$, the judge follows a three-step protocol (question understanding $\rightarrow$ image analysis $\rightarrow$ scoring) and emits four integer scores in $[1,10]$, corresponding one-to-one to the four criteria $\{r^{(\text{vc})}, r^{(\text{vf})}, r^{(\text{ru})}, r^{(\text{tr})}\}$ defined in Sec.~\ref{subsec:reward}. The full system prompt is shown below.

\begin{tcolorbox}[
  colback=lightgray!30,
  colframe=black,
  width=\textwidth,
  arc=0mm,
  boxrule=0.5pt,
]
You are an expert evaluator for a vision-language model (VLM) that reasons by generating intermediate images as visual tools. Your task: evaluate one generated intermediate image within a multi-step reasoning trajectory.\\[4pt]

\textbf{STEP 1 --- UNDERSTAND THE QUESTION \& PROBLEM IMAGE(S)} \\
Read the question carefully. Identify:
\begin{itemize}\itemsep0pt
  \item What specific visual information is needed to answer it?
  \item What is provided by the problem image(s), and what remains ambiguous?
  \item What type of visual operation would best help answer this question?
\end{itemize}

\textbf{STEP 2 --- ANALYZE THE GENERATED IMAGE} \\
Describe concretely what it shows. Compare it carefully against the problem image(s) --- what is preserved, what is changed, what is new, what might be fabricated? Be very precise: zoom in on small details, distinguish between nearby objects, and verify that what the model claims to target is actually what the generated image shows.\\[4pt]

\textbf{STEP 3 --- SCORE on the 4 dimensions below (1--10 each, integers only).}
\begin{enumerate}\itemsep2pt
  \item \texttt{action\_correctness} --- Did the model choose the RIGHT visual operation and execute it correctly? 10 = perfect execution. 1 = completely wrong.
  \item \texttt{visual\_faithfulness} --- Does the generated image faithfully represent the actual content of the problem image(s)? Only penalize fabrications that could mislead reasoning. 10 = pixel-accurate. 1 = entirely fabricated.
  \item \texttt{reasoning\_contribution} --- Does this image provide NEW, CORRECT information that could advance reasoning toward the correct answer? 10 = decisive new evidence. 1 = useless or actively misleading.
  \item \texttt{misleading\_risk} --- Could this image cause a reasonable model to reach a **wrong** answer? 10 = completely safe. 1 = virtually guarantees wrong answer.
\end{enumerate}

Be willing to use the full 1--10 range. A score of 8+ means genuinely good.\\[4pt]

Return **only** valid **json** (no markdown, no extra text):\\
\texttt{\{"action\_correctness": <int>, "visual\_faithfulness": <int>, "reasoning\_contribution": <int>, "misleading\_risk": <int>\}}\\[6pt]
\end{tcolorbox}

\clearpage
\section{Additional Results}\label{app:more_analysis}

\subsection{Ablation Study}
\begin{table}[ht]
\centering
\caption{Ablation study on different components of BRAID across seven spatial reasoning and perception benchmarks. Best results in \textbf{bold} and second best \underline{underlined}.}
\label{tab:ablation}
\setlength{\tabcolsep}{5pt}
\renewcommand{\arraystretch}{1.3}
\resizebox{\textwidth}{!}{%
\begin{tabular}{l cccc ccc >{\columncolor{cyan!10}}c}
\toprule
\multirow{2}{*}{\textbf{Model}}
 & \multicolumn{4}{c}{\textbf{Spatial Reasoning}}
 & \multicolumn{3}{c}{\textbf{Perception}}
 & \multirow{2}{*}{\textbf{Avg.}} \\
\cmidrule(lr){2-5} \cmidrule(lr){6-8}
 & \textbf{MMSI} & \textbf{SAT} & \textbf{MMVP} & \textbf{CV-Bench 3D}
 & \textbf{BLINK} & \textbf{VStar} & \textbf{CV-Bench 2D} & \\
\midrule
BAGEL                                & 26.60 & 44.67 & 75.67 & \textbf{83.41} & 54.76 & 57.59 & 73.53 & 59.46 \\
SFT                                  & \underline{29.20} & 49.67 & 78.67 & 81.67 & \underline{57.02} & 62.82 & \underline{74.85} & 61.99 \\
\midrule
BRAID w/o $r^{\text{vis}}$           & 28.40 & \underline{56.00} & \underline{79.00} & \underline{82.17} & 55.81 & \underline{63.87} & 73.48 & \underline{62.68} \\
BRAID w/o $\mathcal{L}_{\text{NFT}}$ & 28.90 & 54.33 & \underline{79.00} & 81.50 & 56.70 & 62.35 & 71.94 & 62.10 \\
\midrule
BRAID                                & \textbf{31.80} & \textbf{58.67} & \textbf{81.33} & \underline{82.17} & \textbf{57.76} & \textbf{68.35} & \textbf{76.26} & \textbf{65.19} \\
\bottomrule
\end{tabular}%
}
\end{table}
Table~\ref{tab:ablation} reports the complete numerical results of the ablation study discussed in Section~\ref{subsec:ablation}, covering all seven benchmarks for BAGEL, SFT, and the two ablated variants. Several per-benchmark patterns are worth noting. On MMSI, all RL variants outperform SFT, indicating that even without $r^{\text{vis}}$ or $\mathcal{L}_{\text{NFT}}$, the text-branch RL alone provides gains on multi-image spatial tasks. On CV-Bench 3D, both BRAID and w/o $r^{\text{vis}}$ recover the SFT regression (82.17 vs.\ 81.67), whereas w/o $\mathcal{L}_{\text{NFT}}$ does not (81.50), suggesting that optimizing the image branch is necessary to recover 3D spatial understanding. The perception benchmarks (BLINK, VStar, CV-Bench 2D) show the largest gap between the full model and ablations, confirming that fine-grained visual tasks benefit most from the combination of image-branch RL and dense process reward.

\subsection{Analysis of Reasoning Capacity}
\begin{table}[ht]
\centering
\caption{Performance comparison of different methods on Vstar and SAT benchmarks
under majority voting (maj@n) and average (avg@n) at $n\in\{3,5,9,17\}$.}
\label{tab:maj_avg}
\setlength{\tabcolsep}{4pt}
\renewcommand{\arraystretch}{1.1}
\begin{tabular}{llcccccccc}
\toprule
\multirow{2}{*}{\textbf{Benchmark}} & \multirow{2}{*}{\textbf{Method}}
 & \multicolumn{4}{c}{\textbf{maj@n}} & \multicolumn{4}{c}{\textbf{avg@n}} \\
\cmidrule(lr){3-6}\cmidrule(lr){7-10}
 &       & \textbf{$n{=}3$} & \textbf{$n{=}5$} & \textbf{$n{=}9$} & \textbf{$n{=}17$}
         & \textbf{$n{=}3$} & \textbf{$n{=}5$} & \textbf{$n{=}9$} & \textbf{$n{=}17$} \\
\midrule
\multirow{3}{*}{\textbf{Vstar}}
 & BAGEL & 62.82 & 64.39 & 65.53 & 65.97 & 58.81 & 59.16 & 59.39 & 59.22 \\
 & SFT   & 65.97 & 67.54 & 66.49 & 67.53 & 62.13 & 61.15 & 61.08 & 61.19 \\
\cmidrule(l){2-10}
 & BRAID & 66.49 & 69.11 & 68.06 & 70.68 & 64.82 & 64.98 & 64.94 & 63.53 \\
\midrule
\multirow{3}{*}{\textbf{SAT}}
 & BAGEL & 45.33 & 52.00 & 53.67 & 52.33 & 47.67 & 46.93 & 47.04 & 47.78 \\
 & SFT   & 54.67 & 53.33 & 56.67 & 56.67 & 49.55 & 50.27 & 51.44 & 51.59 \\
\cmidrule(l){2-10}
 & BRAID & 55.33 & 53.33 & 60.67 & 61.33 & 58.11 & 59.93 & 61.48 & 60.75 \\
\bottomrule
\end{tabular}
\end{table}
Table~\ref{tab:maj_avg} provides the full Maj@$n$ and Avg@$n$ results across $n\in\{3,5,9,17\}$ on VStar and SAT, corresponding to the curves in Figure~\ref{fig:maj_avg}. BRAID consistently achieves the highest scores at all sampling budgets, with the gap widening as $n$ increases. Notably, on VStar the Maj@17 gap between BRAID and SFT is +3.15, whereas the Avg@17 gap is +2.34, indicating that BRAID benefits from both higher per-sample quality and greater diversity. On SAT, BRAID's Avg@$n$ improves monotonically from 58.11 to 60.75 as $n$ grows, while SFT and BAGEL plateau or decline, suggesting that RL produces a policy whose samples are more complementary and less redundant.

\subsection{Analysis of Training Recipe}
\begin{table}[ht]
\centering
\caption{Effect of data-allocation regimes ($\mathcal{D}^{2:1}$, $\mathcal{D}^{1:1}$, $\mathcal{D}^{\cap}$) at the SFT and RL stages across seven benchmarks. Best results in \textbf{bold} and second best \underline{underlined}. ``--'' indicates that we did not run RL from this SFT checkpoint due to its noticeable weaknesses on several benchmarks.}
\label{tab:data_split_result}
\setlength{\tabcolsep}{5pt}
\renewcommand{\arraystretch}{1.3}
\resizebox{\textwidth}{!}{%
\begin{tabular}{ll cccc ccc >{\columncolor{cyan!10}}c}
\toprule
\multirow{2}{*}{\textbf{Stage}} & \multirow{2}{*}{\textbf{Data Regime}}
 & \multicolumn{4}{c}{\textbf{Spatial Reasoning}}
 & \multicolumn{3}{c}{\textbf{Perception}}
 & \multirow{2}{*}{\textbf{Avg.}} \\
\cmidrule(lr){3-6} \cmidrule(lr){7-9}
 & & \textbf{MMSI} & \textbf{SAT} & \textbf{MMVP} & \textbf{CV-Bench 3D}
 & \textbf{BLINK} & \textbf{VStar} & \textbf{CV-Bench 2D} & \\
\midrule
% \multicolumn{2}{l}{\textit{Baseline}} \\
\multicolumn{2}{l}{BAGEL} & 26.60 & 44.67 & 75.67 & \textbf{83.41} & 54.76 & 57.59 & 73.53 & 59.46 \\
\midrule
\multirow{2}{*}{SFT}
 & $\mathcal{D}^{1:1}$ ($\omega{=}0$)   & 26.70 & 46.67 & 77.33 & 80.91 & \underline{58.34} & 58.63 & 65.24 & 59.12 \\
 & $\mathcal{D}^{2:1}$ ($\omega{=}0$)   & 29.20 & 49.67 & 78.67 & 81.67 & 57.02 & 62.82 & 74.85 & 61.99 \\
\midrule
\multirow{3}{*}{RL}
 & $\mathcal{D}^{1:1}$ ($\omega{=}0$)   & --    & --    & --    & --    & --    & --    & --    & --    \\
 & $\mathcal{D}^{2:1}$ ($\omega{=}0$)   & \underline{31.40} & \underline{49.33} & \underline{79.67} & \underline{81.83} & 57.76 & \underline{64.92} & \underline{74.88} & \underline{62.83} \\
 & $\mathcal{D}^{\cap}$ ($\omega{\approx}0.32$) & \textbf{31.80} & \textbf{58.67} & \textbf{81.33} & 82.17 & \textbf{57.76} & \textbf{68.35} & \textbf{76.26} & \textbf{65.19} \\
\bottomrule
\end{tabular}%
}
\end{table}
Table~\ref{tab:data_split_result} presents the full results under the three data-allocation regimes discussed in Section~\ref{subsec:analysic}. The $\mathcal{D}^{1:1}$ SFT checkpoint underperforms BAGEL on CV-Bench 2D (65.24 vs.\ 73.53), a $-$8.29 drop that we attribute to the reduced SFT budget failing to cover sufficient visual perception data. By contrast, the $\mathcal{D}^{2:1}$ regime provides a stable SFT initialization that enables effective downstream RL. The replay regime $\mathcal{D}^{\cap}$ further improves upon $\mathcal{D}^{2:1}$ RL on all benchmarks except BLINK (tied at 57.76), with the largest gains on SAT (+9.34) and VStar (+3.43). This pattern suggests that replaying SFT prompts under on-policy rollouts is particularly beneficial for tasks requiring multi-step spatial reasoning, where the model can consolidate learned strategies while exploring new ones.

\begin{table}[ht]
\centering
\caption{Effect of the rollout-to-update batch size ratio across seven spatial reasoning and perception benchmarks. ``Ratio'' denotes the rollout batch size to update batch size ratio used during RL training. Best results in \textbf{bold} and second best \underline{underlined}.}
\label{tab:on_off_policy}
\setlength{\tabcolsep}{5pt}
\renewcommand{\arraystretch}{1.3}
\resizebox{\textwidth}{!}{%
\begin{tabular}{l cccc ccc >{\columncolor{cyan!10}}c}
\toprule
\multirow{2}{*}{\textbf{Model}}
 & \multicolumn{4}{c}{\textbf{Spatial Reasoning}}
 & \multicolumn{3}{c}{\textbf{Perception}}
 & \multirow{2}{*}{\textbf{Avg.}} \\
\cmidrule(lr){2-5} \cmidrule(lr){6-8}
 & \textbf{MMSI} & \textbf{SAT} & \textbf{MMVP} & \textbf{CV-Bench 3D}
 & \textbf{BLINK} & \textbf{VStar} & \textbf{CV-Bench 2D} & \\
\midrule
BAGEL & 26.60 & 44.67 & 75.67 & \underline{83.41} & 54.76 & 57.59 & 73.53 & 59.46 \\
SFT   & \underline{29.20} & 49.67 & 78.67 & 81.67 & 57.02 & \underline{62.82} & 74.85 & \underline{61.99} \\
\midrule
Ratio = 1:1 & 24.60 & \underline{50.00} & \underline{79.67} & \textbf{84.83} & \textbf{59.07} & 60.20 & \underline{75.12} & 61.93 \\
Ratio = 2:1 & \textbf{31.80} & \textbf{58.67} & \textbf{81.33} & 82.17 & \underline{57.76} & \textbf{68.35} & \textbf{76.26} & \textbf{65.19} \\
\bottomrule
\end{tabular}%
}
\end{table}
Table~\ref{tab:on_off_policy} reports the complete benchmark results for the rollout-to-update ratio analysis. The 2:1 ratio achieves the best average (65.19 vs.\ 61.93), with pronounced advantages on MMSI (+7.20), SAT (+8.67), and VStar (+8.15). The 1:1 ratio, despite its overall lower performance, achieves the highest score on CV-Bench 3D (84.83) and BLINK (59.07), both exceeding even the 2:1 setting. We hypothesize that the more aggressive update frequency acts as implicit exploration pressure, which benefits benchmarks where the SFT initialization is already close to a local optimum but fails to generalize on tasks requiring stable policy improvement over many steps.

\clearpage
\section{Interleave Reasoning Case}\label{app:case}

We present a qualitative comparison of BRAID and its ablated variants on a visual search task from the VStar benchmark.

\begin{tcolorbox}[center,title=\large\centering{Question and Answer}]
\textbf{Problem Image:}\\[4pt]
\centering\includegraphics[width=0.55\linewidth]{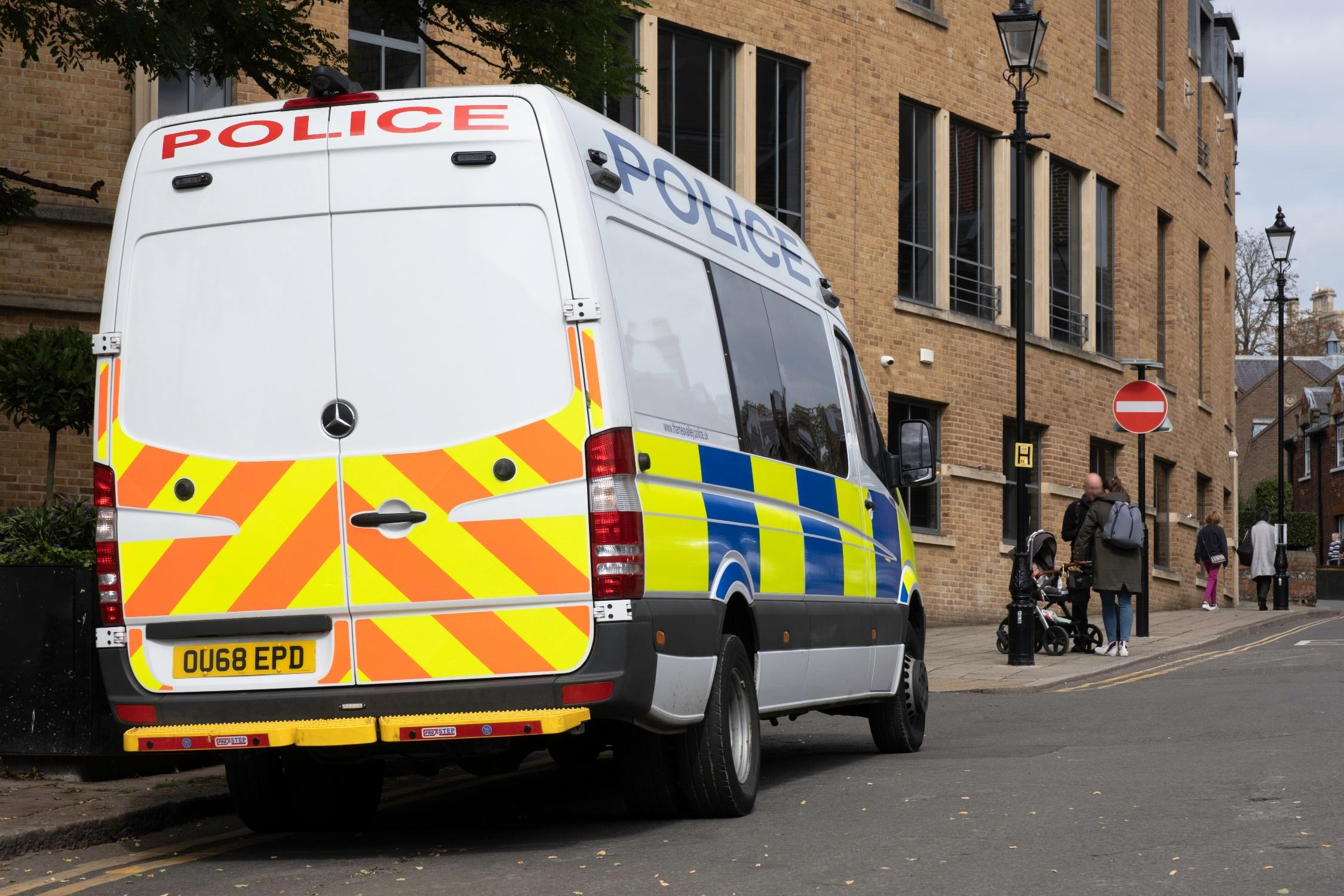}\\[6pt]
\raggedright
\textbf{Question:} What is the license plate number of the police van?\\
\textbf{Options:} (A) OU68 EPD \quad (B) QU68 EPO \quad (C) OV68 EDP \quad (D) OU86 EPD\\[4pt]
\textbf{Ground Truth:} (A) OU68 EPD
\end{tcolorbox}

\begin{tcolorbox}[center,breakable,title=\large\centering{BRAID --- Rollout 1 \textcolor{green!50!black}{(\ding{51} Correct)}}]
\texttt{<think>}To find the correct answer, first examine the image and locate the police van, which is the main vehicle in the scene. Focus on the rear section of the van, as the license plate is always displayed at the back of public service vehicles. On the bottom center of the rear doors, just above the yellow safety strip and below the right door handle, there is a yellow plate with black text. This plate is the vehicle's registration number. Carefully read the sequence of characters on this plate: it consists of a two-letter vanity plate followed by three digits, then another three characters. The style and spacing match a typical license plate, and the characters are clearly separated by spaces. That specific yellow plate, with the visible text inside it, is where you should place the red bounding box.\texttt{</think>}\texttt{<image\_start>}\\[4pt]
\textbf{Generated Image:}\\
\includegraphics[width=0.5\linewidth]{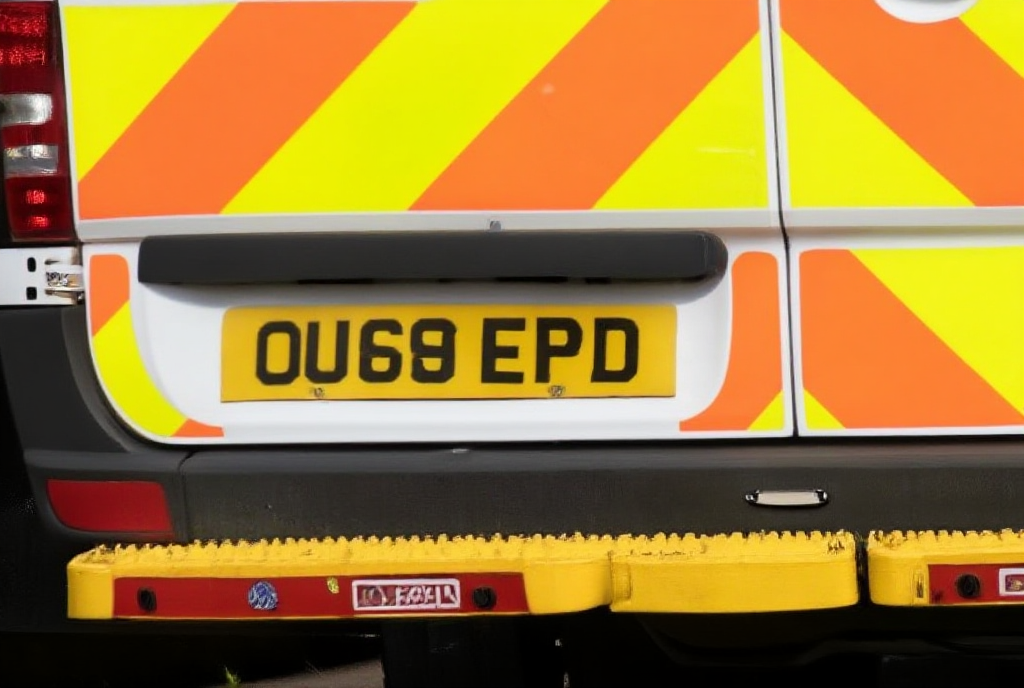}\\[4pt]
\texttt{<image\_end>}\texttt{<think>}With the yellow license plate at the lower center of the rear doors now highlighted, you can focus on the black text printed on the yellow background. The sequence is a two big characters, then a gap, followed by three digits, then another gap and three more characters. Comparing that exact sequence to the options, only the option (A) OU688EPO matches this pattern: the digit in the first sequence is the same as the second one in the options, the second sequence in the second and the two last characters match in the two letters and you can see on the plate. Because the highlighted area is the only area you can clearly read the plate, this identified is the correct answer: A.\texttt{</think>}\texttt{<answer>}A\texttt{</answer>}
\end{tcolorbox}

\begin{tcolorbox}[center,breakable,title=\large\centering{BRAID --- Rollout 2 \textcolor{green!50!black}{(\ding{51} Correct)}}]
\texttt{<think>}To locate the relevant region, first look at the police van in the image. The question asks for the license plate, so focus on the rear of the vehicle. At the bottom center of the back, just above the bumper and between the two rear door sections, there is a rectangular plate with a black background and yellow border. Inside this rectangle, there are black characters arranged in a combination of letters and numbers. This rectangular area, containing the black-and-yellow license plate, is where the red bounding box should be drawn.\texttt{</think>}\texttt{<image\_start>}\\[4pt]
\textbf{Generated Image:}\\
\includegraphics[width=0.6\linewidth]{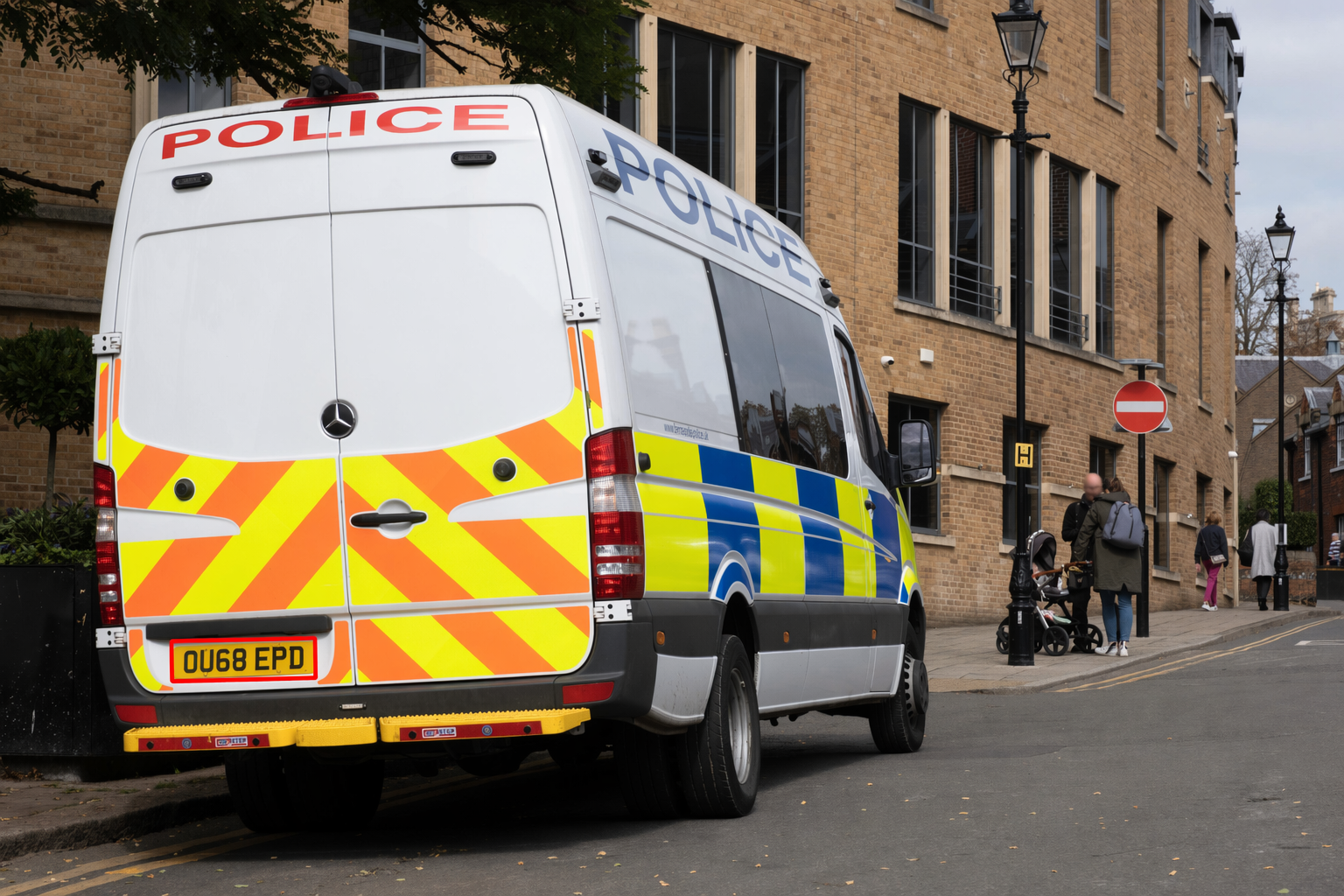}\\[4pt]
\texttt{<image\_end>}\texttt{<think>}With the correct license plate now clearly highlighted, you can concentrate on reading the characters inside the rectangle. The plate shows a two-letter country code starting with ``OU'', then the number ``68'' repeated twice, and the letters ``EPD'' at the end. Matching this full sequence against the answer choices, it corresponds exactly to option (A), so the correct answer is A.\texttt{</think>}\texttt{<answer>}A\texttt{</answer>}
\end{tcolorbox}

\begin{tcolorbox}[center,breakable,title=\large\centering{BRAID w/o $r^{\text{vis}}$ --- Rollout 1 \textcolor{green!50!black}{(\ding{51} Correct, but hallucinated content)}}]
\texttt{<think>}To locate the relevant region, first focus on the back of the police van, since license plates are typically placed there. Scan the lower central part of the rear doors, just above the bumper and below the rear windows. There you can see a rectangular yellow plate with a black border and black text on it. This plate contains the vehicle's registration number. The red bounding box should be drawn tightly around this yellow rectangular plate at the bottom middle of the rear doors, enclosing all the black alphanumeric characters printed on it.\texttt{</think>}\texttt{<image\_start>}\\[4pt]
\textbf{Generated Image:}\\
\includegraphics[width=0.5\linewidth]{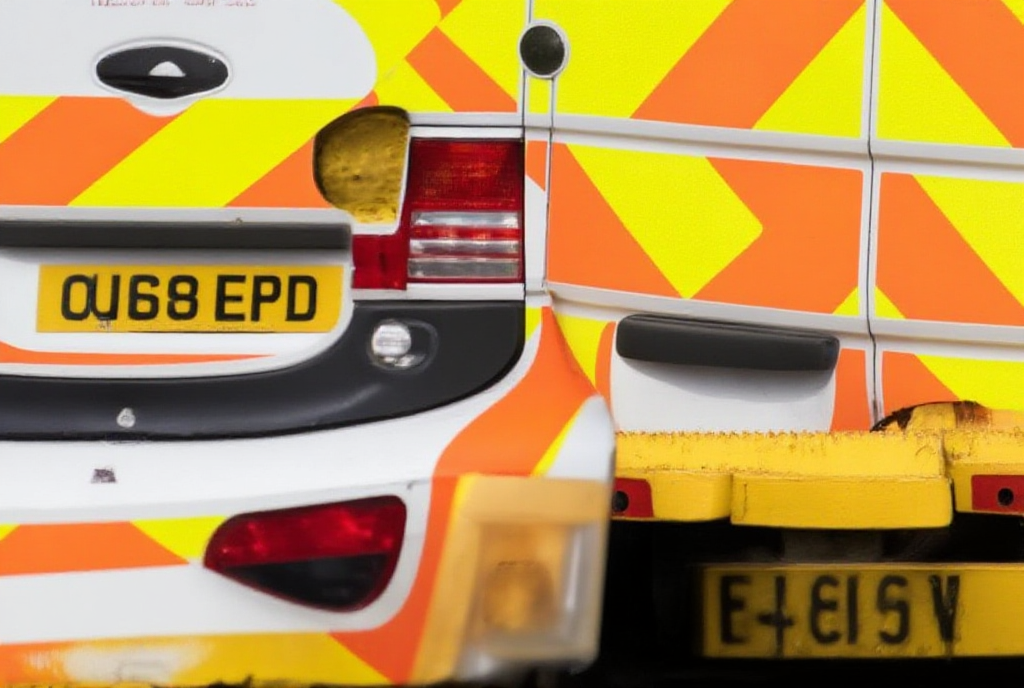}\\[4pt]
\texttt{<image\_end>}\texttt{<think>}With the license plate now clearly highlighted, you can read the characters printed on it. The plate shows a combination starting with the letters ``OU68'' followed by ``EPD'' and ending with the same letter ``D'' at the far right. Comparing this complete sequence with the multiple-choice options, it matches exactly option (A), so the correct answer is A.\texttt{</think>}\texttt{<answer>}A\texttt{</answer>}
\end{tcolorbox}

\begin{tcolorbox}[center,breakable,title=\large\centering{BRAID w/o $\mathcal{L}_{\text{NFT}}$ --- Rollout 1 \textcolor{green!50!black}{(\ding{51} Correct, but misleading image)}}]
\texttt{<think>}To locate the relevant region, first focus on the police van in the image. The question asks about the license plate, so look at the rear of the vehicle. At the bottom center of the back, just above the bumper and between the two rear doors, there is a rectangular plate with a yellow background and black text. This is the license plate whose number needs to be read. The red bounding box should be drawn tightly around this rectangular plate at the license plate position on the back of the police van.\texttt{</think>}\texttt{<image\_start>}\\[4pt]
\textbf{Generated Image:}\\
\includegraphics[width=0.5\linewidth]{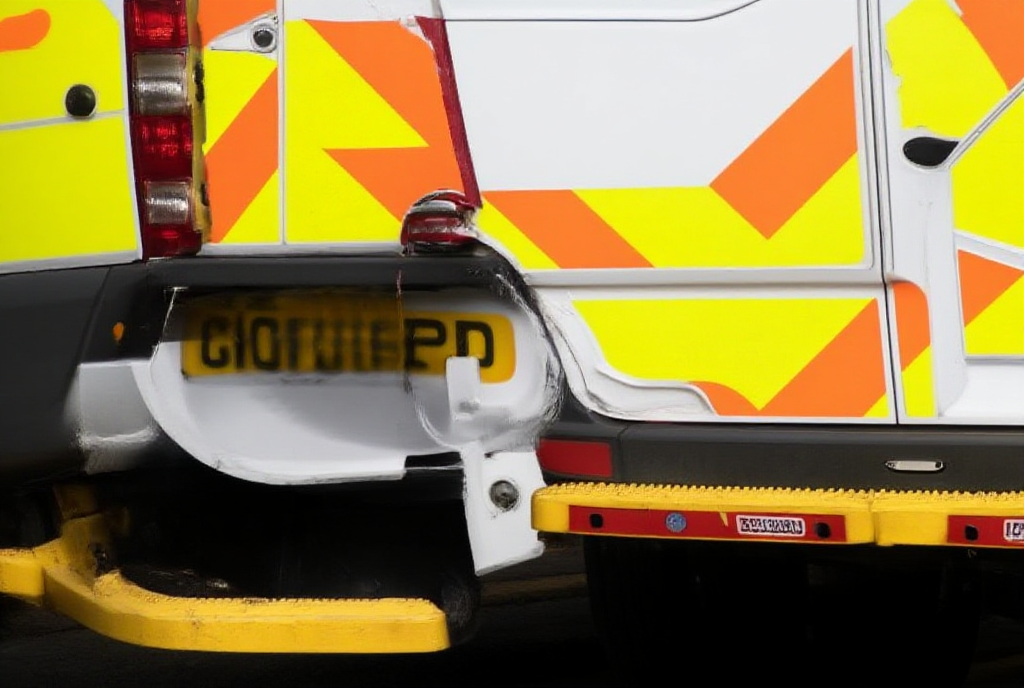}\\[4pt]
\texttt{<image\_end>}\texttt{<think>}From the cropped view of the license plate area, I can now examine the characters more closely. The yellow plate is visible in the left-center portion of the image. Reading the black characters from left to right, the first group appears to be ``GI01'' and the second group reads ``EPD''. The character shapes are fairly clear --- the first letter has a distinct curve consistent with ``G'' rather than ``O'', and the second character is a straight vertical stroke matching ``I''. The digits in the middle appear to be ``01'' based on their rounded and straight forms. Therefore the full registration reads ``GI01 EPD''. Matching this against the available options, this corresponds to option (A).\texttt{</think>}\texttt{<answer>}A\texttt{</answer>}
\end{tcolorbox}

\paragraph{Analysis.}
This case reveals several observations about the role of image quality in TIT-CoT reasoning:

(1)~\textbf{BRAID produces diverse and faithful visual intermediates.} The two BRAID rollouts adopt distinct visual strategies: Rollout~1 generates a tightly cropped view of the license plate for precise character reading, while Rollout~2 produces a full-scene image with the plate region highlighted. Both yield high-fidelity images where ``OU68 EPD'' is clearly legible, directly supporting downstream text reasoning.

(2)~\textbf{Without $r^{\text{vis}}$, images contain hallucinated artifacts.} The w/o $r^{\text{vis}}$ variant generates an image containing a hallucinated second license plate (``E-EI SV'') alongside the actual plate. Although the model correctly identifies the relevant plate here, the fabricated visual content introduces unnecessary ambiguity that would likely mislead reasoning on harder instances.

(3)~\textbf{Without $\mathcal{L}_{\text{NFT}}$, images degrade and mislead reasoning.} The w/o $\mathcal{L}_{\text{NFT}}$ variant generates a distorted, blurry crop where the plate characters are garbled. The text model hallucinates ``GI01 EPD'' from the corrupted visual evidence and arrives at the correct answer through faulty reasoning (incorrectly mapping ``GI01 EPD'' to option A). The reasoning chain is fundamentally unreliable: the model cannot distinguish correct from fabricated visual evidence.

(4)~\textbf{Correct answers can mask fragile reasoning.} All variants arrive at the correct answer, yet only BRAID produces a trajectory where the visual evidence genuinely supports the conclusion. The ablated variants achieve correctness \emph{despite} their visual intermediates, not \emph{because} of them, a distinction that manifests as lower robustness under Maj@$n$ evaluation (Section~\ref{subsec:results}).

% \clearpage
\section{Text-to-Image Generation Quality}\label{app:t2i_case}

Beyond interleaved reasoning, we examine whether the RL training in BRAID also improves standalone text-to-image generation quality. We compare generations from BAGEL, SFT, BRAID w/o $r^{\text{vis}}$, BRAID w/o $\mathcal{L}_{\text{NFT}}$, and the full BRAID on three prompts that test different aspects of instruction following: spatial compositionality, numerical accuracy, and creative concept binding.

\begin{figure}[ht]
\centering
\setlength{\tabcolsep}{2pt}
\begin{tabular}{@{}c@{\hspace{3pt}}c@{\hspace{3pt}}c@{\hspace{3pt}}c@{\hspace{3pt}}c@{\hspace{3pt}}c@{}}
\textbf{Prompt} & \textbf{BAGEL} & \textbf{SFT} & \textbf{w/o $\mathcal{L}_{\text{NFT}}$} & \textbf{w/o $r^{\text{vis}}$} & \textbf{BRAID} \\[3pt]
\rotatebox{90}{\parbox{2.5cm}{\centering A cat sitting inside a fishbowl}} &
\includegraphics[width=0.17\linewidth]{Figs/t2i_case/bagel_case_1.png} &
\includegraphics[width=0.17\linewidth]{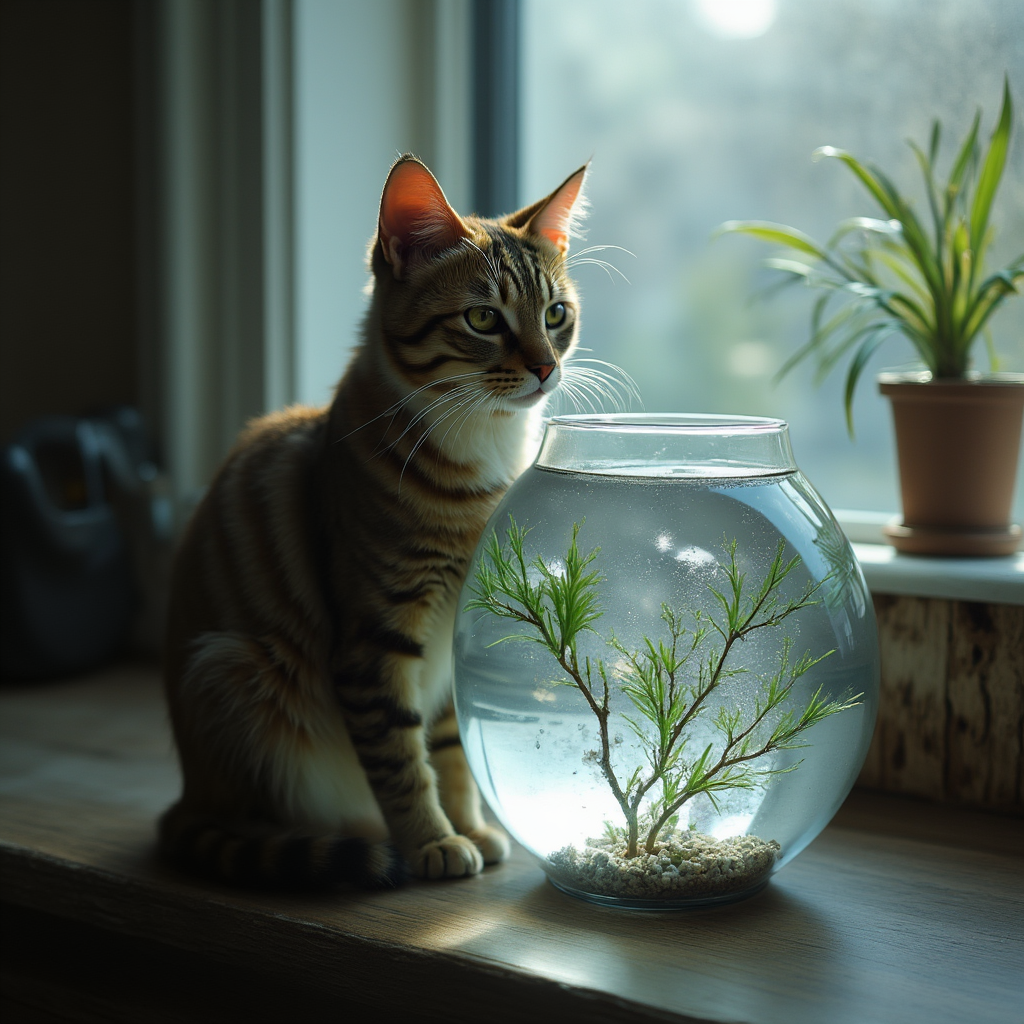} &
\includegraphics[width=0.17\linewidth]{Figs/t2i_case/braid_wo_nft_1.png} &
\includegraphics[width=0.17\linewidth]{Figs/t2i_case/braid_wo_r_vis_case_1.png} &
\includegraphics[width=0.17\linewidth]{Figs/t2i_case/braid_case_1.png} \\[3pt]
\rotatebox{90}{\parbox{2.5cm}{\centering Exactly five balloons}} &
\includegraphics[width=0.17\linewidth]{Figs/t2i_case/bagel_case_2.png} &
\includegraphics[width=0.17\linewidth]{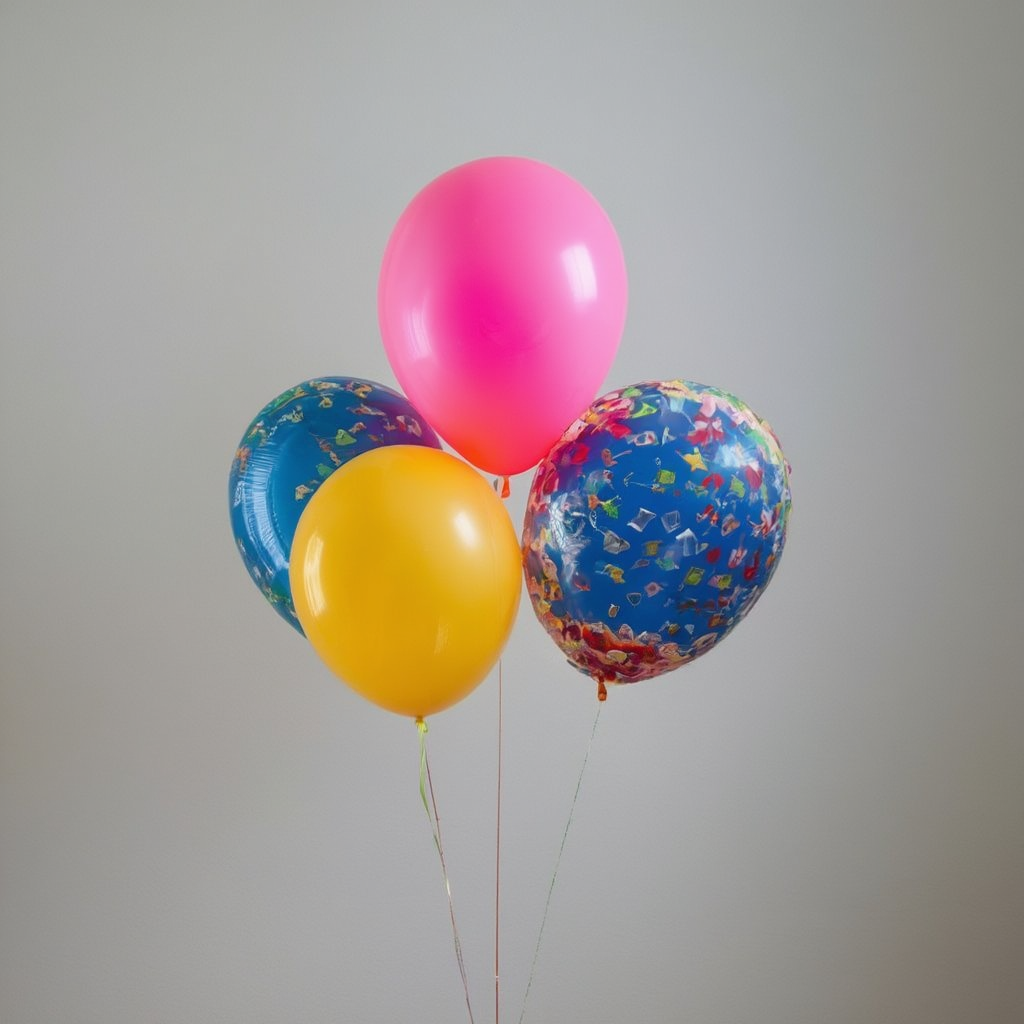} &
\includegraphics[width=0.17\linewidth]{Figs/t2i_case/braid_wo_nft_2.png} &
\includegraphics[width=0.17\linewidth]{Figs/t2i_case/braid_wo_r_vis_case_2.png} &
\includegraphics[width=0.17\linewidth]{Figs/t2i_case/braid_case_2.png} \\[3pt]
\rotatebox{90}{\parbox{2.5cm}{\centering Hourglass with galaxies as sand}} &
\includegraphics[width=0.17\linewidth]{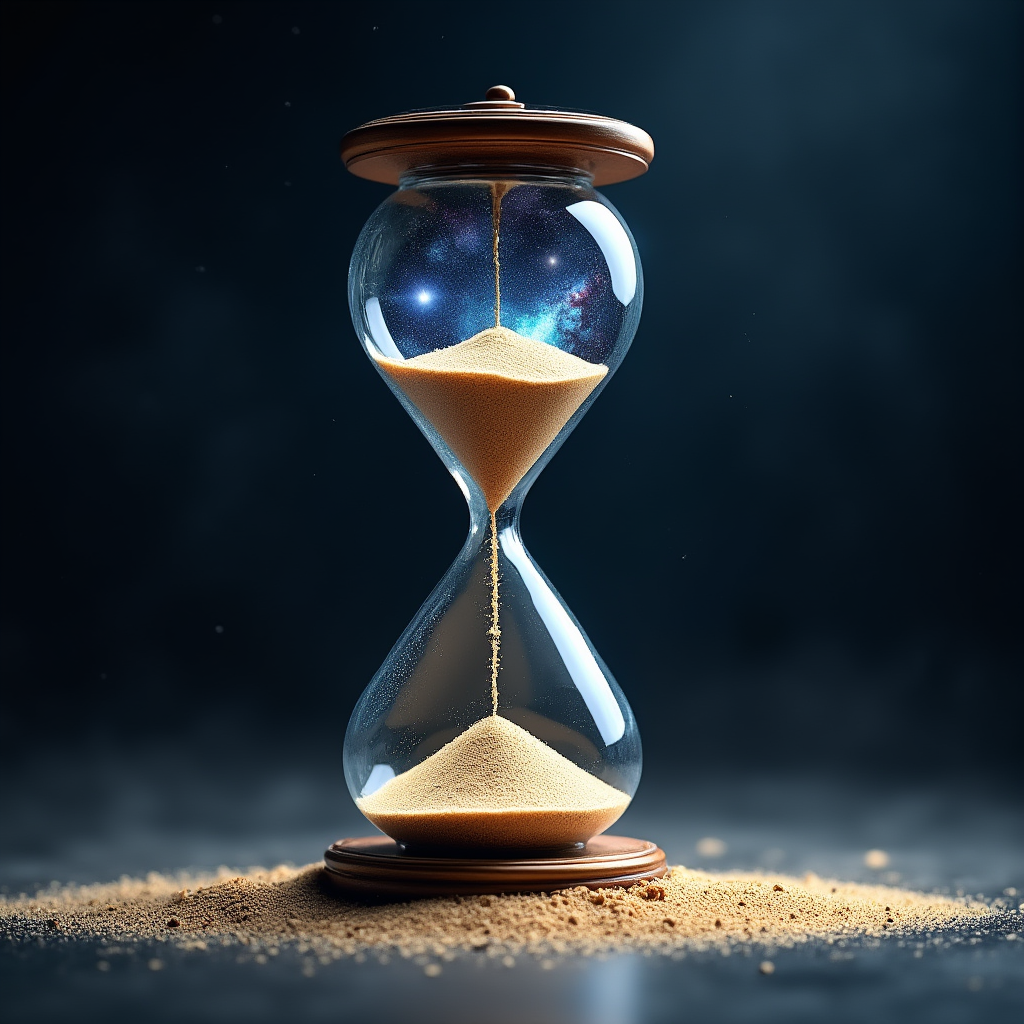} &
\includegraphics[width=0.17\linewidth]{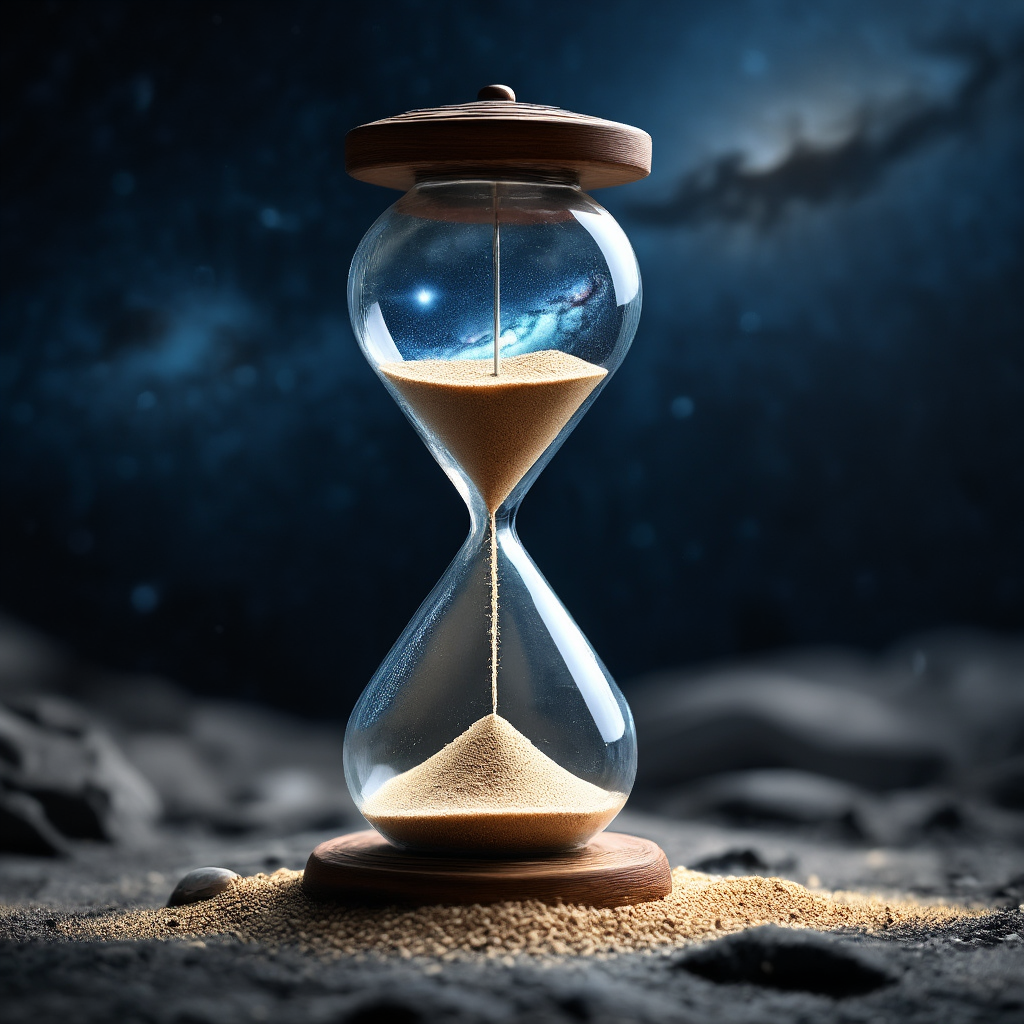} &
\includegraphics[width=0.17\linewidth]{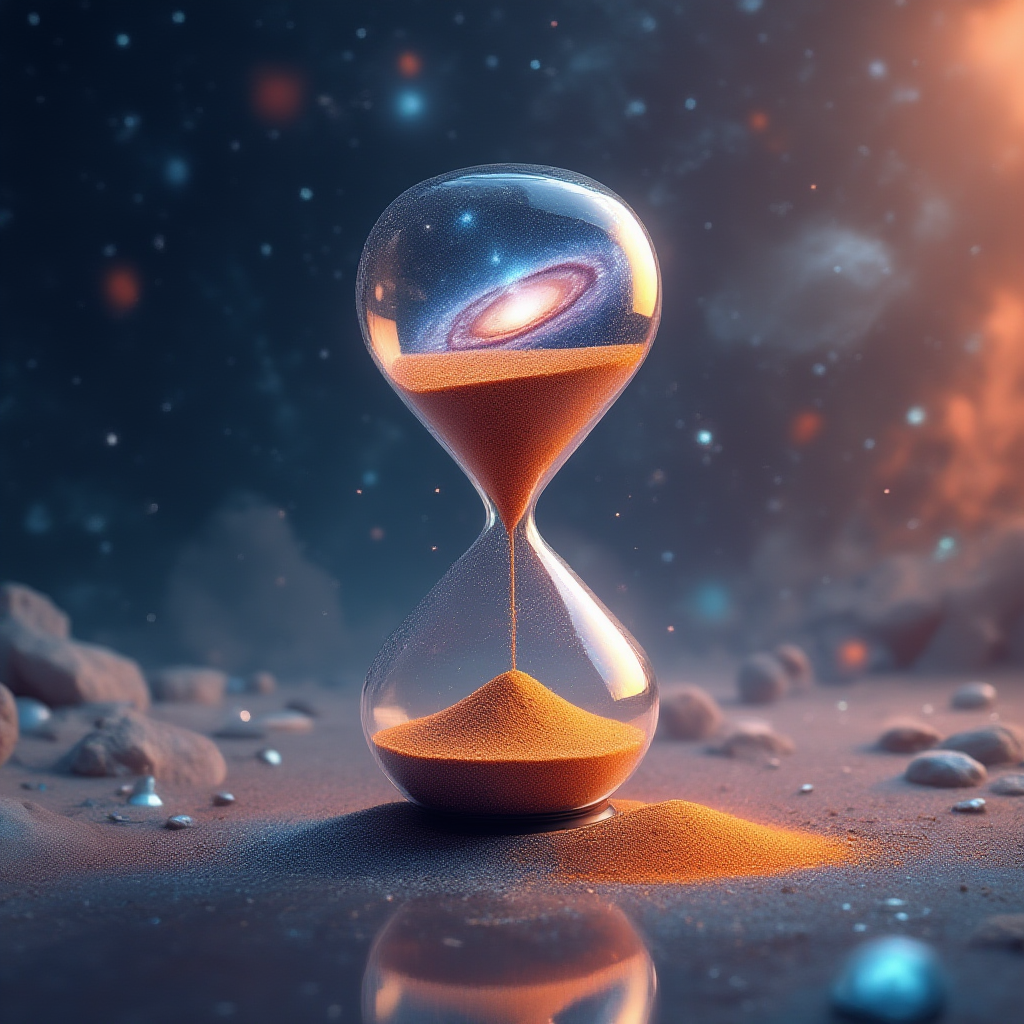} &
\includegraphics[width=0.17\linewidth]{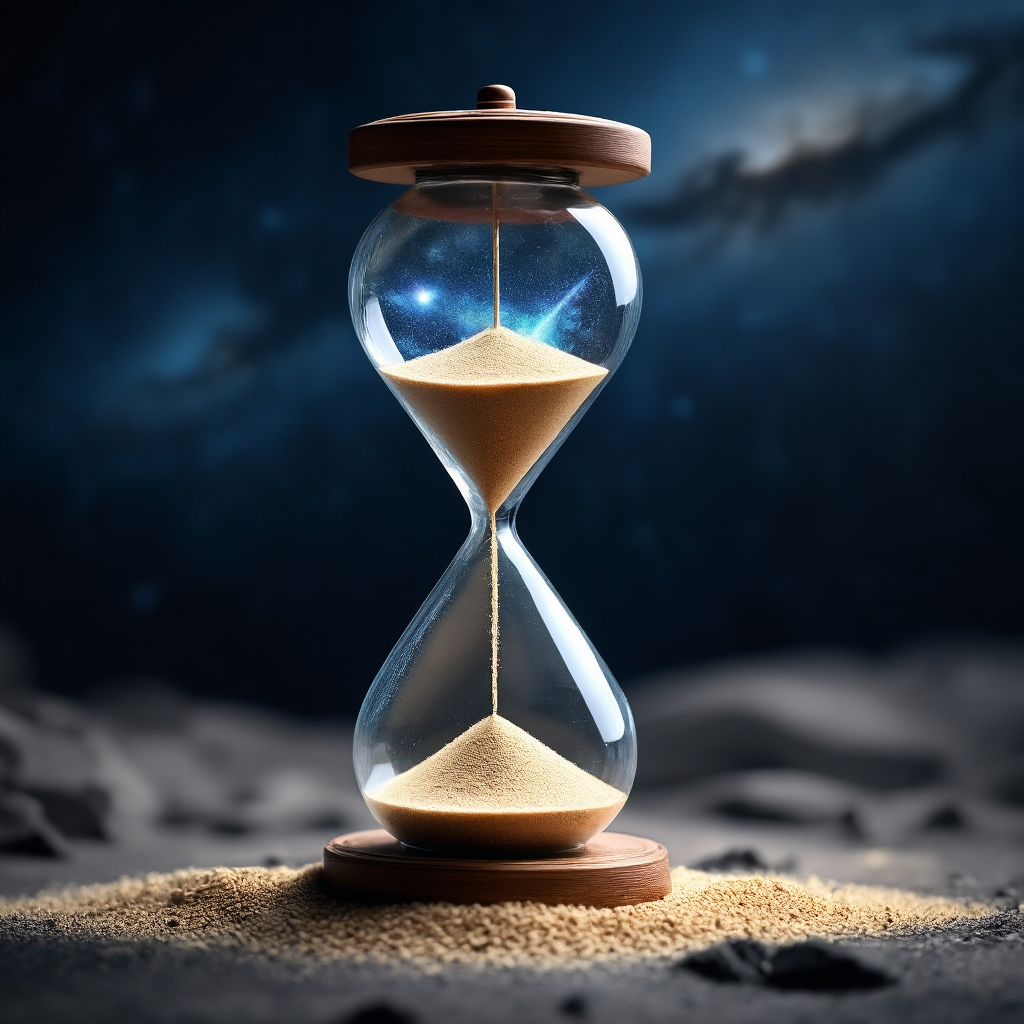} &
\includegraphics[width=0.17\linewidth]{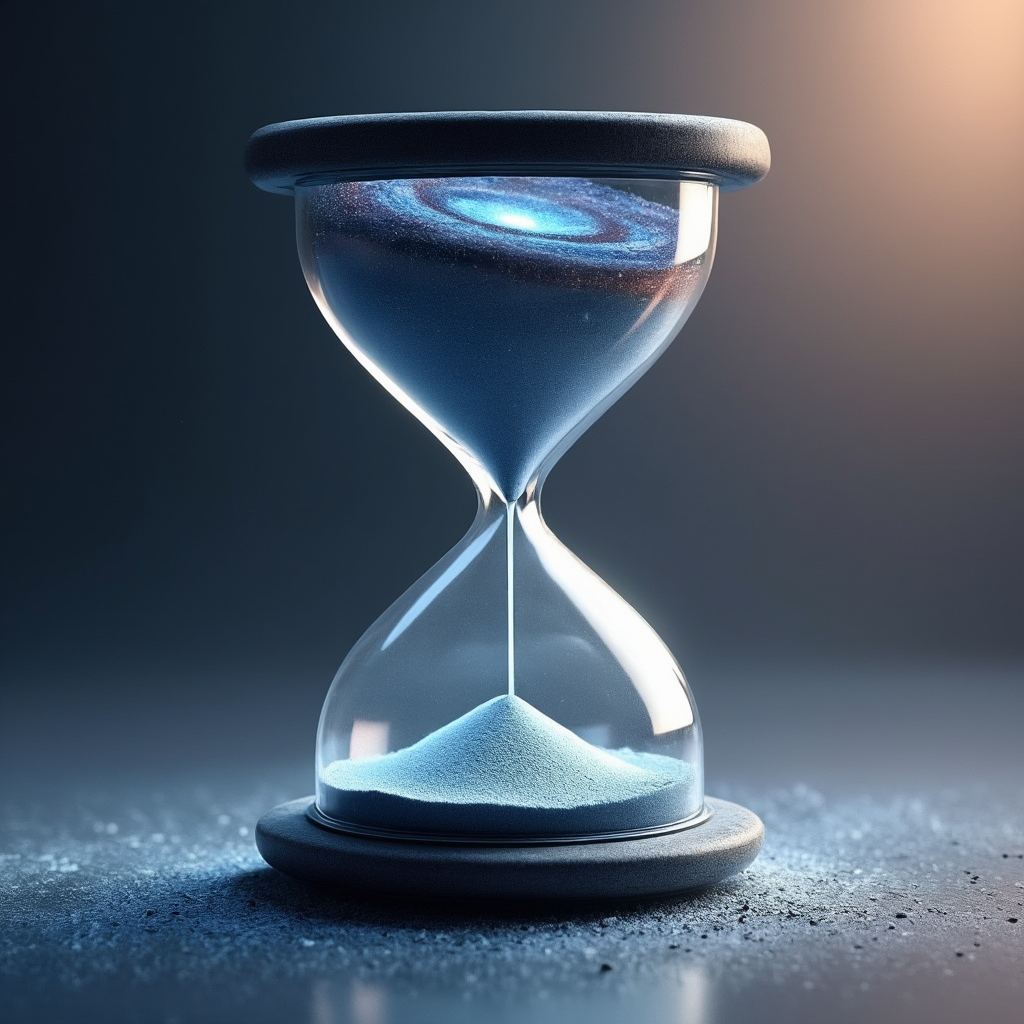} \\
\end{tabular}
\caption{Text-to-image generation comparison across prompts testing spatial compositionality (Row 1), numerical accuracy (Row 2), and creative concept binding (Row 3).}
\label{fig:t2i_case}
\end{figure}

\paragraph{Analysis.}
The generation results in Figure~\ref{fig:t2i_case} reveal that BRAID's RL training yields improvements in both image quality and instruction adherence:

(1)~\textbf{Spatial compositionality.} For ``A cat sitting inside a fishbowl'', BAGEL, SFT, and w/o $\mathcal{L}_{\text{NFT}}$ all place the cat \emph{next to} the fishbowl rather than inside it, failing to satisfy the spatial relation specified by the prompt. Both w/o $r^{\text{vis}}$ and BRAID correctly render the cat inside the bowl, but BRAID additionally preserves higher visual fidelity with a fish coexisting in the scene.

(2)~\textbf{Numerical accuracy.} For ``Exactly five balloons'', BAGEL generates approximately eight, SFT produces four, w/o $\mathcal{L}_{\text{NFT}}$ produces an over-saturated chaotic scene with many balloons, and w/o $r^{\text{vis}}$ generates seven to eight. Only BRAID correctly generates exactly five balloons, demonstrating that the joint RL objective improves precise counting ability.

(3)~\textbf{Creative concept binding.} For ``Hourglass with galaxies as sand'', BAGEL, SFT, and w/o $r^{\text{vis}}$ all produce an hourglass with regular sand and merely place it against a galaxy background, failing to bind the ``galaxies as sand'' concept. The w/o $\mathcal{L}_{\text{NFT}}$ variant shows a galaxy swirl inside the hourglass, and BRAID further improves this with a clearly visible galaxy serving as the flowing sand material.

These observations suggest that the vision-thinking process reward $r^{\text{vis}}$ provides optimization direction that transfers beyond the interleaved reasoning setting to general text-to-image generation. The reward signal, which evaluates visual correctness and faithfulness during TIT-CoT training, implicitly teaches the image branch to better follow compositional instructions. Notably, w/o $\mathcal{L}_{\text{NFT}}$ consistently shows the weakest instruction following among RL-trained variants, confirming that without the DiffusionNFT gradient pathway, the image branch cannot internalize the reward-driven improvements.

% \clearpage
% \input{Texs/7_checklist}

\end{document}